\documentclass{article}
\usepackage{PRIMEarxiv}

\usepackage[T1]{fontenc}
\usepackage{amsmath,amssymb,amsthm,mathtools,bm,mathrsfs}

\usepackage{silence}
\usepackage{microtype}
\usepackage{enumitem}
\usepackage{float}
\usepackage{booktabs,array}
\usepackage{xcolor}
\usepackage{graphicx}
\usepackage[numbers]{natbib}
\usepackage{url}
\usepackage[
    colorlinks=true,
    citecolor=green!50!black,
    linkcolor=red!65!black,
    urlcolor=blue!60!black,
    bookmarksnumbered=true
]{hyperref}
\usepackage{tikz}
\usetikzlibrary{arrows.meta}
\usepackage[nameinlink,noabbrev]{cleveref}
\crefname{definition}{Definition}{Definitions}
\Crefname{definition}{Definition}{Definitions}
\crefname{remark}{Remark}{Remarks}
\crefname{proposition}{Proposition}{Propositions}
\crefname{theorem}{Theorem}{Theorems}
\crefname{assumption}{Assumption}{Assumptions}
\crefname{lemma}{Lemma}{Lemmas}
\crefname{section}{Section}{Sections}
\crefname{figure}{Figure}{Figures}
\crefname{table}{Table}{Tables}
\crefname{appendix}{Appendix}{Appendices}
\Crefname{appendix}{Appendix}{Appendices}
\usepackage[skins,breakable]{tcolorbox}
\theoremstyle{definition}
\newtheorem{definition}{Definition}

\tcolorboxenvironment{definition}{
  colback=yellow!8,
  colframe=yellow!35!black,
  boxrule=0.5pt,
  arc=2pt,
  left=6pt,
  right=6pt,
  top=5pt,
  bottom=5pt
}
\tcolorboxenvironment{proposition}{
  enhanced,
  breakable,
  colback=black!7,
  colframe=black!7,
  boxrule=0pt,
  sharp corners,
  left=6pt,
  right=6pt,
  top=5pt,
  bottom=5pt,
  before skip=8pt,
  after skip=8pt
}

\definecolor{theoremblue}{RGB}{235,244,252}
\tcolorboxenvironment{theorem}{
  enhanced,
  breakable,
  colback=theoremblue,
  colframe=theoremblue,
  boxrule=0pt,
  sharp corners,
  left=6pt,
  right=6pt,
  top=5pt,
  bottom=5pt,
  before skip=8pt,
  after skip=8pt
}
\makeatletter
\renewcommand{\@makefnmark}{%
  \hbox{\@textsuperscript{%
    \normalfont\color{purple!65!black}\@thefnmark}}}
\makeatother
\let\oldeqref\eqref
\renewcommand{\eqref}[1]{%
  {\hypersetup{linkcolor=blue!60!black}\oldeqref{#1}}%
}

\hypersetup{
  pdftitle={Width and Depth Limits Commute during Training
    in Linear Residual Networks},
  pdfauthor={Ruijia Zhang and Soufiane Hayou},
  pdfsubject={Width--depth commutativity for trained linear residual networks},
  pdfkeywords={residual networks, width--depth commutativity, random matrices,
    free probability, Volterra equations}
}
\usepackage{aliascnt}

\newtheorem{theorem}{Theorem}[section]

\newaliascnt{proposition}{theorem}
\newtheorem{proposition}[proposition]{Proposition}
\aliascntresetthe{proposition}

\newaliascnt{lemma}{theorem}
\newtheorem{lemma}[lemma]{Lemma}
\aliascntresetthe{lemma}
\theoremstyle{definition}
\newaliascnt{assumption}{theorem}

\aliascntresetthe{assumption}

\newaliascnt{remark}{theorem}
\newtheorem{remark}[remark]{Remark}
\aliascntresetthe{remark}

\newcommand{\E}{\mathbb E}

\newcommand{\tr}{\operatorname{tr}}
\newcommand{\op}{\mathrm{op}}

\newcommand{\dd}{\,\mathrm d}

\newcommand{\Ltwo}{\mathbb L_2}
\title{Width and Depth Limits Commute During \\ Training in Linear Residual Networks}
\author{
  Ruijia Zhang \qquad Soufiane Hayou \\
  Department of Applied Mathematics and Statistics \\
  Johns Hopkins University
}
\date{}

\begin{document}
\maketitle

\begin{abstract}
Recent work showed that width and depth limits in residual networks with branches scaled with $1/\sqrt{depth}$ commute at \emph{initialization}: no matter how those limits are taken (e.g. width first, then depth, or depth first, then width, or with a fixed width/depth ratio, etc), the network output converges to the same limit at initialization. In this paper, we extend this result to commutativity \emph{during training} for linear residual networks: we show that under the same $1/\sqrt{depth}$ branch scaling (Depth-$\mu$P) and with $\mu$P parameterization for the width, for any fixed number of gradient steps with gradient clipping, the network output converges in $\Ltwo$ as $\min\{width, depth\} \to \infty$, implying commutativity of the width and depth limits during training. We further show that such result does not hold without gradient clipping, and a weaker result can be obtained in that case under the weaker convergence in probability.
\end{abstract}


\section{Introduction}
Modern neural networks are increasingly scaled along multiple architectural
dimensions. The rapid growth and success of large language and vision models has further motivated the study of neural network scaling limits as
a way to understand how model behavior changes with scale \citep{kaplan2020scaling,zhai2022scaling}. Among these limits, width and depth play distinct roles: increasing width gives rise to asymptotic regimes such as neural tangent kernel and mean-field
limits \citep{jacot2018neural,mei2019mean, yang2020feature}, while increasing depth connects
residual networks to many continuous-time dynamics \citep{haber2018stable, Hanin2019product, hayou2021stableresnet, hayou2022on, yang2024tensor, peluchetti2020resnetdiffusion, marion2022scaling,bordelon2023depthwisehyperparametertransferresidual, chizat2026hiddenwidthdeepresnets}. Understanding how these two
limits interact is therefore fundamental when both dimensions grow
simultaneously.

A central question is whether the width $n$ and depth $L$ limits commute, that is,
whether the same asymptotic behavior is obtained regardless of the order or
relative rate at which the two dimensions scale. For residual networks
with branches scaled by $L^{-1/2}$, \citet{hayou2023width} showed that the width and depth limits commute at
initialization, so the same limit is obtained along arbitrary joint
width--depth scaling paths. This leaves open whether commutativity persists
during training. At initialization, the residual
weights are independent across layers, a
property used in existing joint commutativity analysis \citep{hayou2023width,hayou2023commutativewidthdepthscaling}. During training, gradient descent couples the weights across layers. As a result, the arguments developed for initialization output convergence cannot directly extend. It is therefore
unclear whether the width and depth limits continue to commute under the same $L^{-1/2}$ branch scaling. 

We first show that standard gradient descent exhibits a sharp distinction between convergence modes: after a single step, the width and depth limits of the trained output commute in probability, but not in $\Ltwo$. Specifically, we show that for every fixed width $n$, the trained output converges in probability as $L\to\infty$, while its second moment diverges in the large-depth limit. This implies that commutativity cannot hold in $\Ltwo$, but we show that it holds in probability. We then show that, under clipped gradient descent, the width and depth limits commute in $\Ltwo$ for any fixed number of training steps, with convergence rates $n^{-1/2}$ in width and $L^{-1/2}$ in depth.

Our proof proceeds by taking the two limits in sequence. We first establish
the depth limit at finite width by comparing the trained residual network
with a continuous-depth model, with estimates \emph{uniform in width}, which allow us to then
take the infinite-width limit of this continuous-depth system and identify the deterministic trained output.

\paragraph{Related Work.} Infinite-width limits have been studied through Gaussian-process, neural
tangent kernel, and mean-field descriptions \citep{jacot2018neural,mei2019mean,neal2012bayesian,lee2017deep}. Similarly, infinite-depth limits of residual networks were studied as continuous-time descriptions of network dynamics \citep{haber2018stable, hayou2021stableresnet, marion2022scaling, peluchetti2020resnetdiffusion, bordelon2023depthwisehyperparametertransferresidual, chizat2026hiddenwidthdeepresnets}, and more heuristic analyses of training dynamics \citep{yang2024tensor, bordelon2023depthwisehyperparametertransferresidual}. The interaction between these two limits was studied in \cite{hayou2023commutativewidthdepthscaling, hayou2023width}, who proved width--depth
commutativity for the neural covariance of ResNets with residual branches
scaled by $L^{-1/2}$. \citet{hayou2023commutativewidthdepthscaling} extended this analysis to
more general residual scalings, while non-commutative limits have been
identified for some MLP and ResNet parameterizations \citep{li2022neural,noci2023shaped}. More recently, \citet{chizat2026hiddenwidthdeepresnets} studied the large-depth training limit of ResNets scaled with $L^{-1}$ branch multiplier at fixed network width. \citet{chaintron2026resnetsshapessizesconvergence} is arguably the closest work to our paper. The authors showed that commutativity of the width and depth limits holds during training with $L^{-1}$ branch scaling. Such scaling leads to ODE-style infinite-depth dynamics, which is substantially different from the SDE-style dynamics obtained with $L^{-1/2}$. Our work establishes commutativity of the width and depth limits during training under the $L^{-1/2}$ branch scaling consistent with Depth-$\mu$P \citep{yang2024tensor}. These two scaling choices lead to significantly different dynamics and therefore require different proof machinery to establish commutative limits.
\section{Setup and Notation}
\subsection{Model and parameterization}
We consider a linear residual network with input dimension $d$, width $n$,
and depth $L$. For a fixed nonzero input $x\in\mathbb R^d$, the hidden states and
scalar output are given by

\begin{equation}
\label{eq:model}
\begin{aligned}
  &Y_0        = W_0\, x, \\
  &Y_\ell     = Y_{\ell-1} + L^{-1/2}\, W_\ell\, Y_{\ell-1},
               \qquad \ell = 1, \dots, L, \\
  &f_{n,L}(x) = \frac{1}{n}\, \xi^\top Y_L,
\end{aligned}
\end{equation}
where $W_0\in\mathbb R^{n\times d}$ is the input weight matrix,
$W_\ell\in\mathbb R^{n\times n}$ for $\ell=1,\ldots,L$ are the residual
weights, and $\xi\in\mathbb R^n$ is the output weight.\footnote{We consider a one-dimensional output to simplify the main results' statements. The extension to multi-dimensional outputs is given
in \cref{app:finite-training-set}.} Since we will study both the width and depth limits, we parameterize certain hyperparameters w.r.t to these scaling dimensions: we use the $L^{-1/2}$ branch multiplier to stabilize the depth limit, which was suggested by Depth-$\mu$P \citep{yang2024tensor}, and we also stabilize the width limit by randomly initializing the weights as follows: $(W_0)_{ij}\sim\mathcal N(0,d^{-1})$,
$(W_\ell)_{ij}\sim\mathcal N(0,n^{-1})$, and
$\xi_i\sim\mathcal N(0,1)$, which also corresponds to $\mu$P scaling \citep{yang2020feature}.

To simplify the analysis, we keep $W_0$ and $\xi$ fixed and train only the residual weights
$W_1,\ldots,W_L$. Our results can be easily extended to the case where $W_0$ and $\xi$ are also trainable.\footnote{We do not include such extension since it does not add any valuable insights to the proof machinery.} Throughout the paper, $\|z\|_2$ denotes the Euclidean norm of a vector
$z$. For a random vector $Z$, we write
$\|Z\|_{\Ltwo}:=(\E\|Z\|_2^2)^{1/2}$; for a scalar random variable, the
inner norm is understood as the absolute value. For $u,v\in\mathbb R^n$,
set $\langle u,v\rangle_n:=n^{-1}u^\top v$ and
$\|u\|_n:=\langle u,u\rangle_n^{1/2}=n^{-1/2}\|u\|_2$. We use the notation $q_0 := \|x\|_2^2/d$. Conditional on $x$,
initial hidden state satisfies $Y_0\sim\mathcal N(0,q_0 I_n)$.

Throughout the paper, all randomness comes solely from the initialization. 
\subsection{Width--Depth Commutativity}
\label{sec:commutativity}

We formalize width--depth commutativity following
\citet{hayou2023commutativewidthdepthscaling}. Unless stated otherwise, commutativity refers
to convergence in $\Ltwo$ of the network output as $\min\{n, L\} \to \infty$.

\begin{definition}[Width--depth commutativity]
\label{def:commutativity}
Fix a training iteration $k$. We say that the width and depth limits commute
if there exists a deterministic limit $f_\infty^{(k)}$ such that
\[
\left\|
f_{n,L}^{(k)}-f_\infty^{(k)}
\right\|_{\Ltwo}
\longrightarrow 0
\qquad
\text{as } \min(n,L)\to\infty,
\]
where $f_{n,L}^{(k)}$ and $f_\infty^{(k)}$ are both evaluated at some input $x \in \mathbb R^{d}$.
\end{definition}

We also consider a weaker notion of commutativity, requiring only convergence in probability. This is a useful distinction because we will establish commutativity in probability but not in $\Ltwo$ in some cases.

\begin{definition}[Commutativity in probability]
\label{def:commutativity-prob}
Fix a training iteration $k$. We say that the width and depth limits commute
in probability if there exists a deterministic limit $f_\infty^{(k)}$ such
that, for every $\varepsilon>0$,
\[
\mathbb P\!\left(
\left|f_{n,L}^{(k)}-f_\infty^{(k)}\right|>\varepsilon
\right)
\longrightarrow 0
\qquad
\text{as } \min(n,L)\to\infty,
\]
where $f_{n,L}^{(k)}$ and $f_\infty^{(k)}$ are both evaluated at some input $x \in \mathbb R^{d}$.
\end{definition}

Both notions require the limiting output to be independent of the relative
growth of width and depth. Moreover, commutativity in $\Ltwo$ implies
commutativity in probability.
\begin{figure}
\vspace{-2em}
    \centering
    \includegraphics[width=1.0\linewidth]{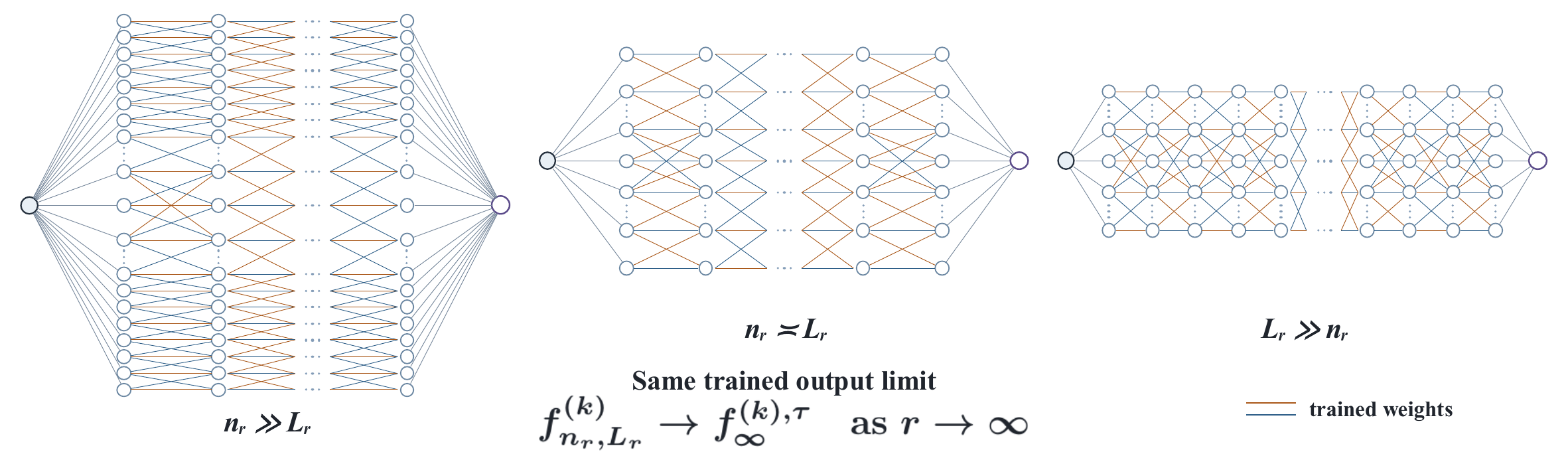}
    \caption{
\textbf{Commutativity of the width and depth limits.}
Illustration of three relative order regimes for $(n_r,L_r)$ as $n_r,L_r\to\infty$.
For the same fixed input and any fixed training iteration $k$, all three
paths converge to the same trained output limit $f_\infty^{(k)}$.
}
    \label{fig:illustration}
\end{figure}

\subsection{Training dynamics}
Without loss of generality, we train model \eqref{eq:model} by minimizing a quadratic loss function evaluated on single input--target pair\footnote{While we consider a single input, the extension to finite training sets is given
in \cref{app:finite-training-set}.}
$(x,y_\star)$
\begin{equation*}
\mathcal L(W_1,\ldots,W_L)
=
\frac12\bigl(f_{n,L}(x)-y_\star\bigr)^2 .
\end{equation*}
This setup is used to alleviate the notation in the proofs, and it was used in \cite{hayou2025proof,yang2024tensor} for similar purposes. Our results can be readily extended to the case of a finite training dataset at the cost of redundant computations.

To minimize the loss, we update the weights using gradient descent (GD)
\begin{equation}
W_\ell^{(k+1)}
=
W_\ell^{(k)}
-
\eta\nabla_{W_\ell^{(k)}}\mathcal L,
\qquad
\ell=1,\ldots,L,
\quad
k=0,\ldots,T-1,
\label{eq:gd-update}
\end{equation}
where $\eta>0$ is the learning rate, fixed independently of $n$ and $L$, and $W_\ell^{(0)}$ are the weights at initialization, specified above. These initialization rules combined with the depth multiplier $L^{-1/2}$ follow the scaling rules suggested by \cite{yang2020feature} and \cite{yang2024tensor} to guarantee stability and feature learning as model width $n$ and depth $L$ grow.

For a training step $k$, let
$A_\ell^{(k)}:=I+L^{-1/2}W_\ell^{(k)}$. The corresponding forward and
backward states are defined by

\begin{equation}
\begin{aligned}
&\textbf{Forward:}  && Y_0^{(k)} = Y_0,
  && Y_\ell^{(k)} = A_\ell^{(k)}\, Y_{\ell-1}^{(k)},
  && \ell = 1, \dots, L, \\[6pt]
&\textbf{Backward:} && B_L^{(k)} = \xi,
  && B_{\ell-1}^{(k)} = \bigl(A_\ell^{(k)}\bigr)^{\!\top} B_\ell^{(k)},
  && \ell = L, \dots, 1.
\end{aligned}
\label{eq:forward-backward}
\end{equation}
Since we consider a single fixed training pair $(x, y_\star)$, we omit it from the notation and write the neural network output at training step $k$, $f_{n,L}^{(k)}$. We also write
$\varepsilon_{n,L}^{(k)}:=f_{n,L}^{(k)}-y_\star$. Straightforward calculations give the gradients of the network output w.r.t the hidden weights $\nabla_{W_\ell^{(k)}} f_{n,L}^{(k)}
=
\frac{1}{n\sqrt L}
B_\ell^{(k)}(Y_{\ell-1}^{(k)})^\top .$ The next section studies commutativity after taking one gradient step.

\section{Width and Depth Limits at \texorpdfstring{$k=1$}{k=1}}

We first develop an intuitive understanding of the problem by analyzing the network output after taking a single gradient step. We show in this case that the width and depth limits commute in probability, but not in
$\Ltwo$ (\Cref{def:commutativity}, \Cref{def:commutativity-prob}). The reason behind this discrepancy is that for a fixed width $n$, the network output diverges in $\Ltwo$ as depth $L \to \infty$, while it converges in probability. We first state the convergence in probability result.

\begin{proposition}[Commutativity in probability after one GD step]
After one gradient step,
\[
f_{n,L}^{(1)}
\xrightarrow[\min\{n,L\}\to\infty]{\mathbb P}
f_\infty^{(1)}
=
\eta y_\star q_0 e,
\]
\label{prop:gd-one-step-prob}
\end{proposition}

The proof is given in \cref{app:gd-one-step-prob}, which also establishes
commutativity in probability after any fixed finite number of gradient steps. The next result shows that commutativity does not extend to
$\Ltwo$. In fact, after one gradient step, the depth limit already fails
in $\Ltwo$ at every fixed finite width.

\begin{proposition}[Depth-wise $\Ltwo$ divergence under gradient descent]
\label{prop:gd-depth-divergence}
After the first gradient iteration, for every fixed finite $n\geq 1$, $\eta>0$,
and $y_\star\in\mathbb R$, $\lim_{L\to\infty}
\mathbb E\bigl|f_{n,L}^{(1)}\bigr|^2
=
+\infty.$

\end{proposition}
Together, \cref{prop:gd-one-step-prob,prop:gd-depth-divergence} show that,
after one gradient step, the width and depth limits commute in probability,
while the depth-first $\Ltwo$ limit does not exist. Indeed, for each $n$ one may choose $L_n\ge n$ sufficiently large so that
$\mathbb E|f_{n,L_n}^{(1)}|^2\ge n$. Then
$\min\{n,L_n\}\to\infty$. Hence, an
unrestricted joint $\Ltwo$ limit cannot hold under ordinary gradient
descent. As shown in \cref{app:gd-depth-divergence}, this failure arises
from the unbounded first GD update as depth increases. This motivates the clipped gradient
descent, a widely used technique to stabilize gradients during training. Specifically, we use the following form of clipped gradient descent with clipping radius
$\tau>0$: 

We denote the clipped training trajectory by the superscript
$\tau$, and define $J_{n,L}^{(k),\tau}
:=
\frac{1}{L}\sum_{\ell=1}^L
\|Y_{\ell-1}^{(k),\tau}\|_n^2
\|B_\ell^{(k),\tau}\|_n^2 .$ Let
$\gamma_{n,L}^{(k),\tau}
:=\eta\varepsilon_{n,L}^{(k),\tau}
\min\{1,\tau/\left(|\varepsilon_{n,L}^{(k),\tau}|
\sqrt{J_{n,L}^{(k),\tau}}\right)\}$,\footnote{If the denominator vanishes,
the multiplier inside the minimum is set to one.}
where $\varepsilon_{n,L}^{(k),\tau}
:=f_{n,L}^{(k),\tau}-y_\star$.
The clipped updates are given by
\begin{equation*}
W_\ell^{(k+1),\tau}
=
W_\ell^{(k),\tau}
-
\frac{\gamma_{n,L}^{(k),\tau}}{n\sqrt L}
B_\ell^{(k),\tau}(Y_{\ell-1}^{(k),\tau})^\top.
\end{equation*}

By definition of $\gamma_{n,L}^{(k),\tau}$, $\left(
\sum_{\ell=1}^L
\|W_\ell^{(k+1),\tau}-W_\ell^{(k),\tau}\|_F^2
\right)^{1/2}
=
|\gamma_{n,L}^{(k),\tau}|
\sqrt{J_{n,L}^{(k),\tau}}
\leq
\eta\tau.$

Thus, clipping enforces uniform control of each GD update, independently
of $n$ and $L$. The next section establishes width-depth commutativity in $\Ltwo$ under gradient clipping.

\section{Width--Depth Commutativity during Training}
\label{sec:main-result}

We now state the main result. For any fixed number of training iterations, the output of a linear residual network trained with clipped gradient descent converges to the same deterministic limit along every
joint path $(n,L)\to(\infty,\infty)$.

\begin{theorem}[Commutative Width and Depth Limits during Training]
\label{thm:finiteT-sharp-main}
Fix $T>0$ number of training steps, and $\tau>0$. There exist deterministic limits
$f_{\infty}^{(0),\tau},\ldots,f_{\infty}^{(T),\tau}$ and a finite constant $C_T$, independent of
$n$ and $L$, such that
\begin{equation}
\max_{0\leq k\leq T}
\left\|
f_{n,L}^{(k),\tau}-f_{\infty}^{(k),\tau}
\right\|_{\Ltwo}
\leq
C_T\left(
n^{-1/2}+L^{-1/2}
\right).
\label{eq:main-output-rate}
\end{equation}
\end{theorem}

The bound in \cref{thm:finiteT-sharp-main} holds along any sequence
$n,L\to\infty$, thus implying the existence of the limit as $\min\{n,L\}\to\infty$ and therefore establishing commutativity in $\Ltwo$.
\begin{remark}[Sharpness in width]
\label{rem:width-sharpness}
The $n^{-1/2}$ term in \cref{thm:finiteT-sharp-main} cannot be improved.
At initialization ($k=0$), independence of $\xi$ and $Y_L$ gives $\mathbb E|f_{n,L}^{(0)}|^2
=
\frac{1}{n}\mathbb E\|Y_L^{(0)}\|_n^2
=
\frac{q_0(1+L^{-1})^L}{n}.$
Since $(1+L^{-1})^L\leq e$, we have
$f_{n,L}^{(0)}\to 0$ in $\Ltwo$ as $n\to\infty$, uniformly in $L$. Hence,
$f_{\infty}^{(0),\tau}=0$. \(
\|f_{n,L}^{(0)}-f_{\infty}^{(0),\tau}\|_{\Ltwo}
=
\sqrt{\frac{q_0(1+L^{-1})^L}{n}}
\geq
\sqrt{\frac{q_0}{n}},\)
establishing the sharpness of the $n^{-1/2}$ bound. The same cannot be said about the depth rate $L^{-1/2}$ as we will later provide empirical results showing that it is unlikely to be sharp.
\end{remark}

\subsection{Understanding the complexity of commutativity during training}
The initialization argument in \cite{hayou2023width} does not extend directly to commutativity during training because gradient descent couples the residual blocks across layers.

At initialization, $A_\ell^{(0)}=I+L^{-1/2}W_\ell$ for $\ell=1,\ldots,L$,
where $\{W_\ell\}_{\ell=1}^L$ are independent Gaussian matrices.
Let $t_\ell=\ell/L$. Since $L^{-1/2}W_\ell$ has the same Gaussian
distribution as a matrix-Brownian increment over an interval of length
$1/L$, one can construct a matrix Brownian motion $M$ on the same
probability space such that
\[
M_{t_\ell}-M_{t_{\ell-1}}=L^{-1/2}W_\ell,
\qquad \ell=1,\ldots,L.
\]
Under this coupling,
\[
Y_\ell^{(0)}
=
\bigl(I+M_{t_\ell}-M_{t_{\ell-1}}\bigr)Y_{\ell-1}^{(0)},
\]
which is the Euler discretization of the solution of the SDE
\[
dY_t=dM_t\,Y_t,
\qquad Y_0=W_0x.
\]

Let $\Phi_{s,t}$ denote the solution operator of this SDE, so that
$Y_t=\Phi_{0,t}Y_0$. \citet{hayou2023width} show that this Euler approximation
converges at rate $O(L^{-1/2})$, with a constant independent of the width
$n$. After taking the depth limit, the resulting finite-width SDE is an
interacting particle system. They then compare each coordinate of this
system with its McKean--Vlasov limit and obtain a $O(n^{-1/2})$
approximation error in $\Ltwo$. Together, these two estimates establish that the width and depth
limits commute at initialization.

However, after one GD update, the residual blocks are no longer independent across layers. Indeed,
\[
A_\ell^{(1),\tau}
= I+\frac{1}{\sqrt L}W_\ell^{(1),\tau} =
A_\ell^{(0)}
-
\frac{\gamma_{n,L}^{(0),\tau}}{nL}
B_\ell^{(0)}(Y_{\ell-1}^{(0)})^\top ,
\]
where $Y_{\ell-1}^{(0)}
=
A_{\ell-1}^{(0)}\cdots A_1^{(0)}Y_0, \quad
B_\ell^{(0)}
=
(A_{\ell+1}^{(0)})^\top\cdots(A_L^{(0)})^\top\xi$ by \eqref{eq:forward-backward}.

The forward state $Y_{\ell-1}^{(0)}$ depends on
$A_1^{(0)},\ldots,A_{\ell-1}^{(0)}$, while the backward state
$B_\ell^{(0)}$ depends on
$A_{\ell+1}^{(0)},\ldots,A_L^{(0)}$.
Therefore, $W_\ell^{(1),\tau}$ depends on the
initial weights for $j\neq\ell$, so the updated residual blocks are coupled across layers and the independent-increment Brownian coupling used at initialization is no longer available. This coupling also changes the structure of the subsequent width limit.
At initialization, the forward state is simply $Y_n^{(0)}(t)=\Phi_{0,t}Y_0.$ 
After one GD update, the discrete forward recursion is
\[
Y_\ell^{(1),\tau}-Y_{\ell-1}^{(1),\tau}
=
\frac{1}{\sqrt L}W_\ell Y_{\ell-1}^{(1),\tau}
-
\frac{\gamma_{n,L}^{(0),\tau}}{nL}
B_\ell^{(0)}
(Y_{\ell-1}^{(0)})^\top
Y_{\ell-1}^{(1),\tau}.
\]
As $L\to\infty$, the Gaussian term converges to the matrix-Brownian
dynamics, while the $L^{-1}$ gradient term becomes a drift in the depth
variable. The continuous-depth state therefore satisfies
\[
\dd Y_n^{(1),\tau}(t)
=
\dd M_t\,Y_n^{(1),\tau}(t)
-
\frac{\gamma_n^{(0),\tau}}{n}
B_n^{(0)}(t)
(Y_n^{(0)}(t))^\top
Y_n^{(1),\tau}(t)\,\dd t .
\]

Integrating this SDE from $0$ to $t$ along the initialization flow $\Phi$
gives
\begin{equation}
Y_n^{(1),\tau}(t)
=
\Phi_{0,t}Y_0
-
\gamma_n^{(0),\tau}
\int_0^t
\Phi_{s,t}B_n^{(0)}(s)
\left\langle
Y_n^{(0)}(s),Y_n^{(1),\tau}(s)
\right\rangle_n
\,\dd s .
\label{eq:one-step-continuous}
\end{equation}

Therefore, proving the width convergence of $Y_n^{(1),\tau}$ also
requires controlling the width limits of $B_n^{(0)}$,
$\gamma_n^{(0),\tau}$, and
$\langle Y_n^{(0)},Y_n^{(1),\tau}\rangle_n$.

\subsection{Proof Overview}
To establish \cref{thm:finiteT-sharp-main}, we decompose the joint limit into
two steps. First, at fixed width $n$, we compare the discrete network with a
continuous-depth model whose output is denoted by $F_n^{(k),\tau}$, obtaining
an $L^{-1/2}$ rate uniformly in $n$. We then compare $F_n^{(k),\tau}$ with its
deterministic infinite-width limit $f_{\infty}^{(k),\tau}$ at rate
$n^{-1/2}$:
\[
f_{n,L}^{(k),\tau}
\;\xrightarrow[\;L\to\infty\;]{\,L^{-1/2}\,}\;
F_n^{(k),\tau}
\;\xrightarrow[\;n\to\infty\;]{\,n^{-1/2}\,}\;
f_{\infty}^{(k),\tau}.
\]

\noindent\textbf{Depth limit.}
For the discrete network, for $t\in I_\ell=(t_{\ell-1},t_\ell]$, define 
\[
V_{n,L}^{(k)}(t)
=
\sum_{m<k}
\frac{\gamma_{n,L}^{(m),\tau}}{n}
B_{n,L}^{(m),\tau}(t_\ell)
\bigl(Y_{n,L}^{(m),\tau}(t_{\ell-1})\bigr)^\top .
\]
Let $Y_n^{(k),\tau}$ and $B_n^{(k),\tau}$ denote the finite-width
continuous-depth forward and backward states, and let
$\gamma_n^{(k),\tau}$ denote the clipped coefficient obtained in the
continuous-depth model.
Their recursive definitions are given in \cref{app:depth-trained}. For the continuous-depth model, define
\[
V_n^{(k)}(t)
=
\sum_{m<k}
\frac{\gamma_n^{(m),\tau}}{n}
B_n^{(m),\tau}(t)
\bigl(Y_n^{(m),\tau}(t)\bigr)^\top .
\]
Let $S_{i+1:j}^{n,L}
:=
A_j^{(0)}\cdots A_{i+1}^{(0)}, \quad
S_{j+1:j}^{n,L}=I,$
denote the discrete initialization propagator. By construction, $S_{i+1:j}^{n,L}$ is the Euler discretization of
$\Phi_{t_i,t_j}$. At the grid points $t_j=j/L$, the discrete and continuous forward
dynamics take the forms
\[
\begin{aligned}
Y_{n,L}^{(k),\tau}(t_j)
&=
S_{1:j}^{n,L}Y_0
-
\frac1L\sum_{\ell=1}^j
S_{\ell+1:j}^{n,L}
V_{n,L}^{(k)}(t_\ell)
Y_{n,L}^{(k),\tau}(t_{\ell-1}),
\\
Y_n^{(k),\tau}(t_j)
&=
\Phi_{0,t_j}Y_0
-
\int_0^{t_j}
\Phi_{s,t_j}
V_n^{(k)}(s)
Y_n^{(k),\tau}(s)\,\dd s .
\end{aligned}
\]

For each fixed width $n$, the standard Euler--Maruyama estimate
\citep{kloeden1977numerical} gives $\left\|
S_{1:j}^{n,L}Y_0-\Phi_{0,t_j}Y_0
\right\|_{\Ltwo}
=
O(L^{-1/2}),$
with a constant that depends on $n$.
This is insufficient for the trained recursion, where
$S_{\ell+1:j}^{n,L}$ and $\Phi_{s,t_j}$ act on $V_{n,L}^{(k)}Y_{n,L}^{(k),\tau}$ and
$V_n^{(k)}Y_n^{(k),\tau}$ generated at
each training iteration. We therefore require an operator-norm estimate
that is uniform in the width. In \cref{app:depth-euler}, we
remove this width dependence and show that
\[
\sup_{n\ge1}\max_{0\le i\le j\le L}
\left\|
\left\|S_{i+1:j}^{n,L}-\Phi_{t_i,t_j}\right\|_{\op}
\right\|_{\Ltwo}
\le
C_{\rm Eul}L^{-1/2},
\]
where $C_{\rm Eul}$ is independent of both $n$ and $L$ (see
\cref{prop:depth-euler}).

We then show by induction that this rate is preserved for any finite number of GD updates, following the proof chain below
\[
\begin{tikzpicture}[
    baseline=(current bounding box.center),
    node distance=0pt,
    every node/.style={inner sep=1pt},
    h/.style={-{Latex[length=2.2mm,width=1.5mm]}, line width=0.9pt},
    v/.style={-{Latex[length=1.8mm,width=1.2mm]}, line width=0.6pt}
]

\node (a1) at (0,1.5)
{$\bigl(Y_{n,L}^{(k),\tau},B_{n,L}^{(k),\tau}\bigr)$};
\node (a2) at (4.1,1.5)
{$\gamma_{n,L}^{(k),\tau}$};
\node (a3) at (7.0,1.5)
{$V_{n,L}^{(k+1)}$};
\node (a4) at (10.8,1.5)
{$\bigl(Y_{n,L}^{(k+1),\tau},B_{n,L}^{(k+1),\tau}\bigr)$};

\node (b1) at (0,0)
{$\bigl(Y_n^{(k),\tau},B_n^{(k),\tau}\bigr)$};
\node (b2) at (4.1,0)
{$\gamma_n^{(k),\tau}$};
\node (b3) at (7.0,0)
{$V_n^{(k+1)}$};
\node (b4) at (10.8,0)
{$\bigl(Y_n^{(k+1),\tau},B_n^{(k+1),\tau}\bigr)$};

\draw[h] (a1.east) -- (a2.west);
\draw[h] (a2.east) -- (a3.west);
\draw[h] (a3.east) -- (a4.west);

\draw[h] (b1.east) -- (b2.west);
\draw[h] (b2.east) -- (b3.west);
\draw[h] (b3.east) -- (b4.west);

\draw[v] (a1.south) -- node[right, xshift=1mm]
{\scriptsize $O(L^{-1/2})$} (b1.north);

\draw[v] (a2.south) -- node[right, xshift=1mm]
{\scriptsize $O(L^{-1/2})$} (b2.north);

\draw[v] (a3.south) -- node[right, xshift=1mm]
{\scriptsize $O(L^{-1/2})$} (b3.north);

\draw[v] (a4.south) -- node[right, xshift=1mm]
{\scriptsize $O(L^{-1/2})$} (b4.north);

\end{tikzpicture}
\]
Clipping yields $\|V_{n,L}^{(k)}\|_{L_t^2(\mathrm F)}
\vee
\|V_n^{(k)}\|_{L_t^2(\mathrm F)}
\le k\eta\tau .$
Since $T$ is fixed, this bound is uniform over $n$ and $L$ for all
$k\le T$. The
$O(L^{-1/2})$ rate is therefore preserved throughout the training induction. This gives the following
width-uniform depth comparison.

\begin{proposition}[Width-uniform depth discretization]
For every fixed integer $T\ge 0$, there exists
$C_{\rm D}<\infty$, independent of $n$ and $L$, such that
\label{prop:main-depth}
\[
\max_{0\le k\le T}
\left\|
f_{n,L}^{(k),\tau}-F_n^{(k),\tau}
\right\|_{\Ltwo}
\le
C_{\rm D}L^{-1/2}.
\]
\end{proposition}

The complete proof is given in \cref{app:depth-trained}.

\noindent\textbf{Width limit.}
The one-step expression \eqref{eq:one-step-continuous} already shows the
structure that persists at later training iterations. At iteration $k$, the forward
state involves $B_n^{(r),\tau}$, $\gamma_n^{(r),\tau}$, and
$\langle Y_n^{(r),\tau},Y_n^{(k),\tau}\rangle_n$ for $r<k$, in addition
to the matrix-Brownian flow $\Phi$. Since $Y_n^{(k),\tau}\in\mathbb R^n$, the state space itself changes
with $n$, so there is no direct statewise comparison across widths.
We therefore introduce an intermediate $n$-dimensional surrogate that
keeps the finite-width matrix flow $\Phi$, while replacing
$\gamma_n^{(r),\tau}$ and the empirical inner products by their limiting
values. Let $y^{(k)}$ and $b^{(k)}$ denote the limiting forward and backward
states, and let $\gamma_k^\tau$ denote the limiting value of
$\gamma_n^{(k),\tau}$ (see Appendix~\ref{app:width-proof}).
Define the surrogate states
\[
\begin{aligned}
\widetilde Y_n^{(k)}(t)
&=
\Phi_{0,t}Y_0
-
\sum_{r<k}\gamma_r^\tau
\int_0^t
\Phi_{s,t}\widetilde B_n^{(r)}(s)
\left\langle y^{(r)}(s),y^{(k)}(s)\right\rangle
\,\dd s,
\\
\widetilde B_n^{(k)}(t)
&=
\Phi_{t,1}^{\top}\xi
-
\sum_{r<k}\gamma_r^\tau
\int_t^1
\Phi_{t,s}^{\top}\widetilde Y_n^{(r)}(s)
\left\langle b^{(r)}(s),b^{(k)}(s)\right\rangle
\,\dd s .
\end{aligned}
\]

Recall that the continuous-depth output is
$F_n^{(k),\tau}:=\langle \xi,Y_n^{(k),\tau}(1)\rangle_n$. The surrogate yields the decomposition
\[
F_n^{(k),\tau}-f_{\infty}^{(k),\tau}
=
\underbrace{
\left\langle
\xi,
Y_n^{(k),\tau}(1)-\widetilde Y_n^{(k)}(1)
\right\rangle_n
}_{\mathrm{(I)}}
+
\underbrace{
\left(
\left\langle
\xi,\widetilde Y_n^{(k)}(1)
\right\rangle_n
-f_{\infty}^{(k),\tau}
\right)
}_{\mathrm{(II)}}.
\]

Term~(I) measures the discrepancy between the finite-width model and the
surrogate, while Term~(II) compares the surrogate output with its
deterministic limit.
To control $\mathrm{(I)}$, we compare the finite-width and surrogate
states inductively through
the following chain:
\[
\begin{tikzpicture}[
    >=Latex,
    h/.style={->, line width=0.9pt},
    v/.style={->, line width=0.7pt},
    every node/.style={align=center}
]

\node (A1) at (0,0)
{$\bigl(Y_n^{(k),\tau},B_n^{(k),\tau}\bigr)$};

\node (A2) at (5.0,0)
{$\gamma_n^{(k),\tau}$};

\node (A3) at (9.4,0)
{$\bigl(Y_n^{(k+1),\tau},B_n^{(k+1),\tau}\bigr)$};

\node (B1) at (0,-1.8)
{$\bigl(\widetilde Y_n^{(k)},\widetilde B_n^{(k)}\bigr)$};

\node (B2) at (5.0,-1.8)
{$\gamma_k^\tau$};

\node (B3) at (9.4,-1.8)
{$\bigl(\widetilde Y_n^{(k+1)},\widetilde B_n^{(k+1)}\bigr)$};

\draw[h] (A1) -- (A2);
\draw[h] (A2) -- (A3);

\draw[h] (B1) -- (B2);
\draw[h] (B2) -- (B3);

\draw[v] (A1) -- node[right] {$O(n^{-1/2})$} (B1);
\draw[v] (A2) -- node[right] {$O(n^{-1/2})$} (B2);
\draw[v] (A3) -- node[right] {$O(n^{-1/2})$} (B3);

\end{tikzpicture}
\]

This gives $\mathrm{(I)}=O(n^{-1/2})$ for every fixed $k$. For $\mathrm{(II)}$, the surrogate output approaches its deterministic
limit in two steps:
\[
\left\langle \xi,\widetilde Y_n^{(k)}(1)\right\rangle_n
\;\xrightarrow{\;n^{-1/2}\;}\;
\mathbb E\!\left[
\left\langle \xi,\widetilde Y_n^{(k)}(1)\right\rangle_n
\,\middle|\,\Phi
\right]
\;\xrightarrow{\;n^{-1}\;}\;
f_{\infty}^{(k),\tau} .
\]
The first step conditions on the matrix flow $\Phi$ and averages out the
Gaussian randomness in $Y_0$ and $\xi$, reducing the conditional mean to
normalized traces of products of $\Phi$. The second step shows that these
finite-width traces converge to their infinite-width limits at rate
$O(n^{-1})$, yielding $f_{\infty}^{(k),\tau}$.

Combining the bounds for $\mathrm{(I)}$ and $\mathrm{(II)}$ gives the
following proposition.

\begin{proposition}[Width limit of the continuous-depth model]
\label{prop:main-width}
For every fixed integer $T\ge 0$, there exists
$C_{\rm W}<\infty$, independent of $n$, such that
\[
\max_{0\le k\le T}
\left\|
F_n^{(k),\tau}-f_{\infty}^{(k),\tau}
\right\|_{\Ltwo}
\le
C_{\rm W}n^{-1/2}.
\]
\end{proposition}
The proof is given in \cref{app:width-proof}.

Propositions~\ref{prop:main-depth} and~\ref{prop:main-width} control the
depth and width errors, respectively. By the triangle inequality,
\[
\max_{0\le k\le T}
\|f_{n,L}^{(k),\tau}-f_{\infty}^{(k),\tau}\|_{\Ltwo}
\le
C_{\rm D}L^{-1/2}+C_{\rm W}n^{-1/2},
\]
which proves \cref{thm:finiteT-sharp-main}.

\section{Experiments}
\label{sec:experiments}

In this section, we empirically study the commutativity of the width and depth
limits during training and characterize their finite-size convergence rates.
Additional experiments in \cref{app:exp} include a single-datum setting
aligned with our theory. Code for reproducing the experiments is available at
\href{https://github.com/Richard-ZZZ/Width_depth_commute}{this repository}.

\subsection{Commutativity during Training}
\label{sec:exp-joint}
We evaluate joint width--depth scaling on synthetic regression and HIGGS
binary classification.

\begin{figure}[t]
\vspace{-2em}
  \centering
  \includegraphics[width=\textwidth]{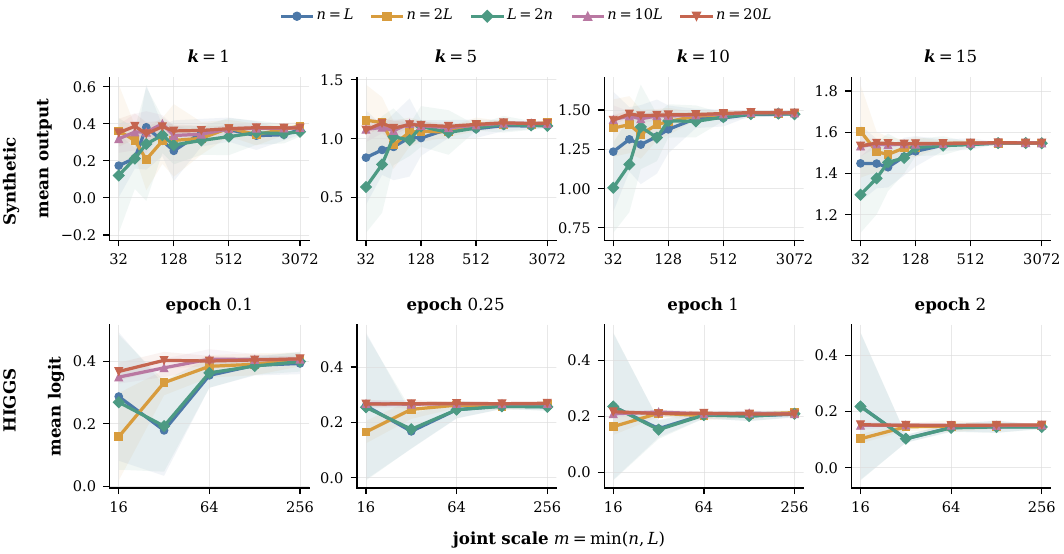}
\caption{
\textbf{Commutativity of the width and depth limits during training.}
Each curve corresponds to a fixed width--depth scaling path and is plotted
against the joint scale $m=\min(n,L)$. The top row shows the synthetic
regression output at training iterations $k=1,5,10,15$, and the bottom row
shows the HIGGS logit margin at four training checkpoints. For each task,
all configurations are evaluated at the same fixed input. Curves show the
mean across 5 model seeds, with shaded regions denoting $\pm1$ standard
deviation.
}
  \label{fig:joint-convergence}
\end{figure}

\paragraph{Synthetic regression.}
We generate $N=128$ samples
$x_i\overset{\mathrm{i.i.d.}}{\sim}\mathcal N(0,\Sigma)$ in dimension
$d=16$, where $\Sigma=Q\Lambda Q^\top$ has a geometrically spaced spectrum
with condition number $10$.
For fixed orthonormal directions $\theta_1,\theta_2$, the regression targets
are
\[
\widetilde y_i
=
\theta_1^\top x_i
+
0.5\bigl((\theta_2^\top x_i)^2-1\bigr),
\qquad
y_i
=
\widetilde y_i
\left(
\frac{1}{N}\sum_{j=1}^{N}\widetilde y_j^2
\right)^{-1/2}.
\]
We train for 15 full-batch GD iterations with quadratic loss.
\paragraph{HIGGS \citep{baldi2014searching}.}
HIGGS is a binary classification benchmark with $28$ input features.
We train the class-$1$ versus class-$0$ logit margin using binary logistic
loss and minibatch SGD for two epochs with batch size $4096$.We use the
same data split and minibatch sequence for every width--depth configuration.

For both tasks, we evaluate five width--depth scaling paths, $n=L$, $n=2L$, $L=2n$, $n=10L$, and $n=20L$, indexed by the common scale
$m=\min(n,L)$. Along each path, both width and depth increase while their
ratio remains fixed. As shown in \cref{fig:joint-convergence}, the outputs differ across paths
at small scales but approach a common value as $m$ increases, consistent
with the commutativity established by our theory.

\begin{figure}[t]
\vspace{-2em}
  \centering
  \includegraphics[width=\textwidth]{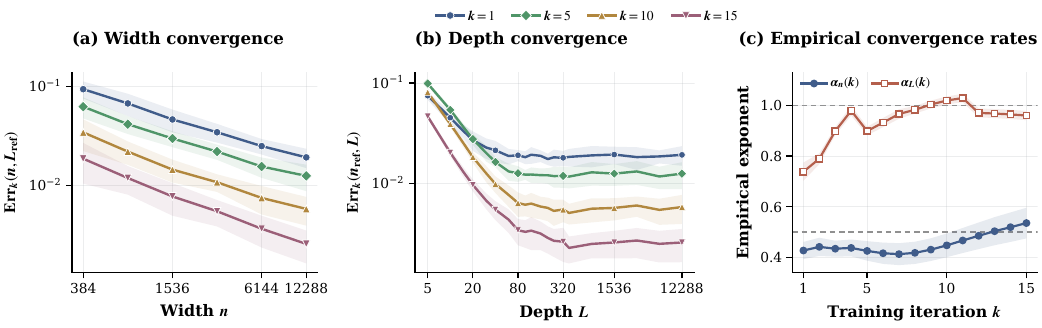}
\caption{
\textbf{Width--depth convergence rates.}
(a) RMS output error as width $n$ increases with depth fixed at
$L=12{,}288$. (b) RMS output error as depth $L$ increases with width fixed at
$n=12{,}288$. Curves in (a)--(b) correspond to training iterations
$k\in\{1,5,10,15\}$. (c) Empirical convergence exponents for width and depth over training. The width rate remains close to $n^{-1/2}$, while the depth rate approaches an empirical $L^{-1}$ rate.
}
  \label{fig:width-depth-rate}
\end{figure}

\subsection{Width--Depth Convergence Rate}
We next characterize how the finite-size error scales with width and depth at different training iterations. Since the limiting output is
not available in closed form, we construct an empirical reference
$\widehat f_\infty^{(k)}$ by averaging the outputs at the largest joint scale
$m=3072$ over the five width--depth scaling paths and five seeds. We measure the RMS output error over
the $N=128$ samples,
\[
\operatorname{Err}_k(n,L)
=
\left(
\frac{1}{N}\sum_{i=1}^{N}
\bigl(
f_{n,L}^{(k)}(x_i)-\widehat f_\infty^{(k)}(x_i)
\bigr)^2
\right)^{1/2}.
\]

To quantify how the finite-size error depends on width and depth, we vary one
dimension while fixing the other at $12{,}288$.
\cref{fig:width-depth-rate}(a)--(b) shows the sweeps at
$k\in\{1,5,10,15\}$. The width error decreases steadily with $n$, whereas
the depth error drops rapidly at small $L$ and then plateaus as finite-width
effects become dominant.

Let $n_{\rm ref}=L_{\rm ref}=12{,}288$, and let
$\operatorname{Err}^{\rm floor}_k$ denote the plateau level of the depth
sweep at iteration $k$. To quantify the convergence rates, we fit
\[
\operatorname{Err}_k(n,L_{\rm ref})\propto n^{-\alpha_n(k)},
\qquad
\operatorname{Err}_k(n_{\rm ref},L)
\propto L^{-\alpha_L(k)},
\]
where the depth exponent is estimated over the decreasing regime of the RMS
error curve.
\cref{fig:width-depth-rate}(c) shows the fitted exponents over training. The width rate remains close to $n^{-1/2}$ throughout training, whereas the depth-dependent error decays more rapidly and approaches an empirical $L^{-1}$ after the initial iterations.

\begin{remark}[Depth rate]
    The $L^{-1/2}$ rate in our \cref{thm:finiteT-sharp-main} is an upper bound induced by the
discrete-to-continuous estimates in the proof, and is not claimed to be
tight. The faster
empirical depth rate suggests that the finite-depth bound may be sharpened.
\end{remark}

\subsection{Nonlinear Activations}
\label{sec:exp-nonlinear}

Our theoretical analysis considers linear residual networks. We next test whether the same width--depth convergence behavior persists
with nonlinear residual blocks,
\[
Y_\ell
=
Y_{\ell-1}
+
\frac{1}{\sqrt L}
W_\ell\phi(Y_{\ell-1}),
\]
where \(\phi\) is applied elementwise. We consider ReLU, tanh, GELU, SiLU,
and sigmoid activations, train for \(10\) updates, and evaluate
\(m\in\{32,48,64,96,128,256\}\). All other experimental settings are the
same as in \cref{sec:exp-joint}.

\begin{figure}[t]
  \centering
  \includegraphics[width=\textwidth]
  {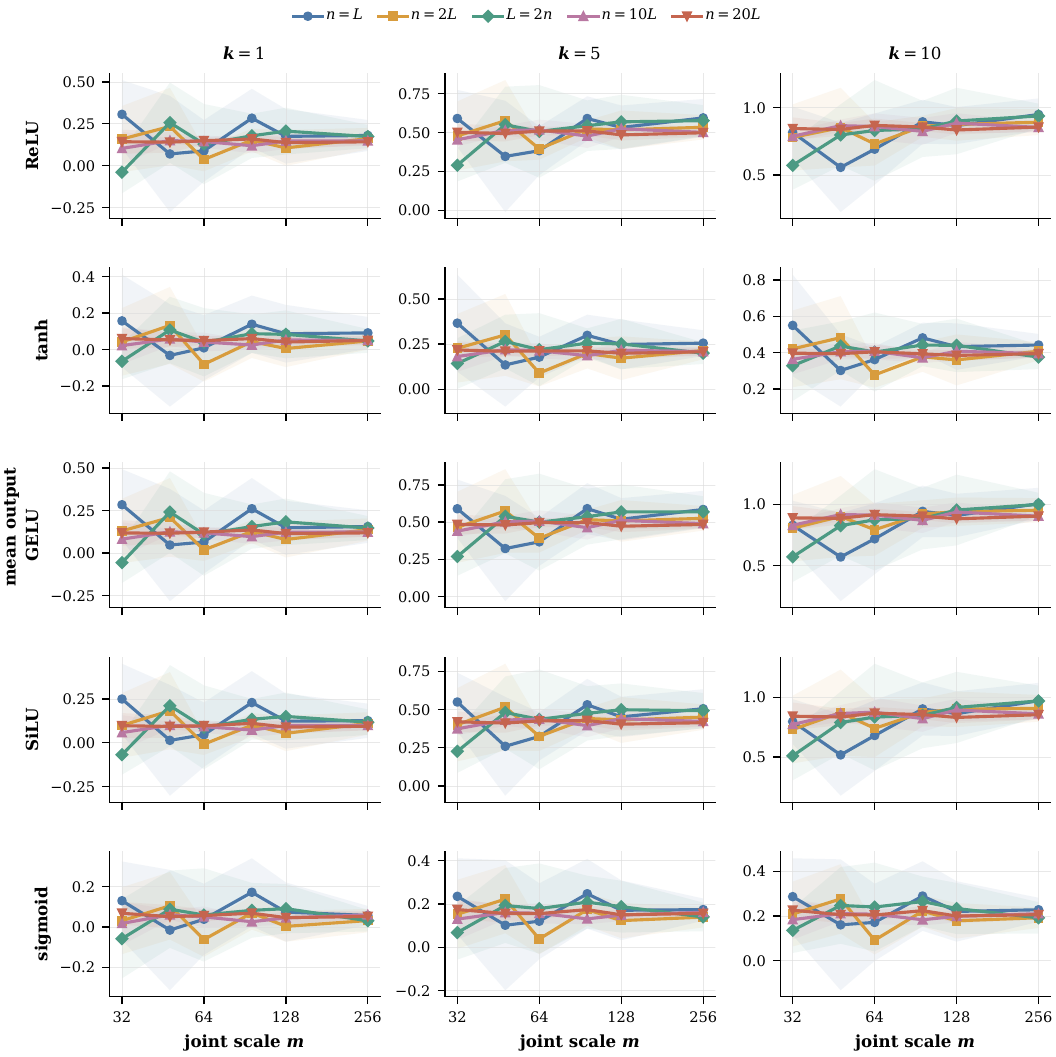}
  \caption{
  \textbf{Width--depth commutativity with nonlinear activations.}
  Rows correspond to activation functions and columns to training iterations.
  Each panel shows the output at synthetic training input \(86\) along the
  five width--depth scaling paths in \cref{sec:exp-joint}. Curves show the
  mean over five model seeds, with shaded regions denoting \(\pm1\) standard
  deviation.
  }
  \label{fig:nonlinear-activation-joint}
\end{figure}

As shown in \cref{fig:nonlinear-activation-joint}, the outputs along the different width--depth paths converge as the scale grows
for all five activations. This behavior persists across training iterations,
providing empirical evidence that the observed commutativity is not specific
to the linear architecture. However, our current theoretical guarantees remain restricted to linear residual networks.

\section{Conclusion}
We studied whether width and depth limits continue to commute during training
in linear residual networks. Under ordinary gradient descent, the limits
commute in probability after any fixed number of training iterations, while
the depth-first $\Ltwo$ limit already fails after one step at every fixed
finite width. Under clipped gradient descent, we prove commutativity in
$\Ltwo$ for any fixed number of training iterations, with errors of order
$n^{-1/2}$ and $L^{-1/2}$.

\subsection*{AI use statement}
We used OpenAI Codex to assist with proof auditing, consistency checking, manuscript revision, and \LaTeX{} formatting.  We reviewed
all AI-assisted work and take responsibility for the final content.

\subsection*{Reproducibility statement}
All theoretical results are proved in the appendices, with the model,
initialization, training dynamics, and limiting objects specified in the main
text. The empirical results use synthetic data and the publicly available
HIGGS dataset. Experimental settings, hyperparameters, and implementation
details are provided in \cref{app:exp}.
\bibliographystyle{unsrtnat}
\bibliography{references}

\appendix
\crefalias{section}{appendix}
\crefalias{subsection}{appendix}
\crefalias{subsubsection}{appendix}
\section{Proof of the \texorpdfstring{$\Ltwo$}{L2} Divergence at
\texorpdfstring{$k=1$}{k=1}}
\label{app:gd-depth-divergence}

This section proves  \cref{prop:gd-depth-divergence}.

\begin{proof}
    At initialization, write
$A_\ell^{(0)}=I+L^{-1/2}W_\ell$,
$Y_\ell:=Y_\ell^{(0)}$,
$B_\ell:=B_\ell^{(0)}$, and
$f_{n,L}:=f_{n,L}^{(0)}$.

Set $u_n:=1/(4n)$ and $b_n:=1+u_n$. Define $E_{n,L}$ as the event on which
\begin{enumerate}[label=(\roman*)]
\item $1\leq (Y_0)_i\leq 2$ and $1\leq \xi_i\leq 2$ for every $i$;
\item $u_n\sqrt L\leq (W_\ell)_{11}\leq 2u_n\sqrt L$ for every
      $\ell=1,\ldots,L$;
\item $0\leq (W_\ell)_{ij}\leq 1$ for every $\ell$ and
      $(i,j)\neq(1,1)$.
\end{enumerate}
Since $q_0>0$, condition~(i) has strictly positive probability.
On $E_{n,L}$, every entry of $A_\ell^{(0)}$ is nonnegative and
$(A_\ell^{(0)})_{11}\geq b_n$ for all $\ell$.

Hence, the forward recursion gives
\begin{equation}
(Y_\ell)_1\geq b_n^\ell,
\qquad
(Y_\ell)_i\geq 1,
\label{eq:rare-forward}
\end{equation}
for every $\ell$ and $i$. Similarly, the backward recursion gives
\begin{equation}
(B_\ell)_1\geq b_n^{L-\ell},
\qquad
(B_\ell)_i\geq 1.
\label{eq:rare-backward}
\end{equation}
Therefore,
\[
f_{n,L}
=
\frac{1}{n}\xi^\top Y_L
\geq
\frac{b_n^L}{n}.
\]
Since $b_n>1$, for all sufficiently large $L$,
\begin{equation}
\varepsilon_{n,L}^{(0)}
=
f_{n,L}-y_\star
\geq
\frac{b_n^L}{2n}
>0.
\label{eq:rare-residual}
\end{equation}

The first GD update is $(A_\ell^{(1)})_{ij}
=
(A_\ell^{(0)})_{ij}
-
\frac{\eta\varepsilon_{n,L}^{(0)}}{nL}
(B_\ell)_i(Y_{\ell-1})_j.$

On $E_{n,L}$, 
\[
\frac{\eta\varepsilon_{n,L}^{(0)}}{nL}
(B_\ell)_i(Y_{\ell-1})_j
\geq
\frac{\eta b_n^L}{2n^2L}
\quad\text{for all }i,j,
\]
while the first coordinate satisfies the stronger bound
\[
\frac{\eta\varepsilon_{n,L}^{(0)}}{nL}
(B_\ell)_1(Y_{\ell-1})_1
\geq
\frac{\eta b_n^{2L-1}}{2n^2L}.
\]

Since $\eta b_n^L/(2n^2L)\to\infty$, all entries of
$A_\ell^{(1)}$ are negative for sufficiently large $L$. Moreover, the
stronger bound on the first coordinate gives
\begin{equation}
\bigl|(A_\ell^{(1)})_{11}\bigr|
\geq
d_{n,L}
:=
\frac{\eta b_n^{2L-1}}{4n^2L},
\qquad
\ell=1,\ldots,L.
\label{eq:rare-dnl}
\end{equation}

Set $P_\ell:=-A_\ell^{(1)}$. On $E_{n,L}$, every entry of $P_\ell$ is
positive. Therefore,
\begin{equation}
\bigl|f_{n,L}^{(1)}\bigr|
=
\frac{1}{n}\xi^\top P_L\cdots P_1Y_0
\geq
\frac{1}{n}d_{n,L}^L .
\label{eq:rare-output-lower}
\end{equation}

It remains to estimate the probability of $E_{n,L}$. Define
\[
p_{n,L}
:=
\mathbb P\!\left(
u_n\sqrt L
\leq
(W_\ell)_{11}
\leq
2u_n\sqrt L
\right),
\qquad
q_n
:=
\mathbb P\!\left(
0\leq (W_\ell)_{ij}\leq 1
\right),
\]
where $(i,j)\neq(1,1)$. These probabilities do not depend on $\ell$,
since all entries of $W_\ell$ are identically distributed. Also, let
\[
r_n
:=
\mathbb P\!\left(
1\leq (Y_0)_i\leq 2,\;
1\leq \xi_i\leq 2
\text{ for all }i
\right)>0.
\]
By independence,
\begin{equation}
\mathbb P(E_{n,L})
=
r_n\,p_{n,L}^{\,L}\,
q_n^{(n^2-1)L}.
\label{eq:rare-event-prob}
\end{equation}

Since $(W_\ell)_{11}\sim\mathcal N(0,n^{-1})$, its density is
$\phi_n(w)=\sqrt{n/(2\pi)}\exp(-nw^2/2)$, which is decreasing on
$[0,\infty)$. Therefore,
\begin{equation}
p_{n,L}
\geq
u_n\sqrt L\sqrt{\frac{n}{2\pi}}
\exp(-2nu_n^2L).
\label{eq:rare-p-lower}
\end{equation}
Combining \eqref{eq:rare-event-prob} and \eqref{eq:rare-p-lower} gives
\begin{equation}
\log \mathbb P(E_{n,L})
\geq
-2nu_n^2L^2
+\frac{1}{2}L\log L
+c_nL
+\log r_n,
\label{eq:rare-log-prob}
\end{equation}
where $c_n:=(n^2-1)\log q_n+\log u_n
+\frac{1}{2}\log(n/(2\pi))$ depends only on $n$.

Using \eqref{eq:rare-output-lower},
\[
\mathbb E\bigl|f_{n,L}^{(1)}\bigr|^2
\geq
\mathbb P(E_{n,L})\,
\frac{d_{n,L}^{2L}}{n^2}.
\]
Together with \eqref{eq:rare-dnl} and \eqref{eq:rare-log-prob}, this yields
\begin{equation}
\log \mathbb E\bigl|f_{n,L}^{(1)}\bigr|^2
\geq
\kappa_n L^2
-\frac{3}{2}L\log L
+O_{n,\eta}(L),
\label{eq:rare-second-moment-lower}
\end{equation}
where $\kappa_n:=4\log b_n-2nu_n^2$.

Since $u_n=1/(4n)\leq1$ and
$\log(1+x)\geq x/2$ for $x\in[0,1]$,
\[
\kappa_n
=
4\log(1+u_n)-2nu_n^2
\geq
2u_n-2nu_n^2
=
\frac{3}{8n}
>0.
\]
Dividing \eqref{eq:rare-second-moment-lower} by $L^2$ and letting
$L\to\infty$ gives
\[
\mathbb E\bigl|f_{n,L}^{(1)}\bigr|^2
\longrightarrow
+\infty.
\]
This proves \cref{prop:gd-depth-divergence}.
\end{proof}

\section{Depth Discretization}
\label{app:depth-proof}
We prove \cref{prop:main-depth} by comparing the trained
residual network with the continuous-depth finite-width model. We first
compare the discrete and continuous representations
\(Y_{n,L}^{(k),\tau}\) and \(Y_n^{(k),\tau}\). The corresponding estimate
for \(B_{n,L}^{(k),\tau}\) and \(B_n^{(k),\tau}\) follows by depth reversal
and transposition; see \cref{app:depth-trained}. Here, \(F_n^{(k),\tau}\) and \(\mathcal J_n^{(k),\tau}\) denote the
continuous-depth finite-width limits of \(f_{n,L}^{(k),\tau}\) and
\(J_{n,L}^{(k),\tau}\), respectively. Moreover,
\(V_{n,L}^{(k)}\) denotes the cumulative parameter update after \(k\)
training steps for the finite-depth network, while \(V_n^{(k)}\) denotes
its continuous-depth limit. Their definitions are given in
\cref{app:depth-trained}.

The depth comparison follows the chain
\[
S_{i+1:j}^{n,L}
\xrightarrow{\,L^{-1/2}\,}
\Phi_{t_i,t_j},
\]
followed by the finite-training induction
\[
\begin{array}{c}
V_{n,L}^{(k)}
\xrightarrow{\,L^{-1/2}\,}
V_n^{(k)}
\;\Longrightarrow\;
\bigl(Y_{n,L}^{(k),\tau},B_{n,L}^{(k),\tau}\bigr)
\xrightarrow{\,L^{-1/2}\,}
\bigl(Y_n^{(k),\tau},B_n^{(k),\tau}\bigr)
\\[1.5mm]
\Longrightarrow\;
\bigl(f_{n,L}^{(k),\tau},J_{n,L}^{(k),\tau}\bigr)
\xrightarrow{\,L^{-1/2}\,}
\bigl(F_n^{(k),\tau},\mathcal J_n^{(k),\tau}\bigr)
\;\Longrightarrow\;
V_{n,L}^{(k+1)}
\xrightarrow{\,L^{-1/2}\,}
V_n^{(k+1)}.
\end{array}
\]

\subsection{Width-uniform base approximation}
\label{app:depth-euler}

Let
\[
t_i:=\frac{i}{L},
\qquad
S_{i+1:j}^{n,L}:=A_j^{(0)}\cdots A_{i+1}^{(0)},
\qquad 0\le i\le j\le L,
\]
with $S_{j+1:j}^{n,L}=I$. Couple the Gaussian residual matrices with an
$n\times n$ matrix Brownian motion $M$ through
\[
M_{t_j}-M_{t_{j-1}}=L^{-1/2}W_j,
\]
and let $\Phi_{s,t}$ solve
\[
\dd\Phi_{s,t}=\dd M_t\,\Phi_{s,t},
\qquad
\Phi_{s,s}=I.
\]

\begin{lemma}[Maximal inequality for matrix-Brownian integrals]
\label{lem:matrix-brownian-maximal}
Let
$M_t=n^{-1/2}(\beta_{\alpha\beta}(t))_{\alpha,\beta=1}^n$, where the
$\beta_{\alpha\beta}$ are independent standard Brownian motions. Let
$\psi$ be a predictable $\mathbb R^n$-valued process satisfying
$\int_a^b\|\psi_s\|_{L^r(\Omega;\mathbb R^n)}^2\,\dd s<\infty$
for some $r\ge2$, and set
$Z_t:=\int_a^t\dd M_s\,\psi_s$ for $a\le t\le b$. Then
\begin{equation}
\left\|
\sup_{a\le u\le t}\|Z_u\|_2
\right\|_{L^r}
\le
K_{r,n}
\left(
\int_a^t
\|\psi_s\|_{L^r(\Omega;\mathbb R^n)}^2\,\dd s
\right)^{1/2},
\qquad
K_{r,n}:=
\frac{r}{r-1}
\sqrt{1+\frac{r-2}{n}}.
\label{eq:matrix-brownian-maximal}
\end{equation}
\end{lemma}

\begin{proof}
Writing $Z_t=(Z_1(t),\ldots,Z_n(t))^\top$, we have
$\dd Z_\alpha(t)=n^{-1/2}\sum_{\beta=1}^n
\psi_\beta(t)\,\dd\beta_{\alpha\beta}(t)$ and hence
$\dd\langle Z_\alpha,Z_\gamma\rangle_t
=\delta_{\alpha\gamma}n^{-1}\|\psi_t\|_2^2\,\dd t$.
It\^o's formula applied to $\|Z_t\|_2^r$ gives
\begin{equation}
\frac{\dd}{\dd t}\mathbb E\|Z_t\|_2^r
=
\frac{r(r+n-2)}{2n}
\mathbb E\!\left[
\|Z_t\|_2^{r-2}\|\psi_t\|_2^2
\right].
\label{eq:matrix-brownian-ito}
\end{equation}
By H\"older's inequality,
\[
\mathbb E\!\left[
\|Z_t\|_2^{r-2}\|\psi_t\|_2^2
\right]
\le
\bigl(\mathbb E\|Z_t\|_2^r\bigr)^{(r-2)/r}
\|\psi_t\|_{L^r(\Omega;\mathbb R^n)}^2.
\]
Substituting this into \eqref{eq:matrix-brownian-ito} yields
\[
\frac{\dd}{\dd t}
\bigl(\mathbb E\|Z_t\|_2^r\bigr)^{2/r}
\le
\frac{r+n-2}{n}
\|\psi_t\|_{L^r(\Omega;\mathbb R^n)}^2.
\]
Since $Z_a=0$, integration gives
\begin{equation}
\|Z_t\|_{L^r(\Omega;\mathbb R^n)}
\le
\sqrt{1+\frac{r-2}{n}}
\left(
\int_a^t
\|\psi_s\|_{L^r(\Omega;\mathbb R^n)}^2\,\dd s
\right)^{1/2}.
\label{eq:matrix-brownian-terminal}
\end{equation}
Since $\|Z_t\|_2$ is a nonnegative submartingale, Doob's inequality and
\eqref{eq:matrix-brownian-terminal} give
\eqref{eq:matrix-brownian-maximal}.
\end{proof}

\begin{proposition}[Width-uniform discrete-to-continuous comparison]
\label{prop:depth-euler}
For every $p\ge2$, there are finite constants
$C_{\rm Eul}(p)$, $C_{\rm base}(p)$, and $C_{\rm reg}(p)$, independent
of $n$ and $L$, such that
\begin{align}
\sup_{n\ge1}\max_{0\le i\le j\le L}
\left\|
\left\|S_{i+1:j}^{n,L}-\Phi_{t_i,t_j}\right\|_{\op}
\right\|_{L^p}
&\le C_{\rm Eul}(p)L^{-1/2},
\label{eq:depth-euler-clean}\\
\sup_{n\ge1}\sup_{0\le s\le1}
\left\|
\sup_{s\le t\le1}\|\Phi_{s,t}\|_{\op}
\right\|_{L^p}
+
\sup_{n,L}\max_{0\le i\le L}
\left\|
\max_{i\le j\le L}\|S_{i+1:j}^{n,L}\|_{\op}
\right\|_{L^p}
&\le C_{\rm base}(p).
\label{eq:depth-base-max-clean}
\end{align}
Moreover, for every $0\le a\le b\le1$,
\begin{equation}
\sup_{n\ge1}
\left\|
\sup_{a\le s\le b}
\bigl(
\|\Phi_{a,s}-I\|_{\op}
+
\|\Phi_{s,b}-I\|_{\op}
\bigr)
\right\|_{L^p}
\le
C_{\rm reg}(p)\sqrt{b-a}.
\label{eq:depth-flow-regularity-clean}
\end{equation}
\end{proposition}

\begin{proof}
Fix $0\le i\le L$ and a deterministic Euclidean unit vector $v$, and set
$U_t:=\Phi_{t_i,t}v$ for $t\in[t_i,1]$. Writing
$M_t=n^{-1/2}(\beta_{\alpha\beta}(t))_{\alpha,\beta=1}^n$, where the
$\beta_{\alpha\beta}$ are independent standard Brownian motions, $U$ solves
$\dd U_t=\dd M_t\,U_t$ with $U_{t_i}=v$.

We first check the assumptions of the standard strong
Euler--Maruyama estimate. For $u,w\in\mathbb R^n$ and $\delta>0$,
independence of the Brownian entries gives
\[
\mathbb E
\left[
\|(M_{t+\delta}-M_t)(u-w)\|_2^2
\right]
=
\delta\|u-w\|_2^2.
\]
Thus the diffusion coefficient is globally Lipschitz with constant one,
uniformly in $n$. The same identity with $u-w$ replaced by $u$ gives
the linear-growth condition, again with a constant independent of $n$.

Under the coupling above, the Euler--Maruyama recursion on
$\{t_j\}_{j=i}^L$ is
\[
U_{t_j}^L
=
\bigl(I+L^{-1/2}W_j\bigr)U_{t_{j-1}}^L,
\qquad
U_{t_i}^L=v,
\]
so $U_{t_j}^L=S_{i+1:j}^{n,L}v$. Therefore, for every fixed $n$ and
$p\ge2$, the standard strong Euler--Maruyama theorem gives
\[
\max_{i\le j\le L}
\left\|
S_{i+1:j}^{n,L}v-\Phi_{t_i,t_j}v
\right\|_{L^p}
\le
C_{n,p}L^{-1/2}.
\]
This estimate controls one deterministic direction, whereas
\eqref{eq:depth-euler-clean} requires an operator-norm bound with a
constant uniform in $n$. We therefore track the dependence on both
the moment order and the width.

For every fixed $n$ and $r\ge2$, the standard moment estimate for
globally Lipschitz SDEs gives
$\int_{t_i}^1\|U_s\|_{L^r}^2\,\dd s<\infty$. Taking $\psi_s=U_s$ in
\cref{lem:matrix-brownian-maximal} and using
$U_t=v+\int_{t_i}^t\dd M_s\,U_s$, set
$F(t):=\|\sup_{t_i\le u\le t}\|U_u\|_2\|_{L^r}$.
Then
\[
F(t)^2
\le
2+2K_{r,n}^2\int_{t_i}^tF(s)^2\,\dd s.
\]
Gronwall's inequality gives
\begin{equation}
\left\|
\sup_{t_i\le t\le1}\|U_t\|_2
\right\|_{L^r}
\le
\sqrt2\,e^{K_{r,n}^2}.
\label{eq:depth-vector-moment-clean}
\end{equation}

We next control the variation of $U$ within one mesh interval. For
$i<\ell\le L$ and $s\in[t_{\ell-1},t_\ell]$,
\[
U_s-U_{t_{\ell-1}}
=
\int_{t_{\ell-1}}^s\dd M_u\,U_u.
\]
\cref{lem:matrix-brownian-maximal} on
$[t_{\ell-1},t_\ell]$, together with
\eqref{eq:depth-vector-moment-clean}, gives
\begin{equation}
\left\|
\sup_{t_{\ell-1}\le s\le t_\ell}
\|U_s-U_{t_{\ell-1}}\|_2
\right\|_{L^r}
\le
\sqrt2\,K_{r,n}e^{K_{r,n}^2}L^{-1/2}.
\label{eq:depth-cell-increment-clean}
\end{equation}

Let $U^L$ denote the continuous Euler interpolation,
\[
U_t^L
=
v+\int_{t_i}^t\dd M_s\,U_{\pi_L(s)}^L,
\qquad
\pi_L(s)=t_{\ell-1}
\quad\text{for }s\in(t_{\ell-1},t_\ell].
\]
At the grid points, $U_{t_j}^L=S_{i+1:j}^{n,L}v$, and
\[
U_t^L-U_t
=
\int_{t_i}^t\dd M_s
\left(
U_{\pi_L(s)}^L-U_{\pi_L(s)}
+
U_{\pi_L(s)}-U_s
\right).
\]

The Euler interpolation $U^L$ has finite $r$-th moments for every fixed
$n$, $L$, and $r\ge2$. Setting
$E(t):=\|\sup_{t_i\le u\le t}\|U_u^L-U_u\|_2\|_{L^r}$,
\cref{lem:matrix-brownian-maximal} and
\eqref{eq:depth-cell-increment-clean} give
\[
E(t)^2
\le
2K_{r,n}^2\int_{t_i}^tE(s)^2\,\dd s
+
4K_{r,n}^4e^{2K_{r,n}^2}L^{-1}.
\]
Gronwall's inequality therefore gives
\begin{equation}
\left\|
\sup_{t_i\le t\le1}
\|U_t^L-U_t\|_2
\right\|_{L^r}
\le
2K_{r,n}^2e^{2K_{r,n}^2}L^{-1/2}.
\label{eq:depth-vector-euler-clean}
\end{equation}

To obtain the operator-norm estimate uniformly in $n$, let $\mathcal N$
be a deterministic $1/2$-net of the Euclidean unit sphere with
$|\mathcal N|\le5^n$, and set
$r=2\lceil\max\{n,p,4\}/2\rceil$. Then $r\ge n,p$,
$5^{n/r}\le5$, and
\begin{equation}
K_{r,n}^2
=
\left(\frac r{r-1}\right)^2
\left(1+\frac{r-2}{n}\right)
\le
2(p+3).
\label{eq:depth-moment-ratio-clean}
\end{equation}

The $1/2$-net inequality gives
\[
\left\|
S_{i+1:j}^{n,L}-\Phi_{t_i,t_j}
\right\|_{\op}
\le
2\max_{v\in\mathcal N}
\left\|
\bigl(S_{i+1:j}^{n,L}-\Phi_{t_i,t_j}\bigr)v
\right\|_2.
\]
Hence
\[
\left\|
\left\|
S_{i+1:j}^{n,L}-\Phi_{t_i,t_j}
\right\|_{\op}
\right\|_{L^p}
\le
2|\mathcal N|^{1/r}
\max_{v\in\mathcal N}
\left\|
\left\|
\bigl(S_{i+1:j}^{n,L}-\Phi_{t_i,t_j}\bigr)v
\right\|_2
\right\|_{L^r}.
\]
Since $r\ge n$, the factor $|\mathcal N|^{1/r}$ is bounded uniformly
in $n$, while \eqref{eq:depth-moment-ratio-clean} gives a bound on
$K_{r,n}$ that is also independent of $n$.  Applying
\eqref{eq:depth-vector-euler-clean}, uniformly over the starting index
$i$, proves \eqref{eq:depth-euler-clean}.

Applying the same net estimate to
\eqref{eq:depth-vector-moment-clean}, and starting the same SDE at an
arbitrary deterministic $s\in[0,1]$, gives
\[
\sup_{n\ge1}\sup_{0\le s\le1}
\left\|
\sup_{s\le t\le1}
\|\Phi_{s,t}\|_{\op}
\right\|_{L^p}
\le
10\sqrt2\,e^{2(p+3)}.
\]
Moreover, since
$\|U_t^L\|_2\le\|U_t\|_2+\|U_t^L-U_t\|_2$,
\eqref{eq:depth-vector-moment-clean} and
\eqref{eq:depth-vector-euler-clean}, followed by the same net estimate,
give
\[
\sup_{n,L}\max_{0\le i\le L}
\left\|
\max_{i\le j\le L}
\|S_{i+1:j}^{n,L}\|_{\op}
\right\|_{L^p}
\le
10\sqrt2\,e^{2(p+3)}
+
40(p+3)e^{4(p+3)}.
\]
Thus, \eqref{eq:depth-base-max-clean} holds with
$C_{\rm base}(p):=
20\sqrt2\,e^{2(p+3)}+40(p+3)e^{4(p+3)}$.

We also need the variation of the base flow over a short depth interval.
Fix $0\le a\le b\le1$.  For every deterministic unit vector $v$,
\[
(\Phi_{a,s}-I)v
=
\int_a^s\dd M_u\,\Phi_{a,u}v,
\qquad a\le s\le b.
\]
Applying \cref{lem:matrix-brownian-maximal} on $[a,b]$, using
\eqref{eq:depth-vector-moment-clean}, and then applying the same net
estimate gives
\[
\sup_{n\ge1}
\left\|
\sup_{a\le s\le b}
\|\Phi_{a,s}-I\|_{\op}
\right\|_{L^p}
\le
C_{\rm reg}(p)\sqrt{b-a}.
\]

It remains to control the flow when the right endpoint is fixed.
For fixed $b$, define
\[
\check M_u:=(M_b-M_{b-u})^\top,
\qquad 0\le u\le b-a.
\]
Then $\check M$ has the same law as $M$, and its forward flow satisfies $\check\Phi_{0,u}=\Phi_{b-u,b}^\top.$

Applying the preceding estimate to $\check\Phi$ and using $\|\Phi_{b-u,b}^\top-I\|_{\op}
=
\|\Phi_{b-u,b}-I\|_{\op}$
gives
\[
\sup_{n\ge1}
\left\|
\sup_{a\le s\le b}
\|\Phi_{s,b}-I\|_{\op}
\right\|_{L^p}
\le
C_{\rm reg}(p)\sqrt{b-a}.
\]
Combining the two estimates proves
\eqref{eq:depth-flow-regularity-clean}.
\end{proof}

\subsection{Finite-training depth comparison}
\label{app:depth-trained}
We now show that the $L^{-1/2}$ rate is preserved through every fixed
training step.

We first define the finite-width continuous-depth training dynamics.
Set $V_n^{(0)}(t):=0$. Given $V_n^{(k)}$, let the forward and backward
states satisfy
\begin{align}
Y_n^{(k),\tau}(t)
&=
\Phi_{0,t}Y_0
-
\int_0^t
\Phi_{s,t}V_n^{(k)}(s)Y_n^{(k),\tau}(s)\,\dd s,
\label{eq:continuous-forward}\\
B_n^{(k),\tau}(t)
&=
\Phi_{t,1}^{\top}\xi
-
\int_t^1
\Phi_{t,s}^{\top}V_n^{(k)}(s)^\top
B_n^{(k),\tau}(s)\,\dd s.
\label{eq:continuous-backward}
\end{align}
Let $F_n^{(k),\tau}:=\langle\xi,Y_n^{(k),\tau}(1)\rangle_n$ and
\[
\mathcal J_n^{(k),\tau}
:=
\int_0^1
\|Y_n^{(k),\tau}(t)\|_n^2
\|B_n^{(k),\tau}(t)\|_n^2\,\dd t.
\]
Define $\gamma_n^{(k),\tau}
:=
\eta\bigl(F_n^{(k),\tau}-y_\star\bigr)
\min\left\{
1,
\frac{\tau}{
|F_n^{(k),\tau}-y_\star|
\sqrt{\mathcal J_n^{(k),\tau}}}
\right\},$
where the multiplier inside the minimum is set to one if the denominator
vanishes. For $k<T$, set
\[
V_n^{(k+1)}(t)
=
V_n^{(k)}(t)
+
\frac{\gamma_n^{(k),\tau}}{n}
B_n^{(k),\tau}(t)
\bigl(Y_n^{(k),\tau}(t)\bigr)^\top.
\]

For $t\in I_\ell:=(t_{\ell-1},t_\ell]$, define the discrete cumulative
update by
\[
V_{n,L}^{(k)}(t)
=
\sum_{m<k}
\frac{\gamma_{n,L}^{(m),\tau}}{n}
B_{n,L}^{(m),\tau}(t_\ell)
\bigl(Y_{n,L}^{(m),\tau}(t_{\ell-1})\bigr)^\top .
\]
Iterating the continuous update gives
\[
V_n^{(k)}(t)
=
\sum_{m<k}
\frac{\gamma_n^{(m),\tau}}{n}
B_n^{(m),\tau}(t)
\bigl(Y_n^{(m),\tau}(t)\bigr)^\top .
\]

For the cumulative updates above, we write $\|V\|_{L_t^2(\mathrm F)}
:=
\left(
\int_0^1\|V(t)\|_{\mathrm F}^2\,\dd t
\right)^{1/2}.$

By the rank-one structure of the updates and the definitions of
$J_{n,L}^{(m),\tau}$ and $\mathcal J_n^{(m),\tau}$,
\begin{align}
\left(
\frac1L\sum_{\ell=1}^L
\left\|
\frac1n
B_{n,L}^{(m),\tau}(t_\ell)
\bigl(Y_{n,L}^{(m),\tau}(t_{\ell-1})\bigr)^\top
\right\|_{\mathrm F}^2
\right)^{1/2}
&=
\sqrt{J_{n,L}^{(m),\tau}}, \notag
\\
\left(
\int_0^1
\left\|
\frac1n
B_n^{(m),\tau}(t)
\bigl(Y_n^{(m),\tau}(t)\bigr)^\top
\right\|_{\mathrm F}^2\,\dd t
\right)^{1/2}
&=
\sqrt{\mathcal J_n^{(m),\tau}}. \notag
\end{align}

By the definition of the clipped coefficients,
\[
|\gamma_{n,L}^{(m),\tau}|
\sqrt{J_{n,L}^{(m),\tau}}
\le\eta\tau,
\qquad
|\gamma_n^{(m),\tau}|
\sqrt{\mathcal J_n^{(m),\tau}}
\le\eta\tau.
\]
Summing over $m<k$ gives
\begin{equation}
\|V_{n,L}^{(k)}\|_{L_t^2(\mathrm F)}
\vee
\|V_n^{(k)}\|_{L_t^2(\mathrm F)}
\le k\eta\tau
\qquad\text{a.s.}
\label{eq:depth-update-budget-clean}
\end{equation}

After reversing the depth variable and transposing the matrices, the
backward equations take the same form as the forward equations. Indeed,
if $B_{\ell-1}=A_\ell^\top B_\ell$, then
\[
B_{L-i}
=
A_{L-i+1}^\top B_{L-i+1},
\qquad
1\le i\le L,
\]
which is a forward recursion in the reversed layer index. At
initialization,
\[
A_{L-i+1}^{(0)\top}
=
I+L^{-1/2}W_{L-i+1}^\top,
\]
and $(W_L^\top,\ldots,W_1^\top)$ has the same joint law as
$(W_1,\ldots,W_L)$.

For the continuous equation,
\[
B(t)
=
\Phi_{t,1}^\top\xi
-
\int_t^1
\Phi_{t,s}^\top V(s)^\top B(s)\,\dd s,
\]
the substitutions $u=1-t$ and $r=1-s$ give
\begin{equation}
B(1-u)
=
\Phi_{1-u,1}^\top\xi
-
\int_0^u
\Phi_{1-u,1-r}^\top
V(1-r)^\top
B(1-r)\,\dd r.
\label{eq:reversed-backward-clean}
\end{equation}
The process $u\mapsto(M_1-M_{1-u})^\top$ has the same law as $M$, and
its forward flow from $r$ to $u$ is $\Phi_{1-u,1-r}^\top$. Moreover,
\[
\int_0^1
\|V(1-r)^\top\|_{\mathrm F}^2\,\dd r
=
\int_0^1
\|V(s)\|_{\mathrm F}^2\,\dd s.
\]
Hence the reversed backward equation has the same form as the forward
equation, with the same Brownian law and the same
$L_t^2(\mathrm F)$ bound on the cumulative update. The estimates below
therefore also hold for the backward equation. We prove only the
forward estimate in detail.

\begin{proof}[Proof of \cref{prop:main-depth}]
At $k=0$, $V_{n,L}^{(0)}=V_n^{(0)}=0$. We prove by induction that, for
every finite $q\ge2$,
\begin{equation}
\sup_{n\ge1}
\left\|
\|V_{n,L}^{(k)}-V_n^{(k)}\|_{L_t^2(\mathrm F)}
\right\|_{L^q}
\le
C_V(k,q)L^{-1/2}.
\label{eq:depth-update-induction-clean}
\end{equation}
\cref{prop:training-bounds} and
\eqref{eq:depth-update-budget-clean} give the moment bounds for the
trained forward states used below.

For $s\in I_\ell$,
\[
Y_n^{(k),\tau}(s)-Y_n^{(k),\tau}(t_{\ell-1})
=
(\Phi_{t_{\ell-1},s}-I)
Y_n^{(k),\tau}(t_{\ell-1})
-
\int_{t_{\ell-1}}^s
\Phi_{u,s}V_n^{(k)}(u)Y_n^{(k),\tau}(u)\,\dd u.
\]
Using \eqref{eq:depth-flow-regularity-clean},
\eqref{eq:depth-update-budget-clean}, and the moment bounds from
\cref{prop:training-bounds}, Cauchy--Schwarz on each
$I_\ell$ gives
\begin{equation}
\sup_{n\ge1}
\left\|
\left(
\sum_{\ell=1}^L
\int_{I_\ell}
\|Y_n^{(k),\tau}(s)-Y_n^{(k),\tau}(t_{\ell-1})\|_n^2\,\dd s
\right)^{1/2}
\right\|_{L^q}
\le
C_{\rm cell}(k,q)L^{-1/2}.
\label{eq:depth-state-regularity-clean}
\end{equation}

Set $r:=4q$. At the grid points,
\begin{align*}
Y_{n,L}^{(k),\tau}(t_j)
&=
S_{1:j}^{n,L}Y_0
-
\frac1L
\sum_{\ell=1}^j
S_{\ell+1:j}^{n,L}
V_{n,L}^{(k)}(t_\ell)
Y_{n,L}^{(k),\tau}(t_{\ell-1}),\\
Y_n^{(k),\tau}(t_j)
&=
\Phi_{0,t_j}Y_0
-
\int_0^{t_j}
\Phi_{s,t_j}V_n^{(k)}(s)
Y_n^{(k),\tau}(s)\,\dd s.
\end{align*}
Since $V_{n,L}^{(k)}$ is constant on each $I_\ell$, subtracting the two
equations gives
\begin{align*}
&
S_{\ell+1:j}^{n,L}V_{n,L}^{(k)}(t_\ell)
Y_{n,L}^{(k),\tau}(t_{\ell-1})
-
\Phi_{s,t_j}V_n^{(k)}(s)Y_n^{(k),\tau}(s)
\\
={}&
\underbrace{
\bigl(S_{\ell+1:j}^{n,L}-\Phi_{s,t_j}\bigr)
V_{n,L}^{(k)}(t_\ell)
Y_{n,L}^{(k),\tau}(t_{\ell-1})
}_{\mathrm{(a)}}
+
\underbrace{
\Phi_{s,t_j}
\bigl(V_{n,L}^{(k)}(t_\ell)-V_n^{(k)}(s)\bigr)
Y_{n,L}^{(k),\tau}(t_{\ell-1})
}_{\mathrm{(b)}}
\\
&+
\underbrace{
\Phi_{s,t_j}V_n^{(k)}(s)
\bigl(
Y_{n,L}^{(k),\tau}(t_{\ell-1})
-
Y_n^{(k),\tau}(s)
\bigr)
}_{\mathrm{(c)}}.
\end{align*}

For (a),
\[
S_{\ell+1:j}^{n,L}-\Phi_{s,t_j}
=
\bigl(S_{\ell+1:j}^{n,L}-\Phi_{t_\ell,t_j}\bigr)
+
\Phi_{t_\ell,t_j}
\bigl(I-\Phi_{s,t_\ell}\bigr).
\]
\cref{prop:depth-euler},
\eqref{eq:depth-update-budget-clean}, and the moment bounds for the
forward states give
\[
\left\|
\max_{j\le L}
\left\|
\sum_{\ell=1}^j
\int_{I_\ell}\mathrm{(a)}\,\dd s
\right\|_n
\right\|_{L^{2r}}
\le
C_a(k,r)L^{-1/2}.
\]

For (b), Cauchy--Schwarz in depth gives
\begin{align*}
\max_{j\le L}
\left\|
\sum_{\ell=1}^j
\int_{I_\ell}\mathrm{(b)}\,\dd s
\right\|_n
\le{}&
\sup_{s\le t}\|\Phi_{s,t}\|_{\op}
\max_\ell
\|Y_{n,L}^{(k),\tau}(t_{\ell-1})\|_n\\
&\times
\|V_{n,L}^{(k)}-V_n^{(k)}\|_{L_t^2(\mathrm F)}.
\end{align*}
H\"older's inequality, \eqref{eq:depth-base-max-clean}, the state moment
bounds, and the induction hypothesis at moment $4r$ therefore give
\[
\left\|
\max_{j\le L}
\left\|
\sum_{\ell=1}^j
\int_{I_\ell}\mathrm{(b)}\,\dd s
\right\|_n
\right\|_{L^{2r}}
\le
C_b(k,r)L^{-1/2}.
\]

For (c), write
\[
Y_{n,L}^{(k),\tau}(t_{\ell-1})-Y_n^{(k),\tau}(s)
=
E_{\ell-1}^{(k)}
+
Y_n^{(k),\tau}(t_{\ell-1})-Y_n^{(k),\tau}(s),
\]
where $E_{\ell-1}^{(k)}
:=Y_{n,L}^{(k),\tau}(t_{\ell-1})
-Y_n^{(k),\tau}(t_{\ell-1})$.
The difference
$Y_n^{(k),\tau}(t_{\ell-1})-Y_n^{(k),\tau}(s)$ is controlled by
\eqref{eq:depth-state-regularity-clean}, while
$E_{\ell-1}^{(k)}$ remains in the linear equation for the current
grid-point difference. \cref{prop:depth-euler} also gives
an $L^{2r}$ bound of order $L^{-1/2}$ for
$(S_{1:j}^{n,L}-\Phi_{0,t_j})Y_0$.

Consequently,
\[
E_j^{(k)}
=
R_j^{(k)}
-
\sum_{\ell=1}^j
\int_{I_\ell}
\Phi_{s,t_j}V_n^{(k)}(s)E_{\ell-1}^{(k)}\,\dd s,
\]
where $E_j^{(k)}
:=Y_{n,L}^{(k),\tau}(t_j)-Y_n^{(k),\tau}(t_j)$ and
$\|\max_j\|R_j^{(k)}\|_n\|_{L^{2r}}
\le C_R(k,r)L^{-1/2}$.
\cref{prop:training-stability} with
$\Lambda=k\eta\tau$ therefore gives
\begin{equation}
\sup_{n\ge1}
\left\|
\max_{0\le j\le L}
\|Y_{n,L}^{(k),\tau}(t_j)-Y_n^{(k),\tau}(t_j)\|_n
\right\|_{L^r}
\le
C_Y(k,r)L^{-1/2}.
\label{eq:depth-forward-state-clean}
\end{equation}

The reversal and transposition argument above gives the corresponding
grid-point bound for the backward states and, for $s\in I_\ell$, the
bound for
$B_n^{(k),\tau}(s)-B_n^{(k),\tau}(t_\ell)$:
\begin{equation}
\sup_{n\ge1}
\left\|
\max_{0\le j\le L}
\|B_{n,L}^{(k),\tau}(t_j)-B_n^{(k),\tau}(t_j)\|_n
\right\|_{L^r}
\le
C_B(k,r)L^{-1/2}.
\label{eq:depth-backward-state-clean}
\end{equation}
and
\begin{equation}
\sup_{n\ge1}
\left\|
\left(
\sum_{\ell=1}^L
\int_{I_\ell}
\|B_n^{(k),\tau}(s)-B_n^{(k),\tau}(t_\ell)\|_n^2\,\dd s
\right)^{1/2}
\right\|_{L^r}
\le
C_{B,\rm reg}(k,r)L^{-1/2}.
\label{eq:depth-backward-regularity-clean}
\end{equation}

It remains to close the induction. The output identity
\[
f_{n,L}^{(k),\tau}-F_n^{(k),\tau}
=
\left\langle
\xi,
Y_{n,L}^{(k),\tau}(1)-Y_n^{(k),\tau}(1)
\right\rangle_n
\]
together with the uniform moment bound for $\|\xi\|_n$ and
\eqref{eq:depth-forward-state-clean} gives
\[
\sup_n
\|f_{n,L}^{(k),\tau}-F_n^{(k),\tau}\|_{L^q}
\le
C_f(k,q)L^{-1/2}.
\]

The forward and backward state bounds, together with
\eqref{eq:depth-state-regularity-clean} and the corresponding bound for
$B_n^{(k),\tau}$, give
\[
\sup_n
\|J_{n,L}^{(k),\tau}-\mathcal J_n^{(k),\tau}\|_{L^q}
\le
C_J(k,q)L^{-1/2},
\]
using
$|a^2b^2-c^2d^2|
\le(a+c)|a-c|b^2+c^2(b+d)|b-d|$.

If $k<T$, define
\[
G_{n,L}^{(k)}(t)
=
\frac{f_{n,L}^{(k),\tau}-y_\star}{n}
B_{n,L}^{(k),\tau}(t_\ell)
\bigl(Y_{n,L}^{(k),\tau}(t_{\ell-1})\bigr)^\top,
\qquad t\in I_\ell,
\]
and
\[
G_n^{(k)}(t)
=
\frac{F_n^{(k),\tau}-y_\star}{n}
B_n^{(k),\tau}(t)
\bigl(Y_n^{(k),\tau}(t)\bigr)^\top.
\]

Since
\[
\begin{aligned}
\left\|
\frac1n
B_{n,L}^{(k),\tau}(t_\ell)
\bigl(Y_{n,L}^{(k),\tau}(t_{\ell-1})\bigr)^\top
\right\|_{\mathrm F}
&=
\|B_{n,L}^{(k),\tau}(t_\ell)\|_n
\|Y_{n,L}^{(k),\tau}(t_{\ell-1})\|_n,
\\
\left\|
\frac1n
B_n^{(k),\tau}(t)
\bigl(Y_n^{(k),\tau}(t)\bigr)^\top
\right\|_{\mathrm F}
&=
\|B_n^{(k),\tau}(t)\|_n
\|Y_n^{(k),\tau}(t)\|_n,
\end{aligned}
\]
the preceding estimates give
\[
\sup_n
\left\|
\|G_{n,L}^{(k)}-G_n^{(k)}\|_{L_t^2(\mathrm F)}
\right\|_{L^q}
\le
C_G(k,q)L^{-1/2}.
\]

Define
$\Pi_\tau(H):=H\min\{1,\tau/\|H\|_{L_t^2(\mathrm F)}\}$,
with multiplier one when $H=0$. Since $\Pi_\tau$ is nonexpansive,
\[
\left\|
\Pi_\tau(G_{n,L}^{(k)})
-
\Pi_\tau(G_n^{(k)})
\right\|_{L_t^2(\mathrm F)}
\le
\left\|
G_{n,L}^{(k)}
-
G_n^{(k)}
\right\|_{L_t^2(\mathrm F)}.
\]

Since
\[
V_{n,L}^{(k+1)}
=
V_{n,L}^{(k)}
+
\eta\Pi_\tau(G_{n,L}^{(k)}),
\qquad
V_n^{(k+1)}
=
V_n^{(k)}
+
\eta\Pi_\tau(G_n^{(k)}),
\]
we obtain
\[
\|V_{n,L}^{(k+1)}-V_n^{(k+1)}\|_{L_t^2(\mathrm F)}
\le
\|V_{n,L}^{(k)}-V_n^{(k)}\|_{L_t^2(\mathrm F)}
+
\eta
\|G_{n,L}^{(k)}-G_n^{(k)}\|_{L_t^2(\mathrm F)}.
\]

The induction hypothesis and the bound on
$\|G_{n,L}^{(k)}-G_n^{(k)}\|_{L_t^2(\mathrm F)}$
therefore give \eqref{eq:depth-update-induction-clean} at $k+1$.

The induction therefore holds for every $k\le T$. Since the output estimate
above holds for every $k\le T$ and $T$ is fixed,
\[
\max_{0\le k\le T}
\|f_{n,L}^{(k),\tau}-F_n^{(k),\tau}\|_{L^p}
\le
C_{\rm D}(T,p)L^{-1/2}.
\]
This proves \cref{prop:main-depth}.
\end{proof}

\section{Width Limit of the Continuous-Depth Model}
\label{app:width-proof}
For
notational convenience, write
$Y_n^{(k)}:=Y_n^{(k),\tau}$,
$B_n^{(k)}:=B_n^{(k),\tau}$,
$F_n^{(k)}:=F_n^{(k),\tau}$, and
$\mathcal J_n^{(k)}:=\mathcal J_n^{(k),\tau}$. Let $y^{(k)}$, $b^{(k)}$, and $\gamma_k^\tau$ denote the deterministic
infinite-width limits defined in \cref{app:deterministic-limit}.
For $a,b\le T$, let
\[
Q_{ab,n}(t):=
\langle Y_n^{(a)}(t),Y_n^{(b)}(t)\rangle_n,
\qquad
R_{ab,n}(t):=
\langle B_n^{(a)}(t),B_n^{(b)}(t)\rangle_n,
\]
and
\[
Q_{ab}(t):=
\langle y^{(a)}(t),y^{(b)}(t)\rangle,
\qquad
R_{ab}(t):=
\langle b^{(a)}(t),b^{(b)}(t)\rangle.
\]
Write also $g_r:=\gamma_r^\tau$.

We prove \cref{prop:main-width} in this section. We introduce
surrogate states $\widetilde Y_n^{(k)}$ and $\widetilde B_n^{(k)}$ between
the finite-width states $Y_n^{(k)},B_n^{(k)}$ and the limiting states
$y^{(k)},b^{(k)}$. We first compare the surrogate quantities with their
limiting values and then compare the finite-width states with the surrogate
states. Backward estimates follow by depth reversal and transposition.

At the output level, the two comparisons meet through
\[
F_n^{(k)}-f_{\infty}^{(k),\tau}
=
\langle \xi,Y_n^{(k)}(1)-\widetilde Y_n^{(k)}(1)\rangle_n
+
\left(
\langle \xi,\widetilde Y_n^{(k)}(1)\rangle_n-f_{\infty}^{(k),\tau}
\right),
\]
where
\[
Y_n^{(k)}
\xrightarrow{\,n^{-1/2}\,}
\widetilde Y_n^{(k)},
\qquad
\langle \xi,\widetilde Y_n^{(k)}(1)\rangle_n
\xrightarrow{\,n^{-1/2}\,}
f_{\infty}^{(k),\tau}.
\]

\subsection{Surrogate construction}

Define
\begin{align}
\widetilde Y_n^{(k)}(t)
&=
\Phi_{0,t}W_0x
-
\sum_{r<k}g_r\int_0^t
\Phi_{s,t}\widetilde B_n^{(r)}(s)Q_{rk}(s)\,\dd s,
\label{eq:main-surrogate-forward-clean}\\
\widetilde B_n^{(k)}(t)
&=
\Phi_{t,1}^{\top}\xi
-
\sum_{r<k}g_r\int_t^1
\Phi_{t,s}^{\top}\widetilde Y_n^{(r)}(s)R_{rk}(s)\,\dd s.
\label{eq:main-surrogate-backward-clean}
\end{align}

In the Volterra system defining the surrogate states, the matrix-Brownian
flow $\Phi$, $W_0x$, and $\xi$ remain finite-width, while the coefficients
in the integral terms are taken from the limiting model.

\begin{proposition}[Surrogate-to-limit comparison]
\label{prop:surrogate-approximation-clean}
For every fixed $T$ and every even $p\ge2$, there exists
$C_{\rm sur}(T,p)<\infty$, independent of $n$, such that
\begin{equation}
\max_{0\le k\le T}
\left\|
\langle \xi,\widetilde Y_n^{(k)}(1)\rangle_n-f_{\infty}^{(k),\tau}
\right\|_{L^p}
\le
C_{\rm sur}(T,p)n^{-1/2}.
\label{eq:surrogate-output-clean}
\end{equation}
In addition, for all $0\le a,b\le T$,
\begin{align}
\sup_{t\in[0,1]}
\left\|
\langle\widetilde Y_n^{(a)}(t),\widetilde Y_n^{(b)}(t)\rangle_n
-Q_{ab}(t)
\right\|_{L^p}
&\le
C_{\rm sur}(T,p)n^{-1/2},
\label{eq:surrogate-forward-contraction-clean}\\
\sup_{t\in[0,1]}
\left\|
\langle\widetilde B_n^{(a)}(t),\widetilde B_n^{(b)}(t)\rangle_n
-R_{ab}(t)
\right\|_{L^p}
&\le
C_{\rm sur}(T,p)n^{-1/2}.
\label{eq:surrogate-backward-contraction-clean}
\end{align}
\end{proposition}

\begin{proof}
At $k=0$,
\[
\widetilde Y_n^{(0)}(t)=\Phi_{0,t}W_0x,
\qquad
\widetilde B_n^{(0)}(t)=\Phi_{t,1}^{\top}\xi.
\]
Now suppose that, for every $r<k$, there exist matrix-valued functions
$K_{r,1},K_{r,2},L_{r,1},L_{r,2}$ such that
\begin{equation}
\widetilde Y_n^{(r)}(t)
=
K_{r,1}(t)W_0x+K_{r,2}(t)\xi,
\qquad
\widetilde B_n^{(r)}(t)
=
L_{r,1}(t)W_0x+L_{r,2}(t)\xi.
\label{eq:surrogate-linear-representation}
\end{equation}
Substituting the second identity in
\eqref{eq:surrogate-linear-representation} into
\eqref{eq:main-surrogate-forward-clean} gives
\[
\begin{aligned}
\widetilde Y_n^{(k)}(t)
=
\left[
\Phi_{0,t}
-
\sum_{r<k}g_r
\int_0^t
\Phi_{s,t}L_{r,1}(s)Q_{rk}(s)\,\dd s
\right]W_0x-
\left[
\sum_{r<k}g_r
\int_0^t
\Phi_{s,t}L_{r,2}(s)Q_{rk}(s)\,\dd s
\right]\xi .
\end{aligned}
\]
Hence, $\widetilde Y_n^{(k)}$ has the same form as
\eqref{eq:surrogate-linear-representation}.

Likewise, substituting the first identity in
\eqref{eq:surrogate-linear-representation} into
\eqref{eq:main-surrogate-backward-clean} gives
\[
\begin{aligned}
\widetilde B_n^{(k)}(t)
=-
\left[
\sum_{r<k}g_r
\int_t^1
\Phi_{t,s}^{\top}K_{r,1}(s)R_{rk}(s)\,\dd s
\right]W_0x+
\left[
\Phi_{t,1}^{\top}
-
\sum_{r<k}g_r
\int_t^1
\Phi_{t,s}^{\top}K_{r,2}(s)R_{rk}(s)\,\dd s
\right]\xi .
\end{aligned}
\]
Thus, $\widetilde B_n^{(k)}$ also has the form
\eqref{eq:surrogate-linear-representation}.

With \eqref{eq:surrogate-linear-representation}, the surrogate output is
\[
\langle \xi,\widetilde Y_n^{(k)}(1)\rangle_n
=
\frac1n\xi^\top K_{k,1}(1)W_0x
+
\frac1n\xi^\top K_{k,2}(1)\xi .
\]
Likewise,
\[
\begin{aligned}
\langle
\widetilde Y_n^{(a)}(t),
\widetilde Y_n^{(b)}(t)
\rangle_n
={}&
\frac1n(W_0x)^\top
K_{a,1}(t)^\top K_{b,1}(t)W_0x
+
\frac1n(W_0x)^\top
K_{a,1}(t)^\top K_{b,2}(t)\xi
\\
&+
\frac1n\xi^\top
K_{a,2}(t)^\top K_{b,1}(t)W_0x
+
\frac1n\xi^\top
K_{a,2}(t)^\top K_{b,2}(t)\xi .
\end{aligned}
\]
The corresponding formula for
$\langle\widetilde B_n^{(a)}(t),\widetilde B_n^{(b)}(t)\rangle_n$
is obtained by replacing $K_{r,i}$ with $L_{r,i}$.

Conditioning on $\Phi$, the matrices $K_{r,i}$ and $L_{r,i}$ are fixed,
while $W_0x$ and $\xi$ remain independent centered Gaussian vectors with
\[
\mathbb E[(W_0x)(W_0x)^\top]=q_0I_n,
\qquad
\mathbb E[\xi\xi^\top]=I_n.
\]
Therefore, for any such matrix $K$,
\[
\mathbb E\!\left[
\frac1n\xi^\top K\xi
\,\middle|\,\Phi
\right]
=
\frac1n
\tr\!\left(
K\,\mathbb E[\xi\xi^\top]
\right)
=
\tr_n(K),
\]
and similarly,
\[
\mathbb E\!\left[
\frac1n(W_0x)^\top K(W_0x)
\,\middle|\,\Phi
\right]
=
q_0\,\tr_n(K).
\]
The mixed terms vanish because $W_0x$ and $\xi$ are independent and centered:
\[
\mathbb E\!\left[
\frac1n(W_0x)^\top K\xi
\,\middle|\,\Phi
\right]
=0.
\]
Applying these identities to the expansions above gives
\begin{align}
\mathbb E\!\left[
\langle \xi,\widetilde Y_n^{(k)}(1)\rangle_n
\,\middle|\,\Phi
\right]
&=
\tr_n\!\bigl(K_{k,2}(1)\bigr),
\label{eq:surrogate-output-conditional-mean}\\
\mathbb E\!\left[
\langle
\widetilde Y_n^{(a)}(t),
\widetilde Y_n^{(b)}(t)
\rangle_n
\,\middle|\,\Phi
\right]
&=
q_0\tr_n\!\bigl(K_{a,1}(t)^\top K_{b,1}(t)\bigr)
+
\tr_n\!\bigl(K_{a,2}(t)^\top K_{b,2}(t)\bigr),
\label{eq:surrogate-forward-conditional-mean}\\
\mathbb E\!\left[
\langle
\widetilde B_n^{(a)}(t),
\widetilde B_n^{(b)}(t)
\rangle_n
\,\middle|\,\Phi
\right]
&=
q_0\tr_n\!\bigl(L_{a,1}(t)^\top L_{b,1}(t)\bigr)
+
\tr_n\!\bigl(L_{a,2}(t)^\top L_{b,2}(t)\bigr).
\label{eq:surrogate-backward-conditional-mean}
\end{align}

For the surrogate output, decompose
\[
\begin{aligned}
\langle \xi,\widetilde Y_n^{(k)}(1)\rangle_n-f_{\infty}^{(k),\tau}
={}&
\left(
\langle \xi,\widetilde Y_n^{(k)}(1)\rangle_n
-
\mathbb E\!\left[
\langle \xi,\widetilde Y_n^{(k)}(1)\rangle_n
\,\middle|\,\Phi
\right]
\right)+
\left(
\mathbb E\!\left[
\langle \xi,\widetilde Y_n^{(k)}(1)\rangle_n
\,\middle|\,\Phi
\right]
-f_{\infty}^{(k),\tau}
\right).
\end{aligned}
\]

Using \eqref{eq:surrogate-linear-representation} at $t=1$ and
\eqref{eq:surrogate-output-conditional-mean},
\[
\langle \xi,\widetilde Y_n^{(k)}(1)\rangle_n
-
\mathbb E\!\left[
\langle \xi,\widetilde Y_n^{(k)}(1)\rangle_n
\,\middle|\,\Phi
\right]
=
\underbrace{
\frac1n\xi^\top K_{k,1}(1)W_0x
}_{\mathrm{I}}
+
\underbrace{
\frac1n\xi^\top K_{k,2}(1)\xi
-
\tr_n\!\bigl(K_{k,2}(1)\bigr)
}_{\mathrm{II}}.
\]

The matrix-flow expansions established in the proof of
\cref{prop:surrogate-trace-clean}, together with
\eqref{eq:depth-base-max-clean}, imply that, for every fixed $T$ and
$q\ge2$,
\[
\sup_{n\ge1}
\max_{\substack{0\le k\le T\\ i\in\{1,2\}}}
\left\|
\sup_{t\in[0,1]}\|K_{k,i}(t)\|_{\op}
\right\|_{L^q}
+
\sup_{n\ge1}
\max_{\substack{0\le k\le T\\ i\in\{1,2\}}}
\left\|
\sup_{t\in[0,1]}\|L_{k,i}(t)\|_{\op}
\right\|_{L^q}
<\infty.
\]
Indeed, for fixed $T$, only finitely many matrix-flow words and word
lengths occur, and each product is controlled by
\eqref{eq:depth-base-max-clean} and H\"older's inequality.

Conditioning on $\Phi$, Term~I is a mixed Gaussian form in $W_0x$ and
$\xi$, while Term~II is a centered Gaussian quadratic form in $\xi$.
The operator-moment bounds above and the standard Gaussian moment
estimates therefore give
\[
\left\|
\langle \xi,\widetilde Y_n^{(k)}(1)\rangle_n
-
\mathbb E\!\left[
\langle \xi,\widetilde Y_n^{(k)}(1)\rangle_n
\,\middle|\,\Phi
\right]
\right\|_{L^p}
\lesssim n^{-1/2}.
\]

Applying \cref{prop:surrogate-trace-clean} to
\eqref{eq:surrogate-output-conditional-mean} gives
\[
\left\|
\mathbb E\!\left[
\langle \xi,\widetilde Y_n^{(k)}(1)\rangle_n
\,\middle|\,\Phi
\right]
-f_{\infty}^{(k),\tau}
\right\|_{L^p}
\lesssim n^{-1/2}.
\]
Together with the estimate for Terms~I and~II, this proves
\eqref{eq:surrogate-output-clean}.

For the forward inner products,
\eqref{eq:surrogate-linear-representation} and
\eqref{eq:surrogate-forward-conditional-mean} give
\[
\begin{aligned}
&\langle\widetilde Y_n^{(a)}(t),\widetilde Y_n^{(b)}(t)\rangle_n
-
\mathbb E\!\left[
\langle\widetilde Y_n^{(a)}(t),\widetilde Y_n^{(b)}(t)\rangle_n
\,\middle|\,\Phi
\right]
\\
={}&
\underbrace{
\frac1n(W_0x)^\top K_{a,1}(t)^\top K_{b,1}(t)W_0x
-
q_0\tr_n\!\bigl(K_{a,1}(t)^\top K_{b,1}(t)\bigr)
}_{\mathrm{I}}
\\
&+
\underbrace{
\frac1n(W_0x)^\top K_{a,1}(t)^\top K_{b,2}(t)\xi
}_{\mathrm{II}}
+
\underbrace{
\frac1n\xi^\top K_{a,2}(t)^\top K_{b,1}(t)W_0x
}_{\mathrm{III}}
\\
&+
\underbrace{
\frac1n\xi^\top K_{a,2}(t)^\top K_{b,2}(t)\xi
-
\tr_n\!\bigl(K_{a,2}(t)^\top K_{b,2}(t)\bigr)
}_{\mathrm{IV}} .
\end{aligned}
\]
Terms~I and~IV are centered Gaussian quadratic forms, while
Terms~II and~III are mixed Gaussian forms. The same Gaussian moment
bounds used for the surrogate output therefore give
\[
\sup_{t\in[0,1]}
\left\|
\langle\widetilde Y_n^{(a)}(t),\widetilde Y_n^{(b)}(t)\rangle_n
-
\mathbb E\!\left[
\langle\widetilde Y_n^{(a)}(t),\widetilde Y_n^{(b)}(t)\rangle_n
\,\middle|\,\Phi
\right]
\right\|_{L^p}
\lesssim n^{-1/2}.
\]

Applying \cref{prop:surrogate-trace-clean} to the two
normalized traces in
\eqref{eq:surrogate-forward-conditional-mean} gives
\[
\sup_{t\in[0,1]}
\left\|
\mathbb E\!\left[
\langle\widetilde Y_n^{(a)}(t),\widetilde Y_n^{(b)}(t)\rangle_n
\,\middle|\,\Phi
\right]
-
Q_{ab}(t)
\right\|_{L^p}
\lesssim n^{-1/2}.
\]
Combining the two estimates proves
\eqref{eq:surrogate-forward-contraction-clean}.

Replacing $K_{r,i}$ by $L_{r,i}$ in the decomposition above and using
\eqref{eq:surrogate-backward-conditional-mean} gives, in the same way,
\eqref{eq:surrogate-backward-contraction-clean}.
\end{proof}

\subsection{Approximation by the surrogate}

The backward comparison follows from the same depth-reversal and
transposition argument as in Appendix~B. We therefore only prove the forward
estimate below.

\begin{proof}[Proof of \cref{prop:main-width}]
We first prove the result for every even $p\ge2$. We proceed by induction over the training step $k$. At $k=0$, no gradient
update happens, and
\[
Y_n^{(0)}(t)
=
\widetilde Y_n^{(0)}(t)
=
\Phi_{0,t}W_0x,
\qquad
B_n^{(0)}(t)
=
\widetilde B_n^{(0)}(t)
=
\Phi_{t,1}^{\top}\xi.
\]
\cref{prop:surrogate-approximation-clean} then gives the
corresponding output and inner-product estimates.

Assume that, for every $r<k$ and every even $q\ge2$,
\[
\left\|
\sup_{t\in[0,1]}
\|Y_n^{(r)}(t)-\widetilde Y_n^{(r)}(t)\|_n
\right\|_{L^q}
+
\left\|
\sup_{t\in[0,1]}
\|B_n^{(r)}(t)-\widetilde B_n^{(r)}(t)\|_n
\right\|_{L^q}
\lesssim n^{-1/2},
\]
and $\|\gamma_n^{(r),\tau}-g_r\|_{L^q}
\lesssim n^{-1/2}.$

For the $k$th training step, define $V_n^{(k)}(s)
:=
\sum_{r<k}\frac{\gamma_n^{(r),\tau}}{n}
B_n^{(r)}(s)
\bigl(Y_n^{(r)}(s)\bigr)^\top.$
Then the finite-width forward state satisfies
\[
Y_n^{(k)}(t)
=
\Phi_{0,t}W_0x
-
\int_0^t
\Phi_{s,t}V_n^{(k)}(s)Y_n^{(k)}(s)\,\dd s.
\]
Let $E_k(t):=
Y_n^{(k)}(t)-\widetilde Y_n^{(k)}(t).$
Subtracting surrogate \eqref{eq:main-surrogate-forward-clean} from $Y_n^{(k)}(t)$ gives
\begin{equation}
E_k(t)
=
-\int_0^t
\Phi_{s,t}V_n^{(k)}(s)E_k(s)\,\dd s
-
\int_0^t
\Phi_{s,t}\mathcal F_k(s)\,\dd s,
\label{eq:width-forward-error-clean}
\end{equation}
where
\[
\begin{aligned}
\mathcal F_k(s)
=
&\sum_{r<k}\Bigl[
\underbrace{
\bigl(\gamma_n^{(r),\tau}-g_r\bigr)
B_n^{(r)}(s)
\left\langle
Y_n^{(r)}(s),\widetilde Y_n^{(k)}(s)
\right\rangle_n
}_{\mathrm{I}}
\\
&+
\underbrace{
g_r\bigl(B_n^{(r)}(s)-\widetilde B_n^{(r)}(s)\bigr)
\left\langle
Y_n^{(r)}(s),\widetilde Y_n^{(k)}(s)
\right\rangle_n
}_{\mathrm{II}}+
\underbrace{
g_r\widetilde B_n^{(r)}(s)
\left(
\left\langle
Y_n^{(r)}(s),\widetilde Y_n^{(k)}(s)
\right\rangle_n
-
Q_{rk}(s)
\right)
}_{\mathrm{III}}
\Bigr].
\end{aligned}
\]

For Term~III, write
\[
\left\langle
Y_n^{(r)}(s),\widetilde Y_n^{(k)}(s)
\right\rangle_n
-
Q_{rk}(s)
=
\underbrace{
\left\langle
Y_n^{(r)}(s)-\widetilde Y_n^{(r)}(s),
\widetilde Y_n^{(k)}(s)
\right\rangle_n
}_{\mathrm{IIIa}}
+
\underbrace{
\left(
\left\langle
\widetilde Y_n^{(r)}(s),\widetilde Y_n^{(k)}(s)
\right\rangle_n
-
Q_{rk}(s)
\right)
}_{\mathrm{IIIb}}.
\]
By Cauchy--Schwarz,
\[
|\mathrm{IIIa}|
\le
\left\|
Y_n^{(r)}(s)-\widetilde Y_n^{(r)}(s)
\right\|_n
\left\|
\widetilde Y_n^{(k)}(s)
\right\|_n.
\]

Since the induction has already established
\[
\|\gamma_n^{(r),\tau}-g_r\|_{L^q}\lesssim n^{-1/2},
\qquad
\left\|
\sup_t\|B_n^{(r)}(t)-\widetilde B_n^{(r)}(t)\|_n
\right\|_{L^q}
\lesssim n^{-1/2},
\]
and
\[
\left\|
\sup_t\|Y_n^{(r)}(t)-\widetilde Y_n^{(r)}(t)\|_n
\right\|_{L^q}
\lesssim n^{-1/2},
\qquad r<k,
\]
Terms~I, II, and IIIa are of order $n^{-1/2}$.  Term~IIIb is of the same
order by \cref{prop:surrogate-approximation-clean}.

\cref{prop:training-bounds}, together with
$\|V_n^{(r)}\|_{L_t^2(\mathrm F)}\le r\eta\tau\le T\eta\tau$,
gives uniform moments for
$\|Y_n^{(r)}\|_n$ and $\|B_n^{(r)}\|_n$.
\cref{prop:surrogate-approximation-clean}, applied with
$a=b$, gives the corresponding bounds for
$\|\widetilde Y_n^{(r)}\|_n$, $\|\widetilde B_n^{(r)}\|_n$, and
$\|\widetilde Y_n^{(k)}\|_n$.
Since $g_r$ is deterministic, Hölder's inequality gives
\[
\left\|
\left(
\int_0^1
\|\mathcal F_k(s)\|_n^2\,\dd s
\right)^{1/2}
\right\|_{L^{2p}}
\lesssim n^{-1/2}.
\]

Equation~\eqref{eq:width-forward-error-clean} has the form covered by the
continuous stability estimate in \cref{prop:training-stability},
with coefficient $V_n^{(k)}$ and forcing term $\mathcal F_k$. Since clipping
gives $\|V_n^{(k)}\|_{L_t^2(\mathrm F)}\le k\eta\tau$, the estimate above
yields
\begin{equation}
\left\|
\sup_{t\in[0,1]}
\|Y_n^{(k)}(t)-\widetilde Y_n^{(k)}(t)\|_n
\right\|_{L^p}
\lesssim n^{-1/2}.
\label{eq:width-forward-state-clean}
\end{equation}
By the same depth-reversal and transposition argument,
\begin{equation}
\left\|
\sup_{t\in[0,1]}
\|B_n^{(k)}(t)-\widetilde B_n^{(k)}(t)\|_n
\right\|_{L^p}
\lesssim n^{-1/2}.
\label{eq:width-backward-state-clean}
\end{equation}

The output error decomposes as
\[
F_n^{(k)}-f_{\infty}^{(k),\tau}
=
\underbrace{
\left\langle
\xi,
Y_n^{(k)}(1)-\widetilde Y_n^{(k)}(1)
\right\rangle_n
}_{\mathrm{(a)}}
+
\underbrace{
\left(
\left\langle
\xi,\widetilde Y_n^{(k)}(1)
\right\rangle_n
-f_{\infty}^{(k),\tau}
\right)
}_{\mathrm{(b)}}.
\]
For Term~(a), Cauchy--Schwarz, Hölder's inequality, and
\eqref{eq:width-forward-state-clean} give
\[
\|\mathrm{(a)}\|_{L^p}
\le
\|\|\xi\|_n\|_{L^{2p}}
\left\|
\sup_{t\in[0,1]}
\|Y_n^{(k)}(t)-\widetilde Y_n^{(k)}(t)\|_n
\right\|_{L^{2p}}
\lesssim n^{-1/2}.
\]
For Term~(b), \cref{prop:surrogate-approximation-clean} gives
$\|\mathrm{(b)}\|_{L^p}\lesssim n^{-1/2}$. Therefore,
\begin{equation}
\|F_n^{(k)}-f_{\infty}^{(k),\tau}\|_{L^p}
\lesssim n^{-1/2}.
\label{eq:width-output-clean}
\end{equation}

For $0\le a,b\le k$,
\[
\begin{aligned}
Q_{ab,n}(t)-Q_{ab}(t)
={}&
\left\langle
Y_n^{(a)}(t)-\widetilde Y_n^{(a)}(t),
Y_n^{(b)}(t)
\right\rangle_n
+
\left\langle
\widetilde Y_n^{(a)}(t),
Y_n^{(b)}(t)-\widetilde Y_n^{(b)}(t)
\right\rangle_n
\\
&+
\left(
\left\langle
\widetilde Y_n^{(a)}(t),
\widetilde Y_n^{(b)}(t)
\right\rangle_n
-
Q_{ab}(t)
\right).
\end{aligned}
\]
The first two terms are controlled by
\eqref{eq:width-forward-state-clean} and the uniform moments of
$Y_n^{(a)}$ and $\widetilde Y_n^{(a)}$, while the last term is controlled
by \cref{prop:surrogate-approximation-clean}. Hence
\[
\sup_{t\in[0,1]}
\|Q_{ab,n}(t)-Q_{ab}(t)\|_{L^p}
\lesssim n^{-1/2}.
\]
Using \eqref{eq:width-backward-state-clean} in the same way gives
\[
\sup_{t\in[0,1]}
\|R_{ab,n}(t)-R_{ab}(t)\|_{L^p}
\lesssim n^{-1/2}.
\]

Since
\[
\mathcal J_n^{(k)}
=
\int_0^1
Q_{kk,n}(t)R_{kk,n}(t)\,\dd t,
\qquad
J_k^\tau
=
\int_0^1
Q_{kk}(t)R_{kk}(t)\,\dd t,
\]
we have
\[
Q_{kk,n}R_{kk,n}-Q_{kk}R_{kk}
=
(Q_{kk,n}-Q_{kk})R_{kk,n}
+
Q_{kk}(R_{kk,n}-R_{kk}).
\]

The bounds for $Q_{kk,n}-Q_{kk}$ and $R_{kk,n}-R_{kk}$ above,
together with Hölder's inequality, therefore give
\[
\|\mathcal J_n^{(k)}-J_k^\tau\|_{L^p}
\lesssim n^{-1/2}.
\]

We have now shown
\[
\|F_n^{(k)}-f_{\infty}^{(k),\tau}\|_{L^p}
+
\|\mathcal J_n^{(k)}-J_k^\tau\|_{L^p}
\lesssim n^{-1/2}.
\]
Applying the preceding estimates at moment $2p$ and
\cref{lem:clipping-stability} gives
\begin{equation}
\|\gamma_n^{(k),\tau}-g_k\|_{L^p}
\lesssim n^{-1/2}.
\label{eq:width-gamma-clean}
\end{equation}
This closes the induction at step $k$.

Since the induction holds for every $k\le T$, and $T$ is fixed,
\eqref{eq:width-output-clean} gives
\[
\max_{0\le k\le T}
\|F_n^{(k)}-f_{\infty}^{(k),\tau}\|_{L^p}
\lesssim n^{-1/2}.
\]
For an arbitrary finite $p\ge2$, choose an even $p'\ge p$.
The $L^{p'}$ estimate proved above implies the corresponding $L^p$
estimate by monotonicity of $L^p$ norms.
This proves \cref{prop:main-width}.
\end{proof}

\section{Commutativity in Probability}
\label{app:gd-one-step-prob}

We prove \cref{prop:gd-one-step-prob} as a special case of the following
finite-horizon result.

\begin{proposition}[Width and Depth Limits commute in probability during training]
\label{prop:gd-finiteT-prob}
Fix $T\in\mathbb N_0$. There exist deterministic limits
$f_\infty^{(0)},\ldots,f_\infty^{(T)}$ such that
\[
\max_{0\le k\le T}
\left|
f_{n,L}^{(k)}-f_\infty^{(k)}
\right|
\xrightarrow[\min\{n,L\}\to\infty]{\mathbb P}
0.
\]
\end{proposition}

\begin{proof}
Define the deterministic unclipped limiting trajectory using the recursion in
\cref{app:deterministic-limit}, with
$\gamma_k^\tau$ replaced by
$\gamma_k:=\eta(f_\infty^{(k)}-y_\star)$.
For fixed $T$, the same finite-step Volterra construction defines finite
$f_\infty^{(k)}$ and $J_k$ for $0\le k\le T$.

For $0\le k<T$, let $G_k
:=
|f_\infty^{(k)}-y_\star|\sqrt{J_k},$
and choose $\tau_T
:=
1+\max_{0\le k<T}G_k.$

By construction, clipping is inactive at every update of the deterministic
$\tau_T$-clipped trajectory. Hence, by induction over $k$,
\begin{equation}
f_\infty^{(k),\tau_T}
=
f_\infty^{(k)},
\qquad
J_k^{\tau_T}
=
J_k,
\qquad
0\le k\le T.
\label{eq:finiteT-limit-agreement}
\end{equation}

For the finite network, define
\[
G_{n,L}^{(k),\tau_T}
:=
|f_{n,L}^{(k),\tau_T}-y_\star|
\sqrt{J_{n,L}^{(k),\tau_T}},
\qquad
0\le k<T.
\]
By \cref{thm:finiteT-sharp-main}, the output satisfies
\[
f_{n,L}^{(k),\tau_T}
\xrightarrow[\min\{n,L\}\to\infty]{\Ltwo}
f_\infty^{(k),\tau_T}.
\]
Moreover, the $J$-estimates established in the proofs of
\cref{prop:main-depth,prop:main-width} give
\[
J_{n,L}^{(k),\tau_T}
\xrightarrow[\min\{n,L\}\to\infty]{\Ltwo}
J_k^{\tau_T}.
\]
Together with \eqref{eq:finiteT-limit-agreement},
\[
G_{n,L}^{(k),\tau_T}
\xrightarrow[\min\{n,L\}\to\infty]{\mathbb P}
G_k,
\qquad
0\le k<T.
\]

Let
\[
E_{n,L}
:=
\left\{
G_{n,L}^{(k),\tau_T}<\tau_T
\text{ for all }0\le k<T
\right\}.
\]
Since $\tau_T-G_k\ge1$ for every $k<T$ and $T$ is fixed, $\mathbb P(E_{n,L})
\longrightarrow1.$

On $E_{n,L}$, the clipping factor is one at every step, so the updates are
exactly the same as ordinary gradient descent. Since the clipped and ordinary trajectories start
from the same initialization, induction over the training steps gives
\[
f_{n,L}^{(k)}
=
f_{n,L}^{(k),\tau_T},
\qquad
0\le k\le T,
\quad\text{on }E_{n,L}.
\]

Therefore, for every $\varepsilon>0$,
\[
\begin{aligned}
&\mathbb P\left(
\max_{0\le k\le T}
|f_{n,L}^{(k)}-f_\infty^{(k)}|
>\varepsilon
\right)\le
\mathbb P(E_{n,L}^c)
+
\mathbb P\left(
\max_{0\le k\le T}
|f_{n,L}^{(k),\tau_T}
-f_\infty^{(k),\tau_T}|
>\varepsilon
\right).
\end{aligned}
\]
The first term tends to zero by the definition of $E_{n,L}$, while the
second tends to zero by \cref{thm:finiteT-sharp-main}. This proves
\cref{prop:gd-finiteT-prob}.
\end{proof}

\begin{lemma}[Initialization limits]
\label{lem:initialization-limits}
As $\min\{n,L\}\to\infty$, $f_{n,L}^{(0)}
\xrightarrow{\Ltwo}
0,
J_{n,L}^{(0)}
\xrightarrow{\Ltwo}
q_0\mathrm e.$
Moreover, for any fixed
$\tau>|y_\star|\sqrt{q_0\mathrm e}$,
\[
f_\infty^{(1),\tau}
=
\eta y_\star q_0\mathrm e.
\]
\end{lemma}
\begin{proof}
At initialization, clipping has not yet been applied, so we omit the
superscript $\tau$. Independence of $\xi$ and $Y_L^{(0)}$ gives
\[
\mathbb E|f_{n,L}^{(0)}|^2
=
\frac{q_0(1+L^{-1})^L}{n}
\le
\frac{q_0\mathrm e}{n}.
\]
Hence,
$f_{n,L}^{(0)}\to0$ in $\Ltwo$ as
$\min\{n,L\}\to\infty$.

For the gradient term, the $k=0$ estimates established in the proofs of
\cref{prop:main-depth,prop:main-width} give
\(
J_{n,L}^{(0)}-\mathcal J_n^{(0)}
\xrightarrow[L\to\infty]{\Ltwo}
0
\quad\text{uniformly in } n,
\)
and
\(
\mathcal J_n^{(0)}
\xrightarrow[n\to\infty]{\Ltwo}
J_0,
\)
where $J_0$ denotes the limiting value at initialization. Therefore,
\[
J_{n,L}^{(0)}
\xrightarrow[\min\{n,L\}\to\infty]{\Ltwo}
J_0.
\]

It remains to identify $J_0$. The initialization covariance calculation of
\citet{hayou2023commutativewidthdepthscaling} applies here with the identity
activation. In their ReLU setting, the diagonal covariance equation contains
the factor
$\mathbb E[\operatorname{ReLU}(Z)^2]=1/2$.
Here $\phi(z)=z$, so
$\mathbb E[\phi(Z)^2]=1$, and the corresponding forward covariance satisfies
\[
\frac{\dd}{\dd t}Q_{00}(t)
=
Q_{00}(t),
\qquad
Q_{00}(0)=q_0.
\]
Thus, $Q_{00}(t)=q_0\mathrm e^t.$
Applying the same calculation after depth reversal, with
$R_{00}(1)=1$, gives \(
R_{00}(t)=\mathrm e^{1-t}.
\)
Consequently,
\[
J_0
=
\int_0^1
Q_{00}(t)R_{00}(t)\,\dd t
=
q_0\mathrm e.
\]

Now fix any $\tau>|y_\star|\sqrt{q_0\mathrm e}$.
Since $f_\infty^{(0),\tau}=0$ and $J_0=q_0\mathrm e$,
\[
|f_\infty^{(0),\tau}-y_\star|\sqrt{J_0}
=
|y_\star|\sqrt{q_0\mathrm e}
<
\tau.
\]
Hence clipping is inactive at the first update of the limiting system,
and $\gamma_0^\tau=-\eta y_\star$.

Applying \eqref{eq:limit-volterra} at $k=1$ and taking the output inner
product gives
\[
f_\infty^{(1),\tau}
=
-\gamma_0^\tau
\int_0^1Q_{00}(t)R_{00}(t)\,\dd t
=
-\gamma_0^\tau J_0
=
\eta y_\star q_0\mathrm e.
\]
\end{proof}

It remains to identify the one-step limit in
\cref{prop:gd-one-step-prob}. At initialization,
$f_\infty^{(0)}=0$ and $J_0=q_0\mathrm e$, as established in
\cref{lem:initialization-limits}. Hence
$\gamma_0=-\eta y_\star$. Applying the limiting recursion at $k=1$ gives
\[
f_\infty^{(1)}
=
-\gamma_0J_0
=
\eta y_\star q_0\mathrm e.
\]
Taking $T=1$ in \cref{prop:gd-finiteT-prob} therefore proves
\cref{prop:gd-one-step-prob}.

\section{Random-Matrix Estimates}
\label{app:random-matrix}

The surrogate output in \cref{app:width-proof} has the form
\[
\langle \xi,\widetilde Y_n^{(k)}(1)\rangle_n
=
\frac1n\xi^\top K_{k,1}(1)W_0x
+
\frac1n\xi^\top K_{k,2}(1)\xi .
\]
The estimates in this appendix establish the two comparisons
\[
\frac1n\xi^\top K_{k,1}(1)W_0x
\xrightarrow{\,n^{-1/2}\,}
0,
\qquad
\frac1n\xi^\top K_{k,2}(1)\xi
\xrightarrow{\,n^{-1/2}\,}
\tr_n\!\bigl(K_{k,2}(1)\bigr)
\xrightarrow{\,n^{-1}\,}
f_{\infty}^{(k),\tau} .
\]
We first establish an
$n^{-1}$ trace comparison for products of $\Phi$ and $\Phi^\top$, and then
extend it to the matrix expressions $K_{k,i}$ and $L_{k,i}$ appearing in the
surrogate system. This gives the $n^{-1/2}$ approximation of the surrogate
output and the corresponding surrogate inner products required in
\cref{app:width-proof}.

The trace comparison below connects the finite-width matrix flow with its
infinite-width limit:
\[
\begin{array}{ccccc}
\text{matrix Brownian motion } M_t
& \longrightarrow &
\text{matrix flow } \Phi_{s,t}
& \longrightarrow &
\tr_n(\mathsf w_n)
\\[2mm]
\Big\Downarrow
&&
\Big\Downarrow
&&
\Big\Downarrow
\\[2mm]
\text{circular Brownian motion } c_t
& \longrightarrow &
\text{circular flow } u_{s,t}
& \longrightarrow &
\varphi(\mathsf w_\infty).
\end{array}
\]
The upper row is the finite-width system introduced in
\cref{app:depth-euler}. The lower row gives its limit in noncommutative
distribution: products of \(\Phi_{s,t}\) and \(\Phi_{s,t}^{\top}\) converge
in normalized trace to the corresponding moments of \(u_{s,t}\) and
\(u_{s,t}^*\).

Let \((\mathfrak A,\varphi)\) be a tracial noncommutative probability
space, and let \(c_t\) be a standard circular Brownian motion in
\(\mathfrak A\). For \(0\le s\le t\le1\), define \(u_{s,t}\) by
\[
u_{s,t}
=
1+\int_s^t \dd c_r\,u_{s,r},
\qquad
u_{s,s}=1.
\]

\subsection{Deterministic limiting system}
\label{app:deterministic-limit}

We now define the infinite-width training dynamics. As shown by the linear
representation in \eqref{eq:surrogate-linear-representation}, the finite-width
forward and backward states are linear combinations of the two independent
Gaussian vectors \(W_0x\) and \(\xi\). Thus, a finite-width state has the form $A_nW_0x+B_n\xi$ for matrix coefficients \(A_n\) and \(B_n\).

The base flow \(\Phi_{s,t}\) acts linearly on such a state:
\[
A_nW_0x+B_n\xi
\;\xmapsto{\ \Phi_{s,t}\ }\;
(\Phi_{s,t}A_n)W_0x+(\Phi_{s,t}B_n)\xi.
\]

In the width limit, the matrix coefficients are represented by elements
\(h_1,h_2\in\mathcal H_0:=L^2(\mathfrak A,\varphi)\), with inner product
\(\langle h,g\rangle_{\mathcal H_0}
:=\operatorname{Re}\varphi(h^*g)\). We therefore represent a limiting
state by \((h_1,h_2)\in\mathcal H:=\mathcal H_0\oplus\mathcal H_0\).

Since \(u_{s,t}\) is the noncommutative limit of \(\Phi_{s,t}\), the
corresponding limiting action is
\[
(h_1,h_2)
\;\xmapsto{\ U_{s,t}\ }\;
(u_{s,t}h_1,u_{s,t}h_2),
\]
that is,
\[
U_{s,t}(h_1,h_2):=(u_{s,t}h_1,u_{s,t}h_2).
\]

The finite-width trained flow is defined by
\eqref{eq:trained-volterra-clean}, with base flow \(\Phi_{s,t}\) and
cumulative update \(V_n^{(k)}\). The limiting flow \(P_{s,t}^{(k)}\) is
defined by the corresponding Volterra equation with \(U_{s,t}\) and
\(V^{(k)}\). At initialization, \(V^{(0)}=0\), so
\(P_{s,t}^{(0)}:=U_{s,t}\).

Given the stages below \(k\), define
\[
V^{(k)}(t)h
:=
\sum_{r<k}\gamma_r^\tau
b^{(r)}(t)\langle y^{(r)}(t),h\rangle_{\mathcal H},
\]
and let
\begin{equation}
P_{s,t}^{(k)}
=
U_{s,t}
-
\int_s^t
U_{r,t}V^{(k)}(r)P_{s,r}^{(k)}\,\dd r.
\label{eq:limit-volterra}
\end{equation}

\subsection{Matrix-Brownian trace comparison}
\label{app:circular-flow}

We first establish the trace comparison underlying the width limit. At finite
width, we consider normalized traces of products of the matrix flow
$\Phi_{s,t}$ and its transpose. Their infinite-width limits are obtained from
the corresponding products of the circular flow $u_{s,t}$ and its adjoint.
At the trace level,
\[
\underbrace{
\tr_n\!\left(
\Phi_{a_1,b_1}\,
\Phi_{a_2,b_2}^{\top}\cdots
\Phi_{a_d,b_d}
\right)
}_{\text{finite width}}
\quad
\xrightarrow{\;\;n^{-1}\;\;}
\quad
\underbrace{
\varphi\!\left(
u_{a_1,b_1}\,
u_{a_2,b_2}^{*}\cdots
u_{a_d,b_d}
\right)
}_{\text{infinite width}} .
\]
The proposition below makes this comparison precise for arbitrary fixed
products and uniformly over the depth intervals.

\begin{proposition}[Matrix--Brownian trace comparison]
\label{prop:flow-trace-clean}
Let $\mathsf w_n$ be a matrix-flow word \footnotemark of length $d$ and let
$\mathsf w_\infty$ be its corresponding circular word.
For every even integer $p\ge2$, there is a finite constant
$C_{\rm tr}(d,p)$ such that
\begin{equation}
\left\|
\tr_n(\mathsf w_n)-\varphi(\mathsf w_\infty)
\right\|_{L^p}
\le
C_{\rm tr}(d,p)n^{-1}.
\label{eq:flow-trace-clean}
\end{equation}
The constant is independent of $n$ and of all intervals appearing in
the word.
\end{proposition}
\footnotetext{
A matrix-flow word of length $d$ is an ordered product of $d$ factors
$\Phi_{a,b}$ or $\Phi_{a,b}^{\top}$, where $0\le a\le b\le1$; the
intervals may overlap or coincide. Its circular counterpart is obtained
by replacing $\Phi_{a,b}$ with $u_{a,b}$ and
$\Phi_{a,b}^{\top}$ with $u_{a,b}^{*}$, without changing the order.
}
\begin{proof}
Fix an auxiliary mesh \(t_\ell:=\ell/N\).

\paragraph{Matrix Euler approximation.}
For every interval \([a,b]\) appearing in the word, set
\(a_N:=\lfloor Na\rfloor/N\) and \(b_N:=\lceil Nb\rceil/N\).
The flow property gives
\[
\Phi_{a_N,b_N}
=
\Phi_{b,b_N}\Phi_{a,b}\Phi_{a_N,a}.
\]
Hence, \eqref{eq:depth-base-max-clean} and
\eqref{eq:depth-flow-regularity-clean} imply
\[
\left\|
\|\Phi_{a,b}-\Phi_{a_N,b_N}\|_{\op}
\right\|_{L^p}
\lesssim N^{-1/2},
\]
uniformly in \(n\).

For \(t_i\le t_j\), define
\[
P_{i+1:j}^{n,N}
:=
\bigl(I+N^{-1/2}G_{j,N}\bigr)\cdots
\bigl(I+N^{-1/2}G_{i+1,N}\bigr),
\qquad
G_{\ell,N}
:=
\sqrt N\bigl(M_{t_\ell}-M_{t_{\ell-1}}\bigr).
\]
The matrices \(G_{\ell,N}\) are independent, and their entries are
i.i.d.\ \(\mathcal N(0,n^{-1})\). By
\eqref{eq:depth-euler-clean},
\[
\sup_{n\ge1}\max_{i\le j}
\left\|
\|P_{i+1:j}^{n,N}-\Phi_{t_i,t_j}\|_{\op}
\right\|_{L^p}
\lesssim N^{-1/2}.
\]
A telescoping expansion over the \(d\) factors of the word, together with
\eqref{eq:depth-base-max-clean} and H\"older's inequality, gives
\begin{equation}
\sup_{n\ge1}
\left\|
\tr_n(\mathsf w_n)-\tr_n(\mathsf w_{n,N})
\right\|_{L^p}
\le
C_{\rm mat}(d,p)N^{-1/2}.
\label{eq:trace-mesh-matrix}
\end{equation}

\paragraph{Circular Euler approximation.}
Let \(\bar u_{i,t}^{N}\) solve
\[
\bar u_{i,t}^{N}
=
1+\int_{t_i}^{t}\dd c_r\,
\bar u_{i,\pi_N(r)}^{N},
\]
where \(\pi_N(r)\) is the left endpoint of the mesh cell containing \(r\).
At the mesh points,
\[
\bar u_{i,t_j}^{N}
=
(1+\Delta c_{j,N})\cdots(1+\Delta c_{i+1,N}).
\]
The operator-norm Burkholder--Gundy inequality
\citep[Theorem~3.2.1]{biane1998stochastic} gives uniform operator-norm
bounds for \(u\) and \(\bar u^N\). Applying the same inequality to their
difference gives
\[E_i(t)
\le
\frac{C_0}{N}
+
C_1\int_{t_i}^{t}E_i(r)\,\dd r,
\qquad
E_i(t)
:=
\sup_{t_i\le q\le t}
\|u_{t_i,q}-\bar u_{i,q}^{N}\|_\infty^2,\]
Gronwall's inequality gives
\[
E_i(t)
\le
\frac{C_0}{N}
\exp\!\bigl(C_1(t-t_i)\bigr)
\le
\frac{C_0e^{C_1}}{N}.
\]
Hence,
\[
\sup_i\sup_{t_i\le t\le1}
\|u_{t_i,t}-\bar u_{i,t}^{N}\|_\infty
\le
C_{\rm free}N^{-1/2}.
\]

For each factor of \(\mathsf w_\infty\), the estimate above together with
the same endpoint approximation gives an \(O(N^{-1/2})\) operator-norm
error between that factor and its Euler approximation. Using the uniform
operator-norm bounds for the remaining factors and telescoping over the
\(d\) factors of the word,
\[
\|\mathsf w_\infty-\mathsf w_{\infty,N}\|_\infty
\le
C(d)N^{-1/2}.
\]
Since \(|\varphi(z)|\le\|z\|_\infty\),
\begin{equation}
\left|
\varphi(\mathsf w_\infty)
-
\varphi(\mathsf w_{\infty,N})
\right|
\le
C_{\rm free}(d)N^{-1/2}.
\label{eq:trace-mesh-free}
\end{equation}

\paragraph{Uniform control of the Euler expansions.}
Before comparing the matrix and circular Euler words, we first show that
their Wick expansions can be summed with constants independent of the
auxiliary mesh size \(N\).

Set $g_{\ell,N}
:=
\sqrt N\bigl(c_{t_\ell}-c_{t_{\ell-1}}\bigr).$
Then, \(g_{\ell,N}\) are freely independent standard circular elements,
and the matrix and circular Euler factors are
\(I+N^{-1/2}G_{\ell,N}\) and
\(1+N^{-1/2}g_{\ell,N}\), respectively.

Fix a mesh index \(\ell\). Across \(q\) copies of a word of length \(d\),
this index appears in at most \(qd\) Euler factors. When these factors are
expanded, each occurrence of the index \(\ell\) supplies either the identity
term or the increment term. If the increment term is selected from exactly
\(2m\) of these factors, the resulting scaling is \(N^{-m}\). There are at
most \(\binom{qd}{2m}\) ways to select these factors and
\((2m-1)!!\) ways to pair the resulting \(2m\) increments. Hence, the total
absolute Wick weight for the fixed index \(\ell\) is bounded by
\[
A_{q,d}(N)
:=
\sum_{m=0}^{\lfloor qd/2\rfloor}
\binom{qd}{2m}(2m-1)!!\,N^{-m}
\le
1+\frac{K_{q,d}}{N}.
\]
Since there are \(N\) mesh indices,
\begin{equation}
A_{q,d}(N)^N
\le
\left(1+\frac{K_{q,d}}{N}\right)^N
\le
\mathrm e^{K_{q,d}}.
\label{eq:uniform-euler-diagram-mass}
\end{equation}
Thus, the absolute sums of the Wick expansions used below are bounded
uniformly in \(N\).

\paragraph{Mean comparison.}
Consider a monomial in the Euler expansion containing \(2m\) Gaussian
factors,
\[
\mathcal M_{n,N}
:=
N^{-m}\tr_n(H_1\cdots H_{2m}),
\]
where each \(H_j\) is either \(G_{\ell_j,N}\) or
\(G_{\ell_j,N}^{\top}\). Let \(\mathcal P_2(2m)\) denote the set of
pairings of \(\{1,\ldots,2m\}\).

The mixed-trace expansions of
\citet[Lemmas~3.8 and~4.2]{redelmeier2014real}
express \(\mathbb E\mathcal M_{n,N}\) as a sum over
\(\rho\in\mathcal P_2(2m)\). Since the matrices \(G_{\ell,N}\) are
independent across \(\ell\), a pairing is nonzero only if
\(\ell_r=\ell_s\) for every \(\{r,s\}\in\rho\).

For such a pairing, let \(V(\rho)\) be the number of free matrix indices
remaining after the Wick contractions, and define
\(\chi(\rho):=V(\rho)-m+1\). The \(m\) Gaussian covariances give
\(n^{-m}\), the normalized trace gives \(n^{-1}\), and the \(V(\rho)\)
free indices give \(n^{V(\rho)}\). Hence, the corresponding power of
\(n\) is \(n^{V(\rho)-m-1}=n^{\chi(\rho)-2}\), and
\[
\mathbb E\mathcal M_{n,N}
=
N^{-m}
\sum_{\substack{\rho\in\mathcal P_2(2m)\\
\ell_r=\ell_s\ \text{for every }\{r,s\}\in\rho}}
n^{\chi(\rho)-2}.
\]

By \citet[Lemma~5.2]{redelmeier2014real}, \(\chi(\rho)\le2\). By
\citet[Lemma~5.8]{redelmeier2014real}, the pairings with
\(\chi(\rho)=2\) are precisely the noncrossing pairings that match an
untransposed factor \(G_{\ell,N}\) with a transposed factor
\(G_{\ell,N}^{\top}\) having the same mesh index.

Let \(\mathcal M_{\infty,N}\) be the circular monomial obtained by
replacing \(G_{\ell,N}\) with \(g_{\ell,N}\) and
\(G_{\ell,N}^{\top}\) with \(g_{\ell,N}^{*}\). The free Wick formula
contains exactly the same leading noncrossing pairings. Therefore,
the terms with \(\chi(\rho)=2\) sum to
\(\varphi(\mathcal M_{\infty,N})\), while every other pairing satisfies
\(\chi(\rho)\le1\) and is smaller by at least a factor \(n^{-1}\).
Consequently,
\[
\left|
\mathbb E\mathcal M_{n,N}
-
\varphi(\mathcal M_{\infty,N})
\right|
\le
n^{-1}N^{-m}
\#\left\{
\rho\in\mathcal P_2(2m):
\ell_r=\ell_s
\text{ for every }\{r,s\}\in\rho
\right\}.
\]

Summing over all monomials in the Euler expansion and applying
\eqref{eq:uniform-euler-diagram-mass} with \(q=1\), we obtain
\begin{equation}
\left|
\mathbb E\tr_n(\mathsf w_{n,N})
-
\varphi(\mathsf w_{\infty,N})
\right|
\le
C_{\rm mean}(d)n^{-1},
\label{eq:trace-euler-mean-uniform}
\end{equation}
uniformly in \(N\).

\paragraph{Fluctuation bound.}
Set \(Z_{n,N}:=\tr_n(\mathsf w_{n,N})\). By the triangle inequality,
\[
\left\|
Z_{n,N}-\varphi(\mathsf w_{\infty,N})
\right\|_{L^p}
\le
\left\|
Z_{n,N}-\mathbb EZ_{n,N}
\right\|_{L^p}
+
\left|
\mathbb EZ_{n,N}
-
\varphi(\mathsf w_{\infty,N})
\right|.
\]
The second term is \(O(n^{-1})\) by
\eqref{eq:trace-euler-mean-uniform}. It remains to control the centered
fluctuation.

For \(r\ge2\), let
\(\kappa_r(Z_{n,N},\ldots,Z_{n,N})\) denote the \(r\)-th cumulant.
By \citet[Lemma~4.4]{redelmeier2014real}, its expansion contains only
pairings that connect all \(r\) trace cycles. Each such pairing carries
a power \(n^{\chi-2r}\). Since
\(\chi\le2\) by \citet[Lemma~5.2]{redelmeier2014real}, and the total
absolute Wick weight is bounded uniformly in \(N\) by
\eqref{eq:uniform-euler-diagram-mass},
\begin{equation}
\left|
\kappa_r(Z_{n,N},\ldots,Z_{n,N})
\right|
\le
C_r(d)n^{2-2r},
\qquad r\ge2,
\label{eq:trace-euler-cumulant}
\end{equation}
uniformly in \(N\).

Let \(\mathcal P_{\ge2}(p)\) denote the partitions of
\(\{1,\ldots,p\}\) whose blocks all have size at least two. Since
\(Z_{n,N}-\mathbb EZ_{n,N}\) is centered and \(p\) is even, the
moment--cumulant formula gives
\[
\mathbb E
\bigl|Z_{n,N}-\mathbb EZ_{n,N}\bigr|^p
=
\sum_{\pi\in\mathcal P_{\ge2}(p)}
\prod_{B\in\pi}
\kappa_{|B|}(Z_{n,N},\ldots,Z_{n,N}).
\]
If \(\pi\) has \(c\) blocks, then \(c\le p/2\) and
\(\sum_{B\in\pi}|B|=p\). Hence, by
\eqref{eq:trace-euler-cumulant},
\[
\prod_{B\in\pi}
\left|
\kappa_{|B|}(Z_{n,N},\ldots,Z_{n,N})
\right|
\le
C(d,p)
n^{2c-2p}
\le
C(d,p)n^{-p}.
\]
Since \(\mathcal P_{\ge2}(p)\) is finite,
\[
\mathbb E
\bigl|Z_{n,N}-\mathbb EZ_{n,N}\bigr|^p
\le
C_{\rm cen}(d,p)n^{-p},
\]
and therefore
\begin{equation}
\left\|
Z_{n,N}-\mathbb EZ_{n,N}
\right\|_{L^p}
\le
C_{\rm cen}(d,p)n^{-1}.
\label{eq:trace-euler-centered-uniform}
\end{equation}

Combining \eqref{eq:trace-euler-mean-uniform} and
\eqref{eq:trace-euler-centered-uniform}, we obtain
\[
\left\|
\tr_n(\mathsf w_{n,N})
-
\varphi(\mathsf w_{\infty,N})
\right\|_{L^p}
\le
C_{\rm Eul}(d,p)n^{-1},
\]
uniformly in \(N\).

Finally, \eqref{eq:trace-mesh-matrix} and
\eqref{eq:trace-mesh-free} give
\[
\left\|
\tr_n(\mathsf w_n)
-
\varphi(\mathsf w_\infty)
\right\|_{L^p}
\le
C(d,p)N^{-1/2}
+
C_{\rm Eul}(d,p)n^{-1}.
\]
Letting \(N\to\infty\) proves
\eqref{eq:flow-trace-clean}.
\end{proof}

\subsection{Trace comparison for surrogate expressions}

\cref{app:width-proof} writes the surrogate states as
\[
\widetilde Y_n^{(k)}(t)
=
K_{k,1}(t)W_0x+K_{k,2}(t)\xi,
\qquad
\widetilde B_n^{(k)}(t)
=
L_{k,1}(t)W_0x+L_{k,2}(t)\xi.
\]
Accordingly,
\[
\langle \xi,\widetilde Y_n^{(k)}(1)\rangle_n
=
\frac1n\xi^\top K_{k,1}(1)W_0x
+
\frac1n\xi^\top K_{k,2}(1)\xi .
\]

The conditional expectations in
\eqref{eq:surrogate-output-conditional-mean},
\eqref{eq:surrogate-forward-conditional-mean}, and
\eqref{eq:surrogate-backward-conditional-mean}
reduce the width comparison to trace estimates for the coefficient matrices.
For the surrogate output, the required chain is
\[
\frac1n\xi^\top K_{k,1}(1)W_0x
\xrightarrow{\;n^{-1/2}\;}
0,
\qquad
\frac1n\xi^\top K_{k,2}(1)\xi
\xrightarrow{\;n^{-1/2}\;}
\tr_n\!\bigl(K_{k,2}(1)\bigr)
\xrightarrow{\;n^{-1}\;}
f_{\infty}^{(k),\tau} .
\]
The proposition below gives the corresponding $n^{-1}$ trace estimates for
the matrices in
\eqref{eq:surrogate-output-conditional-mean}--%
\eqref{eq:surrogate-backward-conditional-mean}.

\begin{proposition}[Trace comparison for surrogate coefficients]
\label{prop:surrogate-trace-clean}
Fix $T\ge0$ and an even $p\ge2$. There exists a finite constant
$C_{\rm st}(T,p)$, independent of $n$, such that
\begin{equation}
\max_{0\le k\le T}
\left\|
\tr_n\!\bigl(K_{k,2}(1)\bigr)-f_{\infty}^{(k),\tau}
\right\|_{L^p}
\le
C_{\rm st}(T,p)n^{-1}.
\label{eq:surrogate-output-trace-clean}
\end{equation}
Furthermore,
\begin{equation}
\max_{0\le a,b\le T}
\sup_{t\in[0,1]}
\left\|
q_0\tr_n\!\bigl(K_{a,1}(t)^\top K_{b,1}(t)\bigr)
+
\tr_n\!\bigl(K_{a,2}(t)^\top K_{b,2}(t)\bigr)
-
Q_{ab}(t)
\right\|_{L^p}
\le
C_{\rm st}(T,p)n^{-1},
\label{eq:surrogate-forward-trace-clean}
\end{equation}
and
\begin{equation}
\max_{0\le a,b\le T}
\sup_{t\in[0,1]}
\left\|
q_0\tr_n\!\bigl(L_{a,1}(t)^\top L_{b,1}(t)\bigr)
+
\tr_n\!\bigl(L_{a,2}(t)^\top L_{b,2}(t)\bigr)
-
R_{ab}(t)
\right\|_{L^p}
\le
C_{\rm st}(T,p)n^{-1}.
\label{eq:surrogate-backward-trace-clean}
\end{equation}
\end{proposition}

\begin{proof}
We prove the forward estimates involving $K_{k,i}$. The corresponding
bounds for $L_{k,i}$ follow from the depth-reversal and transposition
argument of Section~\ref{app:depth-trained}.

We first establish the following structural claim: for every $k\le T$ and
$i\in\{1,2\}$, $K_{k,i}(t)$ is a finite sum of deterministic multiple
depth integrals of matrix-flow words.

Using the backward coefficient recursions in the forward recursions for
$K_{k,i}$ gives
\begin{align}
K_{k,1}(t)
={}&
\Phi_{0,t}
+
\sum_{q<r<k} g_rg_q
\int_0^t\int_s^1
Q_{rk}(s)R_{qr}(v)\,
\Phi_{s,t}\Phi_{s,v}^{\top}K_{q,1}(v)
\,\dd v\,\dd s,
\label{eq:K1-closed-recursion}\\
K_{k,2}(t)
={}&
-\sum_{r<k}g_r
\int_0^t
Q_{rk}(s)\,
\Phi_{s,t}\Phi_{s,1}^{\top}\,\dd s
\nonumber\\
&+
\sum_{q<r<k} g_rg_q
\int_0^t\int_s^1
Q_{rk}(s)R_{qr}(v)\,
\Phi_{s,t}\Phi_{s,v}^{\top}K_{q,2}(v)
\,\dd v\,\dd s.
\label{eq:K2-closed-recursion}
\end{align}

The claim follows by induction on $k$. At $k=0$,
$K_{0,1}(t)=\Phi_{0,t}$ and $K_{0,2}(t)=0$. Assume that the claim holds
for every $q<k$. In
\eqref{eq:K1-closed-recursion}--\eqref{eq:K2-closed-recursion},
each occurrence of $K_{q,i}$ is multiplied by
$\Phi_{s,t}\Phi_{s,v}^{\top}$ and integrated over $s$ and $v$.
Hence every matrix-flow word in the expansion of $K_{q,i}$ is replaced by $\Phi_{s,t}\Phi_{s,v}^{\top}\mathsf w_n,$
which is again a matrix-flow word. The remaining coefficients
$g_rg_qQ_{rk}(s)R_{qr}(v)$ are deterministic. Thus, $K_{k,1}$ and
$K_{k,2}$ are finite sums of deterministic multiple depth integrals of
matrix-flow words.

Let $\bar K_{k,i}(t)$ denote the corresponding expression obtained by
replacing every matrix-flow word with its circular word. For fixed $T$, only finitely many word lengths occur in these expansions.
Since \cref{prop:flow-trace-clean} is uniform over the depth
variables, Minkowski's inequality gives
\begin{equation}
\max_{\substack{0\le k\le T\\ i\in\{1,2\}}}
\sup_{t\in[0,1]}
\left\|
\tr_n\!\bigl(K_{k,i}(t)\bigr)
-
\varphi\!\bigl(\bar K_{k,i}(t)\bigr)
\right\|_{L^p}
\le C_{\rm st}^{(1)}(T,p)n^{-1}.
\label{eq:surrogate-single-coefficient-trace}
\end{equation}

The same argument applies to the products in
\eqref{eq:surrogate-forward-conditional-mean}. Expanding both coefficient
matrices expresses
$K_{a,i}(t)^\top K_{b,j}(t)$ as a finite sum of multiple depth integrals
whose matrix-valued factors are again matrix-flow words. Hence
\begin{equation}
\max_{0\le a,b\le T}
\sup_{t\in[0,1]}
\left\|
\tr_n\!\bigl(K_{a,i}(t)^\top K_{b,j}(t)\bigr)
-
\varphi\!\bigl(
\bar K_{a,i}(t)^*\bar K_{b,j}(t)
\bigr)
\right\|_{L^p}
\le C_{\rm st}^{(2)}(T,p)n^{-1}.
\label{eq:surrogate-product-coefficient-trace}
\end{equation}

It remains to identify the circular coefficients with the limiting system.
Define
\[
\bar y^{(k)}(t)
:=
\bigl(
\sqrt{q_0}\,\bar K_{k,1}(t),
\bar K_{k,2}(t)
\bigr),
\qquad
\bar b^{(k)}(t)
:=
\bigl(
\sqrt{q_0}\,\bar L_{k,1}(t),
\bar L_{k,2}(t)
\bigr).
\]
Replacing $\Phi$ by $u$ and $\Phi^\top$ by $u^*$ in the coefficient
recursions shows that $\bar y^{(k)}$ and $\bar b^{(k)}$ satisfy
\eqref{eq:limit-volterra}. By uniqueness,
$\bar y^{(k)}=y^{(k)}$ and $\bar b^{(k)}=b^{(k)}$.

Consequently,
\[
f_{\infty}^{(k),\tau}
=
\operatorname{Re}\varphi\!\bigl(\bar K_{k,2}(1)\bigr),
\]
and
\[
Q_{ab}(t)
=
q_0\operatorname{Re}\varphi\!\bigl(
\bar K_{a,1}(t)^*\bar K_{b,1}(t)
\bigr)
+
\operatorname{Re}\varphi\!\bigl(
\bar K_{a,2}(t)^*\bar K_{b,2}(t)
\bigr).
\]
Similarly,
\[
R_{ab}(t)
=
q_0\operatorname{Re}\varphi\!\bigl(
\bar L_{a,1}(t)^*\bar L_{b,1}(t)
\bigr)
+
\operatorname{Re}\varphi\!\bigl(
\bar L_{a,2}(t)^*\bar L_{b,2}(t)
\bigr).
\]
Therefore
\eqref{eq:surrogate-single-coefficient-trace} and
\eqref{eq:surrogate-product-coefficient-trace} give
\eqref{eq:surrogate-output-trace-clean},
\eqref{eq:surrogate-forward-trace-clean}, and
\eqref{eq:surrogate-backward-trace-clean}.
\end{proof}

\section{Stability under Clipped Training}
\label{app:training-stability}

The training update at depth $t$ depends on both the forward and backward
states. At initialization,
\[
Y^{(0)}(t)=\Phi_{0,t}Y_0,
\qquad
B^{(0)}(t)=\Phi_{t,1}^{\top}\xi .
\]
Thus $Y^{(0)}(t)$ depends on the matrix-Brownian increments over $[0,t]$,
whereas $B^{(0)}(t)$ depends on those over $[t,1]$. Let
$\mathcal F_t^M:=\sigma(M_s:0\le s\le t)$.

Since
\[
V^{(k)}(t)
=
\sum_{r<k}\frac{\gamma_r^\tau}{n}
B^{(r)}(t)\bigl(Y^{(r)}(t)\bigr)^\top
\]
contains the backward states, it is not $\mathcal F_t^M$-measurable.
Consequently, the predictability condition in
\cref{lem:matrix-brownian-maximal} is not available during training.

We therefore use the Volterra identity
\begin{equation}
\Phi_{s,t}^{V}
=
\Phi_{s,t}
-
\int_s^t
\Phi_{r,t}V(r)\Phi_{s,r}^{V}\,\dd r.
\label{eq:trained-volterra-clean}
\end{equation}

Taking operator norms gives
\[
\|\Phi_{s,t}^{V}\|_{\op}
\le
\|\Phi_{s,t}\|_{\op}
+
\int_s^t
\|\Phi_{r,t}\|_{\op}
\|V(r)\|_{\op}
\|\Phi_{s,r}^{V}\|_{\op}\,\dd r,
\]
where the factors involving $\Phi$ are controlled by
\eqref{eq:depth-base-max-clean}. Thus it remains to control the size of
the cumulative training update. By global clipping, $\|V^{(k)}\|_{L_t^2(\mathrm F)}
\le
k\eta\tau.$
Section~\ref{app:training-bounds} uses this bound to control
$\Phi^{V^{(k)}}$ and hence the moments of $Y^{(k)}$ and $B^{(k)}$.
Section~\ref{app:training-stability-estimates} shows that the
$O(L^{-1/2})$ and $O(n^{-1/2})$ comparison rates are preserved when
forming the next clipped update.

At stage $k$, the induction closes through
\[
V^{(k)}
\longrightarrow
(Y^{(k)},B^{(k)},f^{(k)},J^{(k)})
\longrightarrow
\gamma^{(k)}
\longrightarrow
V^{(k+1)}.
\]

\subsection{Moment control during training}
\label{app:training-bounds}

\begin{proposition}[Uniform bounds for trained flows]
\label{prop:training-bounds}
Let $\Lambda<\infty$ and suppose that $\|V\|_{L_t^2(\mathrm F)}
\le
\Lambda
\text{ a.s.}$ Let $\Phi_{s,t}^V$ be the trained propagator defined by
\[
\Phi_{s,t}^{V}
=
\Phi_{s,t}
-
\int_s^t
\Phi_{r,t}V(r)\Phi_{s,r}^{V}\,\dd r.
\]
Then this equation has a unique solution. Moreover, for every $p\ge2$,
there exists $C_{\rm train}(p,\Lambda)<\infty$, independent of $n$, such
that
\begin{equation}
\sup_{0\le s\le1}
\left\|
\sup_{s\le t\le1}
\|\Phi_{s,t}^{V}\|_{\op}
\right\|_{L^p}
\le
C_{\rm train}(p,\Lambda).
\label{eq:trained-flow-continuous-clean}
\end{equation}
The same bound holds for the reversed and transposed flow.

For the discrete trained states, for every fixed $T$ and $p\ge2$, there
exists $C_{\rm state}(T,p)<\infty$, independent of $n$ and $L$,
such that
\[
\max_{0\le k\le T}
\left\|
\max_{0\le j\le L}
\|Y_{n,L}^{(k),\tau}(t_j)\|_n
\right\|_{L^p}
+
\max_{0\le k\le T}
\left\|
\max_{0\le j\le L}
\|B_{n,L}^{(k),\tau}(t_j)\|_n
\right\|_{L^p}
\le
C_{\rm state}(T,p).
\]
\end{proposition}

\begin{proof}
We first prove pathwise well-posedness. Since
\[
\|V(r)\|_{\op}\le \|V(r)\|_{\mathrm F},
\]
Cauchy--Schwarz gives
\begin{equation}
\int_0^1\|V(r)\|_{\op}\,\dd r
\le
\|V\|_{L_t^2(\mathrm F)}
\le
\Lambda .
\label{eq:trained-update-L1-clean}
\end{equation}

Fix a realization for which \eqref{eq:trained-update-L1-clean} holds.
Since $(s,t)\mapsto\Phi_{s,t}$ is continuous on
$\{0\le s\le t\le1\}$,
\[
K_\Phi
:=
\sup_{0\le s\le t\le1}
\|\Phi_{s,t}\|_{\op}
<\infty .
\]
Iterating the Volterra equation gives
\[
\Phi_{s,t}^V
=
\Phi_{s,t}
+
\sum_{m\ge1}(-1)^m I_m(s,t),
\]
where
\[
I_m(s,t)
:=
\int_{s<r_1<\cdots<r_m<t}
\Phi_{r_m,t}V(r_m)
\Phi_{r_{m-1},r_m}\cdots
V(r_1)\Phi_{s,r_1}
\,\dd\mathbf r .
\]
By submultiplicativity of the operator norm,
\[
\|I_m(s,t)\|_{\op}
\le
K_\Phi^{m+1}
\int_{s<r_1<\cdots<r_m<t}
\prod_{i=1}^m\|V(r_i)\|_{\op}
\,\dd\mathbf r .
\]
The integrand is symmetric in $r_1,\ldots,r_m$, so
\[
\|I_m(s,t)\|_{\op}
\le
\frac{K_\Phi^{m+1}}{m!}
\left(
\int_s^t\|V(r)\|_{\op}\,\dd r
\right)^m
\le
K_\Phi\frac{(K_\Phi\Lambda)^m}{m!}.
\]
Hence, the Volterra series converges absolutely in operator norm.

For uniqueness, let $\Psi_{s,t}$ and $\widetilde\Psi_{s,t}$ be two
solutions and set
$D_{s,t}:=\Psi_{s,t}-\widetilde\Psi_{s,t}$. Then
\[
D_{s,t}
=
-\int_s^t
\Phi_{r,t}V(r)D_{s,r}\,\dd r.
\]
Therefore,
\[
\|D_{s,t}\|_{\op}
\le
K_\Phi
\int_s^t
\|V(r)\|_{\op}\,
\|D_{s,r}\|_{\op}\,\dd r.
\]
Gronwall's inequality gives $D_{s,t}=0$ for all $s\le t$.
Thus, the Volterra equation has a unique pathwise solution.

For $m\ge1$, submultiplicativity of the operator norm gives
\[
\begin{aligned}
\sup_{t\ge s}\|I_m(s,t)\|_{\op}
\le{}&
\int_{s<r_1<\cdots<r_m<1}
\sup_{t\ge r_m}\|\Phi_{r_m,t}\|_{\op}
\prod_{i=1}^{m-1}\|\Phi_{r_i,r_{i+1}}\|_{\op}
\|\Phi_{s,r_1}\|_{\op}
\prod_{i=1}^m\|V(r_i)\|_{\op}
\,\dd\mathbf r .
\end{aligned}
\]

By Cauchy--Schwarz in $r_1,\ldots,r_m$,
\[
\begin{aligned}
\sup_{t\ge s}\|I_m(s,t)\|_{\op}
\le{}&
\left(
\int_{s<r_1<\cdots<r_m<1}
\left[
\sup_{u\ge r_m}\|\Phi_{r_m,u}\|_{\op}
\prod_{i=1}^{m-1}\|\Phi_{r_i,r_{i+1}}\|_{\op}
\|\Phi_{s,r_1}\|_{\op}
\right]^2
\,\dd\mathbf r
\right)^{1/2}
\\
&\times
\left(
\int_{s<r_1<\cdots<r_m<1}
\prod_{i=1}^m\|V(r_i)\|_{\op}^2
\,\dd\mathbf r
\right)^{1/2}.
\end{aligned}
\]

Since $\|V(r)\|_{\op}\le\|V(r)\|_{\mathrm F}$ and the integrand is
symmetric in $r_1,\ldots,r_m$,
\[
\int_{s<r_1<\cdots<r_m<1}
\prod_{i=1}^m\|V(r_i)\|_{\op}^2\,\dd\mathbf r
\le
\frac1{m!}
\left(
\int_s^1\|V(r)\|_{\mathrm F}^2\,\dd r
\right)^m
\le
\frac{\Lambda^{2m}}{m!}.
\]
Hence
\[
\begin{aligned}
\sup_{t\ge s}\|I_m(s,t)\|_{\op}
\le
\frac{\Lambda^m}{\sqrt{m!}}
\left(
\int_{s<r_1<\cdots<r_m<1}
\left[
\sup_{u\ge r_m}\|\Phi_{r_m,u}\|_{\op}
\prod_{i=1}^{m-1}\|\Phi_{r_i,r_{i+1}}\|_{\op}
\|\Phi_{s,r_1}\|_{\op}
\right]^2
\,\dd\mathbf r
\right)^{1/2}.
\end{aligned}
\]

For fixed $s<r_1<\cdots<r_m$, the factors
\[
\Phi_{s,r_1},\,
\Phi_{r_1,r_2},\ldots,\,
\Phi_{r_{m-1},r_m},
\qquad
\sup_{u\ge r_m}\|\Phi_{r_m,u}\|_{\op},
\]
depend on disjoint matrix-Brownian increments. Hence they are independent,
and \eqref{eq:depth-base-max-clean} gives
\[
\left\|
\sup_{u\ge r_m}\|\Phi_{r_m,u}\|_{\op}
\prod_{i=1}^{m-1}\|\Phi_{r_i,r_{i+1}}\|_{\op}
\|\Phi_{s,r_1}\|_{\op}
\right\|_{L^p}
\le
C_{\rm base}(p)^{m+1}.
\]
Since $p\ge2$, Minkowski's inequality yields
\[
\begin{aligned}
&
\left\|
\left(
\int_{s<r_1<\cdots<r_m<1}
\left[
\sup_{u\ge r_m}\|\Phi_{r_m,u}\|_{\op}
\prod_{i=1}^{m-1}\|\Phi_{r_i,r_{i+1}}\|_{\op}
\|\Phi_{s,r_1}\|_{\op}
\right]^2
\,\dd\mathbf r
\right)^{1/2}
\right\|_{L^p}\le
\frac{C_{\rm base}(p)^{m+1}}{\sqrt{m!}}.
\end{aligned}
\]
Combining this with the preceding estimate gives
\begin{equation}
\left\|
\sup_{t\ge s}\|I_m(s,t)\|_{\op}
\right\|_{L^p}
\le
C_{\rm base}(p)
\frac{
\bigl(C_{\rm base}(p)\Lambda\bigr)^m
}{m!}.
\label{eq:trained-mth-moment-clean}
\end{equation}

Using the Volterra expansion and the triangle inequality in $L^p$,
\[
\begin{aligned}
\left\|
\sup_{t\ge s}\|\Phi_{s,t}^V\|_{\op}
\right\|_{L^p}
&\le
\left\|
\sup_{t\ge s}\|\Phi_{s,t}\|_{\op}
\right\|_{L^p}
+
\sum_{m\ge1}
\left\|
\sup_{t\ge s}\|I_m(s,t)\|_{\op}
\right\|_{L^p}
\\
&\le
C_{\rm base}(p)
\sum_{m\ge0}
\frac{\bigl(C_{\rm base}(p)\Lambda\bigr)^m}{m!}
\\
&=
C_{\rm base}(p)
\exp\!\bigl(C_{\rm base}(p)\Lambda\bigr).
\end{aligned}
\]
The right-hand side is independent of $s$ and $n$, which proves
\eqref{eq:trained-flow-continuous-clean}.

For the backward flow, set $u:=1-t$ and define
\[
\check M_u:=(M_1-M_{1-u})^\top,
\qquad
\check V(u):=V(1-u)^\top .
\]
Then $\check M$ has the same law as $M$, and $\|\check V\|_{L_t^2(\mathrm F)}
=
\|V\|_{L_t^2(\mathrm F)}
\le \Lambda.$
Therefore, we can apply the same estimate again to the reversed and transposed flow. Applying these bounds to $Y_0$ and $\xi$, together with their Gaussian
moment bounds, gives the corresponding uniform moments for the forward
and backward states.

The same argument gives the discrete state bounds used in
Section~\ref{app:depth-trained}. By \eqref{eq:depth-update-budget-clean},
for every $k\le T$,
\[
\frac1L\sum_{\ell=1}^L
\|V_{n,L}^{(k)}(t_\ell)\|_{\mathrm F}^2
\le
(T\eta\tau)^2.
\]

From
\[
Y_{n,L}^{(k),\tau}(t_j)
=
S_{1:j}^{n,L}Y_0
-
\frac1L\sum_{\ell_2=1}^j
S_{\ell_2+1:j}^{n,L}
V_{n,L}^{(k)}(t_{\ell_2})
Y_{n,L}^{(k),\tau}(t_{\ell_2-1}),
\]
substituting the same identity for
$Y_{n,L}^{(k),\tau}(t_{\ell_2-1})$ gives
\[
\begin{aligned}
Y_{n,L}^{(k),\tau}(t_j)
={}&
S_{1:j}^{n,L}Y_0
-
\frac1L\sum_{\ell_2=1}^j
S_{\ell_2+1:j}^{n,L}
V_{n,L}^{(k)}(t_{\ell_2})
S_{1:\ell_2-1}^{n,L}Y_0
\\
&+
\frac1{L^2}
\sum_{\ell_1<\ell_2\le j}
S_{\ell_2+1:j}^{n,L}
V_{n,L}^{(k)}(t_{\ell_2})
S_{\ell_1+1:\ell_2-1}^{n,L}
V_{n,L}^{(k)}(t_{\ell_1})
Y_{n,L}^{(k),\tau}(t_{\ell_1-1}).
\end{aligned}
\]

Continuing this substitution gives the finite expansion
\[
Y_{n,L}^{(k),\tau}(t_j)
=
\sum_{m=0}^j(-1)^m Y_{j,m}^{(k)},
\]
where $Y_{j,0}^{(k)}
=
S_{1:j}^{n,L}Y_0,$
and, for $m\ge1$,
\[
Y_{j,m}^{(k)}
=
\frac1{L^m}
\sum_{1\le\ell_1<\cdots<\ell_m\le j}
S_{\ell_m+1:j}^{n,L}
V_{n,L}^{(k)}(t_{\ell_m})
S_{\ell_{m-1}+1:\ell_m-1}^{n,L}
\cdots
V_{n,L}^{(k)}(t_{\ell_1})
S_{1:\ell_1-1}^{n,L}Y_0.
\]
Thus the order-$m$ term contains $m$ update factors, with base products
over the layer blocks between successive update indices.

By Cauchy--Schwarz over
$1\le\ell_1<\cdots<\ell_m\le L$ and
\eqref{eq:depth-update-budget-clean},
\[
\frac1{L^m}
\sum_{\ell_1<\cdots<\ell_m}
\prod_{i=1}^m
\|V_{n,L}^{(k)}(t_{\ell_i})\|_{\mathrm F}^2
\le
\frac{(T\eta\tau)^{2m}}{m!}.
\]
For each ordered tuple, the base products depend on disjoint blocks of
residual matrices and are therefore independent. Using
\eqref{eq:depth-base-max-clean} and the same Cauchy--Schwarz and
Minkowski argument as above gives
\[
\left\|
\max_{0\le j\le L}
\|Y_{j,m}^{(k)}\|_n
\right\|_{L^p}
\le
C_{\rm base}(2p)
\frac{
\bigl(C_{\rm base}(2p)T\eta\tau\bigr)^m
}{m!}
\left\|\|Y_0\|_n\right\|_{L^{2p}}.
\]
Summing over $m$ yields
\[
\left\|
\max_{0\le j\le L}
\|Y_{n,L}^{(k),\tau}(t_j)\|_n
\right\|_{L^p}
\le
C_{\rm base}(2p)
\exp\!\bigl(C_{\rm base}(2p)T\eta\tau\bigr)
\left\|\|Y_0\|_n\right\|_{L^{2p}}.
\]

The Gaussian initialization gives
$\sup_n\|\|Y_0\|_n\|_{L^{2p}}<\infty$.
Applying the same argument after reversal and transposition, with terminal
state $\xi$, gives the corresponding bound for
$B_{n,L}^{(k),\tau}$. Taking the maximum over $k\le T$ proves the
discrete state bound.
\end{proof}

\subsection{Error propagation during training}
\label{app:training-stability-estimates}
Let $H_n$ denote $\mathbb R^n$ equipped with the normalized norm
$\|z\|_n:=n^{-1/2}\|z\|_2$.
\begin{proposition}[Bounds for the discrete and continuous error equations]
\label{prop:training-stability}
Fix $p\ge2$ and $\Lambda<\infty$, and let $V$ satisfy
$\|V\|_{L_t^2(\mathrm F)}\le\Lambda$ a.s.

For integers $1\le a\le b\le L$, consider the discrete equation
\begin{equation}
H_{a,b}
=
\mathcal R_{a,b}
-
\sum_{j=a}^b
\int_{I_j}
\Phi_{r,t_b}V(r)H_{a,j-1}\,\dd r,
\qquad
H_{a,a-1}=\mathcal R_{a,a-1},
\label{eq:grid-stability-clean}
\end{equation}
there exists $C_{\rm stab}(p,\Lambda)<\infty$, independent of
$n$, $L$, and $a$, such that
\begin{equation}
\left\|
\max_{a\le b\le L}\|H_{a,b}\|_n
\right\|_{L^p}
\le
C_{\rm stab}(p,\Lambda)
\left\|
\max_{b\ge a-1}\|\mathcal R_{a,b}\|_n
\right\|_{L^{2p}}.
\label{eq:grid-stability-bound-clean}
\end{equation}
The same estimate holds after reversal and transposition.

For the continuous equation
\begin{equation}
E(t)
=
-\int_0^t
\Phi_{s,t}V(s)E(s)\,\dd s
-
\int_0^t
\Phi_{s,t}F(s)\,\dd s,
\label{eq:continuous-stability-clean}
\end{equation}
we have
\begin{equation}
\left\|
\sup_{t\in[0,1]}\|E(t)\|_n
\right\|_{L^p}
\le
C_{\rm stab}(p,\Lambda)
\|F\|_{L^{2p}(\Omega;L_t^2(H_n))}.
\label{eq:continuous-stability-bound-clean}
\end{equation}
The same estimate holds after reversal and transposition.
\end{proposition}

\begin{proof}
We first consider the continuous equation. Define
$\widetilde E(t):=-\int_0^t\Phi_{s,t}^V F(s)\,\dd s$.
Using the Volterra identity \eqref{eq:trained-volterra-clean},
\[
\widetilde E(t)
=
-\int_0^t\Phi_{s,t}F(s)\,\dd s
+
\int_0^t\int_s^t
\Phi_{r,t}V(r)\Phi_{s,r}^V F(s)
\,\dd r\,\dd s.
\]
The double integral is over $0\le s\le r\le t$. By Fubini's theorem,
\[
\int_0^t\int_s^t
\Phi_{r,t}V(r)\Phi_{s,r}^V F(s)
\,\dd r\,\dd s
=
\int_0^t
\Phi_{r,t}V(r)
\left(
\int_0^r\Phi_{s,r}^V F(s)\,\dd s
\right)
\dd r.
\]
By the definition of $\widetilde E$,
\[
\int_0^r\Phi_{s,r}^V F(s)\,\dd s
=
-\widetilde E(r).
\]
Hence
\[
\widetilde E(t)
=
-\int_0^t\Phi_{s,t}F(s)\,\dd s
-
\int_0^t\Phi_{r,t}V(r)\widetilde E(r)\,\dd r.
\]
Thus $\widetilde E$ satisfies \eqref{eq:continuous-stability-clean}.
The same Gronwall argument used above gives uniqueness, and therefore
\begin{equation}
E(t)
=
-\int_0^t\Phi_{s,t}^V F(s)\,\dd s.
\label{eq:continuous-error-representation}
\end{equation}

It follows that
\[
\sup_{t\in[0,1]}\|E(t)\|_n
\le
\left(
\int_0^1
\sup_{t\ge s}\|\Phi_{s,t}^V\|_{\op}^2\,\dd s
\right)^{1/2}
\|F\|_{L_t^2(H_n)}.
\]
By H\"older's inequality,
\[
\begin{aligned}
\left\|
\sup_{t\in[0,1]}\|E(t)\|_n
\right\|_{L^p}
\le{}&
\left\|
\left(
\int_0^1
\sup_{t\ge s}\|\Phi_{s,t}^V\|_{\op}^2\,\dd s
\right)^{1/2}
\right\|_{L^{2p}}
\\
&\times
\|F\|_{L^{2p}(\Omega;L_t^2(H_n))}.
\end{aligned}
\]
Minkowski's inequality and \cref{prop:training-bounds} give
\[
\left\|
\left(
\int_0^1
\sup_{t\ge s}\|\Phi_{s,t}^V\|_{\op}^2\,\dd s
\right)^{1/2}
\right\|_{L^{2p}}
\le
C_{\rm train}(2p,\Lambda).
\]
Hence \eqref{eq:continuous-stability-bound-clean} holds with
$C_{\rm cont}(p,\Lambda):=C_{\rm train}(2p,\Lambda)$.

We next consider the discrete equation. Picard iteration of
\eqref{eq:grid-stability-clean} gives a finite expansion whose order-$m$
terms are indexed by $a\le j_1<\cdots<j_m\le b$. Let
$H_{a,b}^{[m]}$ denote the term containing $m$ factors of $V$.
By submultiplicativity,
\[
\begin{aligned}
\max_{b\ge a}\|H_{a,b}^{[m]}\|_n
\le{}&
\max_{c\ge a-1}\|\mathcal R_{a,c}\|_n
\sum_{a\le j_1<\cdots<j_m\le L}
\int_{I_{j_1}\times\cdots\times I_{j_m}}
\sup_{b\ge j_m}\|\Phi_{r_m,t_b}\|_{\op}
\\
&\times
\prod_{i=1}^{m-1}
\|\Phi_{r_i,t_{j_{i+1}-1}}\|_{\op}
\prod_{i=1}^m\|V(r_i)\|_{\op}
\,\dd\mathbf r .
\end{aligned}
\]
Applying Cauchy--Schwarz gives
\[
\begin{aligned}
\max_{b\ge a}\|H_{a,b}^{[m]}\|_n
\le{}&
\left(
\sum_{a\le j_1<\cdots<j_m\le L}
\int_{I_{j_1}\times\cdots\times I_{j_m}}
\left[
\sup_{b\ge j_m}\|\Phi_{r_m,t_b}\|_{\op}
\prod_{i=1}^{m-1}
\|\Phi_{r_i,t_{j_{i+1}-1}}\|_{\op}
\right]^2
\,\dd\mathbf r
\right)^{1/2}
\\
&\times
\left(
\sum_{a\le j_1<\cdots<j_m\le L}
\int_{I_{j_1}\times\cdots\times I_{j_m}}
\prod_{i=1}^m\|V(r_i)\|_{\op}^2
\,\dd\mathbf r
\right)^{1/2}
\max_{c\ge a-1}\|\mathcal R_{a,c}\|_n .
\end{aligned}
\]

Since $r_i\in I_{j_i}$ and $j_1<\cdots<j_m$, we have
$r_1<\cdots<r_m$. The sets
$I_{j_1}\times\cdots\times I_{j_m}$ corresponding to distinct index
tuples are disjoint. Therefore,
\[
\sum_{a\le j_1<\cdots<j_m\le L}
\int_{I_{j_1}\times\cdots\times I_{j_m}}
\prod_{i=1}^m\|V(r_i)\|_{\op}^2\,\dd\mathbf r
\le
\int_{0<r_1<\cdots<r_m<1}
\prod_{i=1}^m\|V(r_i)\|_{\op}^2\,\dd\mathbf r.
\]
Using $\|V(r)\|_{\op}\le\|V(r)\|_{\mathrm F}$ and symmetry in
$r_1,\ldots,r_m$,
\[
\int_{0<r_1<\cdots<r_m<1}
\prod_{i=1}^m\|V(r_i)\|_{\op}^2\,\dd\mathbf r
\le
\frac1{m!}
\left(
\int_0^1\|V(r)\|_{\mathrm F}^2\,\dd r
\right)^m
\le
\frac{\Lambda^{2m}}{m!}.
\]

For fixed $j_1<\cdots<j_m$ and $r_i\in I_{j_i}$, the factors
$\Phi_{r_i,t_{j_{i+1}-1}}$, $1\le i<m$, depend on Brownian increments
before $r_{i+1}$, while
$\sup_{b\ge j_m}\|\Phi_{r_m,t_b}\|_{\op}$ depends only on increments
after $r_m$. These intervals are disjoint, so the corresponding random
variables are independent. By \eqref{eq:depth-base-max-clean},
\[
\left\|
\sup_{b\ge j_m}\|\Phi_{r_m,t_b}\|_{\op}
\prod_{i=1}^{m-1}
\|\Phi_{r_i,t_{j_{i+1}-1}}\|_{\op}
\right\|_{L^{2p}}
\le
C_{\rm base}(2p)^m.
\]
Since each $I_j$ has length $L^{-1}$,
\[
\sum_{a\le j_1<\cdots<j_m\le L}
|I_{j_1}|\cdots|I_{j_m}|
=
\binom{L-a+1}{m}L^{-m}
\le
\frac1{m!}.
\]
Minkowski's inequality therefore gives
\[
\begin{aligned}
&\left\|
\left(
\sum_{a\le j_1<\cdots<j_m\le L}
\int_{I_{j_1}\times\cdots\times I_{j_m}}
\left[
\sup_{b\ge j_m}\|\Phi_{r_m,t_b}\|_{\op}
\prod_{i=1}^{m-1}
\|\Phi_{r_i,t_{j_{i+1}-1}}\|_{\op}
\right]^2
\,\dd\mathbf r
\right)^{1/2}
\right\|_{L^{2p}}
\le
\frac{C_{\rm base}(2p)^m}{\sqrt{m!}}.
\end{aligned}
\]

Combining the preceding estimates,
\[
\begin{aligned}
\max_{b\ge a}\|H_{a,b}^{[m]}\|_n
\le{}&
\frac{\Lambda^m}{\sqrt{m!}}
\left(
\sum_{a\le j_1<\cdots<j_m\le L}
\int_{I_{j_1}\times\cdots\times I_{j_m}}
\left[
\sup_{b\ge j_m}\|\Phi_{r_m,t_b}\|_{\op}
\prod_{i=1}^{m-1}
\|\Phi_{r_i,t_{j_{i+1}-1}}\|_{\op}
\right]^2
\,\dd\mathbf r
\right)^{1/2}
\\
&\times
\max_{c\ge a-1}\|\mathcal R_{a,c}\|_n .
\end{aligned}
\]
Taking $L^p$ norms and applying H\"older's inequality,
\[
\left\|
\max_{b\ge a}\|H_{a,b}^{[m]}\|_n
\right\|_{L^p}
\le
\frac{\bigl(C_{\rm base}(2p)\Lambda\bigr)^m}{m!}
\left\|
\max_{c\ge a-1}\|\mathcal R_{a,c}\|_n
\right\|_{L^{2p}}.
\]
Since $H_{a,b}^{[0]}=\mathcal R_{a,b}$, summing over $m$ gives
\[
\left\|
\max_{a\le b\le L}\|H_{a,b}\|_n
\right\|_{L^p}
\le
\exp\!\bigl(C_{\rm base}(2p)\Lambda\bigr)
\left\|
\max_{c\ge a-1}\|\mathcal R_{a,c}\|_n
\right\|_{L^{2p}}.
\]
Thus \eqref{eq:grid-stability-bound-clean} holds with
$C_{\rm disc}(p,\Lambda):=
\exp(C_{\rm base}(2p)\Lambda)$.

Reversal and transposition preserve the Brownian law, the
$L_t^2(\mathrm F)$ bound on $V$, and the relevant norms. Hence the same
estimates hold for the backward equations. Finally, taking
$C_{\rm stab}(p,\Lambda):=
\max\{C_{\rm cont}(p,\Lambda),C_{\rm disc}(p,\Lambda)\}$
proves the proposition.
\end{proof}

\begin{lemma}[Clipped coefficient comparison]
\label{lem:clipping-stability}
Fix $k\le T$ and $p\ge2$. Suppose
\[
\|F_n^{(k)}-f_{\infty}^{(k),\tau}\|_{L^{2p}}
+
\|\mathcal J_n^{(k)}-J_k^\tau\|_{L^{2p}}
\le
C_k n^{-1/2},
\]
and $\sup_n\|F_n^{(k)}\|_{L^{2p}}<\infty$. Then
\[
\|\gamma_n^{(k),\tau}-g_k\|_{L^p}
\le
C_{k,p}n^{-1/2}.
\]
\end{lemma}

\begin{proof}
We first establish a deterministic local bound for the clipping map.

If $|f_{\infty}^{(k),\tau}-y_\star|\sqrt{J_k^\tau}<\tau$, continuity implies that
there exists $\rho_k>0$ such that
$|F_n^{(k)}-f_{\infty}^{(k),\tau}|
+|\mathcal J_n^{(k)}-J_k^\tau|\le\rho_k$
implies
$|F_n^{(k)}-y_\star|\sqrt{\mathcal J_n^{(k)}}<\tau$.
In this case,
$\gamma_n^{(k),\tau}=\eta(F_n^{(k)}-y_\star)$ and
$g_k=\eta(f_{\infty}^{(k),\tau}-y_\star)$, and therefore
\[
|\gamma_n^{(k),\tau}-g_k|
=
\eta|F_n^{(k)}-f_{\infty}^{(k),\tau}|.
\]

Suppose instead that
$|f_{\infty}^{(k),\tau}-y_\star|\sqrt{J_k^\tau}\ge\tau$.
Since $\tau>0$, we have $f_{\infty}^{(k),\tau}-y_\star\neq0$ and $J_k^\tau>0$.
By continuity, there exists $\rho_k>0$ such that
$|F_n^{(k)}-f_{\infty}^{(k),\tau}|
+|\mathcal J_n^{(k)}-J_k^\tau|\le\rho_k$
implies
$\operatorname{sgn}(F_n^{(k)}-y_\star)
=\operatorname{sgn}(f_{\infty}^{(k),\tau}-y_\star)$
and $\mathcal J_n^{(k)}\ge J_k^\tau/2$.
Hence
\[
\begin{aligned}
|\gamma_n^{(k),\tau}-g_k|
\le{}&
\eta\bigl|
|F_n^{(k)}-y_\star|
-
|f_{\infty}^{(k),\tau}-y_\star|
\bigr|
+
\eta\tau
\left|
(\mathcal J_n^{(k)})^{-1/2}
-
(J_k^\tau)^{-1/2}
\right|.
\end{aligned}
\]
Here we used
$|\min\{a,b\}-\min\{c,d\}|
\le |a-c|+|b-d|$.
Since $x\mapsto x^{-1/2}$ has bounded derivative on
$[J_k^\tau/2,\infty)$, there is $L_k<\infty$ such that
\[
|\gamma_n^{(k),\tau}-g_k|
\le
L_k\left(
|F_n^{(k)}-f_{\infty}^{(k),\tau}|
+
|\mathcal J_n^{(k)}-J_k^\tau|
\right)
\]
whenever
$|F_n^{(k)}-f_{\infty}^{(k),\tau}|
+|\mathcal J_n^{(k)}-J_k^\tau|\le\rho_k$.

Combining the two cases, there exist $\rho_k>0$ and $L_k<\infty$ such that
\[
|\gamma_n^{(k),\tau}-g_k|
\le
L_k\left(
|F_n^{(k)}-f_{\infty}^{(k),\tau}|
+
|\mathcal J_n^{(k)}-J_k^\tau|
\right),
\]
whenever
$|F_n^{(k)}-f_{\infty}^{(k),\tau}|
+|\mathcal J_n^{(k)}-J_k^\tau|
\le\rho_k$.

On the event
$|F_n^{(k)}-f_{\infty}^{(k),\tau}|
+|\mathcal J_n^{(k)}-J_k^\tau|\le\rho_k$,
the local estimate gives
\[
\left\|
(\gamma_n^{(k),\tau}-g_k)
\mathbf 1_{\{
|F_n^{(k)}-f_{\infty}^{(k),\tau}|
+
|\mathcal J_n^{(k)}-J_k^\tau|
\le\rho_k
\}}
\right\|_{L^p}
\le
L_k C_k n^{-1/2}.
\]

For the complementary event, Markov's inequality gives
\[
\mathbb P\!\left(
|F_n^{(k)}-f_{\infty}^{(k),\tau}|
+
|\mathcal J_n^{(k)}-J_k^\tau|
>
\rho_k
\right)^{1/(2p)}
\le
\frac{C_k}{\rho_k}n^{-1/2}.
\]
By the definition of $\gamma_n^{(k),\tau}$,
$|\gamma_n^{(k),\tau}|
\le \eta|F_n^{(k)}-y_\star|$.
Hence, H\"older's inequality gives
\[
\left\|
(\gamma_n^{(k),\tau}-g_k)
\mathbf 1_{\{
|F_n^{(k)}-f_{\infty}^{(k),\tau}|
+
|\mathcal J_n^{(k)}-J_k^\tau|
>
\rho_k
\}}
\right\|_{L^p}
\le
\left(
\eta\|F_n^{(k)}-y_\star\|_{L^{2p}}
+
|g_k|
\right)
\frac{C_k}{\rho_k}n^{-1/2}.
\]
The assumption
$\sup_n\|F_n^{(k)}\|_{L^{2p}}<\infty$
implies
$\sup_n\|F_n^{(k)}-y_\star\|_{L^{2p}}<\infty$.
Hence, there exists $C_{k,p}<\infty$, independent of $n$, such that
\[
\|\gamma_n^{(k),\tau}-g_k\|_{L^p}
\le
C_{k,p}n^{-1/2}.
\]
\end{proof}

\subsection{Extension to Finite Training Sets and Vector-Valued Outputs}
\label{app:finite-training-set}

We state the main theory for a single training pair and a scalar output to
simplify the notation. We show here that the same commutativity result holds
for any fixed finite training set and any fixed finite output dimension.

Let \(\mathcal D=\{(x_i,y_i^\star)\}_{i=1}^{N_{\rm tr}}\), with
\(y_i^\star\in\mathbb R^{d_{\rm out}}\), where \(N_{\rm tr}\) and
\(d_{\rm out}\) are fixed independently of \(n\) and \(L\). Let \(\xi_1,\ldots,\xi_{d_{\rm out}}\in\mathbb R^n\) be independent
standard Gaussian vectors, fixed throughout training, and write
\(\Xi:=[\xi_1,\ldots,\xi_{d_{\rm out}}]\). The network
output is \(f_{n,L}(x)=n^{-1}\Xi^\top Y_L(x)\in\mathbb R^{d_{\rm out}}\),
and we minimize
\[
\mathcal L
=
\frac{1}{2N_{\rm tr}}
\sum_{i=1}^{N_{\rm tr}}
\|f_{n,L}(x_i)-y_i^\star\|_2^2.
\]

For sample \(i\), define the residual
\(e_i^{(k),\tau}:=f_{n,L}^{(k),\tau}(x_i)-y_i^\star
\in\mathbb R^{d_{\rm out}}\).
Define the backward matrix
\[
\mathbf B_\ell^{(k),\tau}
:=
\bigl(A_{\ell+1}^{(k),\tau}\bigr)^\top
\cdots
\bigl(A_L^{(k),\tau}\bigr)^\top
\Xi
\in\mathbb R^{n\times d_{\rm out}},
\]
which is independent of the sample index, and set
\[
C_\ell^{(k),\tau}
:=
\frac{1}{N_{\rm tr}}
\sum_{i=1}^{N_{\rm tr}}
Y_{\ell,i}^{(k),\tau}
\bigl(e_i^{(k),\tau}\bigr)^\top
\in\mathbb R^{n\times d_{\rm out}}.
\]
A direct differentiation gives
\[
\nabla_{W_\ell^{(k)}}\mathcal L
=
\frac{1}{n\sqrt L}
\mathbf B_\ell^{(k),\tau}
\bigl(C_{\ell-1}^{(k),\tau}\bigr)^\top,
\]
so every layer gradient has rank at most \(d_{\rm out}\).

Define
\[
J_{n,L}^{(k),\tau}
:=
\frac1L
\sum_{\ell=1}^L
\left\|
\frac1n
\mathbf B_\ell^{(k),\tau}
\bigl(C_{\ell-1}^{(k),\tau}\bigr)^\top
\right\|_{\mathrm F}^2.
\]
Then \(\sqrt{J_{n,L}^{(k),\tau}}\) is the Euclidean norm of the gradient
over all residual matrices. Under global clipping,
\[
W_\ell^{(k+1),\tau}
=
W_\ell^{(k),\tau}
-
\frac{\alpha_{n,L}^{(k),\tau}}{n\sqrt L}
\mathbf B_\ell^{(k),\tau}
\bigl(C_{\ell-1}^{(k),\tau}\bigr)^\top,
\]
where
\(\alpha_{n,L}^{(k),\tau}
:=\eta\min\{1,\tau/\sqrt{J_{n,L}^{(k),\tau}}\}\),
with multiplier one when \(J_{n,L}^{(k),\tau}=0\).

\begin{proposition}[Fixed finite training sets and output dimension]
\label{prop:finite-training-set}
Fix \(T<\infty\), \(\tau>0\), \(N_{\rm tr}<\infty\), and
\(d_{\rm out}<\infty\). For every \(1\le i\le N_{\rm tr}\) and
\(0\le k\le T\), there exists a deterministic limit
\(f_{i,\infty}^{(k),\tau}\in\mathbb R^{d_{\rm out}}\) such that
\[
\max_{\substack{0\le k\le T\\1\le i\le N_{\rm tr}}}
\left\|
f_{n,L}^{(k),\tau}(x_i)
-
f_{i,\infty}^{(k),\tau}
\right\|_{L^2(\Omega;\mathbb R^{d_{\rm out}})}
\le
C_{T,\tau,\mathcal D,d_{\rm out}}
\left(n^{-1/2}+L^{-1/2}\right),
\]
where the constant is independent of \(n\) and \(L\).
\end{proposition}

\begin{proof}
We verify the depth and width comparisons separately.

\paragraph{Depth comparison.}
Write \(B_{a,n,L}^{(k),\tau}\) and \(B_{a,n}^{(k),\tau}\) for the
\(a\)th columns of \(\mathbf B_{n,L}^{(k),\tau}\) and
\(\mathbf B_n^{(k),\tau}\), respectively, and similarly write
\(C_{a,n,L}^{(k),\tau}\) and \(C_{a,n}^{(k),\tau}\) for the columns of
\(\mathbf C_{n,L}^{(k),\tau}\) and \(\mathbf C_n^{(k),\tau}\).
Then, for \(t\in I_\ell\),
\[
\begin{aligned}
C_{a,n,L}^{(k),\tau}(t_{\ell-1})
&=
\frac1{N_{\rm tr}}
\sum_{i=1}^{N_{\rm tr}}
e_{i,n,L,a}^{(k),\tau}
Y_{i,n,L}^{(k),\tau}(t_{\ell-1}),
&
C_{a,n}^{(k),\tau}(t)
&=
\frac1{N_{\rm tr}}
\sum_{i=1}^{N_{\rm tr}}
e_{i,n,a}^{(k),\tau}
Y_{i,n}^{(k),\tau}(t),
\end{aligned}
\]
where \(e_{i,n,L,a}^{(k),\tau}\) and
\(e_{i,n,a}^{(k),\tau}\) denote the \(a\)th coordinates of the discrete
and continuous residual vectors.

The gradient fields can therefore be written as the finite sums
\[
\begin{aligned}
G_{n,L}^{(k)}(t)
&=
\frac1n
\sum_{a=1}^{d_{\rm out}}
B_{a,n,L}^{(k),\tau}(t_\ell)
\bigl(C_{a,n,L}^{(k),\tau}(t_{\ell-1})\bigr)^\top,
&
G_n^{(k)}(t)
&=
\frac1n
\sum_{a=1}^{d_{\rm out}}
B_{a,n}^{(k),\tau}(t)
\bigl(C_{a,n}^{(k),\tau}(t)\bigr)^\top .
\end{aligned}
\]

Let \(\Pi_\tau\) denote the metric projection onto the radius-\(\tau\)
ball of \(L_t^2(\mathrm F)\). The cumulative updates satisfy
\[
V_{n,L}^{(k+1)}
=
V_{n,L}^{(k)}+\eta\Pi_\tau(G_{n,L}^{(k)}),
\qquad
V_n^{(k+1)}
=
V_n^{(k)}+\eta\Pi_\tau(G_n^{(k)}),
\]
and hence
\(\|V_{n,L}^{(k)}\|_{L_t^2(\mathrm F)}
\vee\|V_n^{(k)}\|_{L_t^2(\mathrm F)}
\le k\eta\tau\).
Thus, \cref{prop:training-bounds,prop:training-stability} apply at every
fixed training step.

Each \(Y_{i,n,L}^{(k),\tau}\) satisfies the same forward equation as in
the single-sample proof, and each \(B_{a,n,L}^{(k),\tau}\) satisfies the
same backward equation as in the scalar-output proof. Hence,
\eqref{eq:depth-forward-state-clean} and
\eqref{eq:depth-backward-state-clean} give
\[
\begin{aligned}
\max_{1\le i\le N_{\rm tr}}
\left\|
\max_{0\le j\le L}
\|Y_{i,n,L}^{(k),\tau}(t_j)
-Y_{i,n}^{(k),\tau}(t_j)\|_n
\right\|_{L^p}
&\lesssim L^{-1/2},
\\
\max_{1\le a\le d_{\rm out}}
\left\|
\max_{0\le j\le L}
\|B_{a,n,L}^{(k),\tau}(t_j)
-B_{a,n}^{(k),\tau}(t_j)\|_n
\right\|_{L^p}
&\lesssim L^{-1/2}.
\end{aligned}
\]

The first estimate also gives
\[
\max_{1\le i\le N_{\rm tr}}
\left\|
f_{n,L}^{(k),\tau}(x_i)-F_{i,n}^{(k),\tau}
\right\|_{L^p(\Omega;\mathbb R^{d_{\rm out}})}
\lesssim L^{-1/2},
\]
since \(d_{\rm out}\) is fixed.

We next compare the columns of \(\mathbf C\). For \(t\in I_\ell\),
\[
\begin{aligned}
C_{a,n,L}^{(k),\tau}(t_{\ell-1})
-
C_{a,n}^{(k),\tau}(t)
&=
\frac1{N_{\rm tr}}
\sum_{i=1}^{N_{\rm tr}}
\bigl(
e_{i,n,L,a}^{(k),\tau}
-
e_{i,n,a}^{(k),\tau}
\bigr)
Y_{i,n,L}^{(k),\tau}(t_{\ell-1})
\\
& \quad+
\frac1{N_{\rm tr}}
\sum_{i=1}^{N_{\rm tr}}
e_{i,n,a}^{(k),\tau}
\bigl(
Y_{i,n,L}^{(k),\tau}(t_{\ell-1})
-
Y_{i,n}^{(k),\tau}(t)
\bigr).
\end{aligned}
\]
By the output estimate above,
\eqref{eq:depth-forward-state-clean},
\eqref{eq:depth-state-regularity-clean}, and
\cref{prop:training-bounds},
\[
\max_{1\le a\le d_{\rm out}}
\left\|
\left(
\sum_{\ell=1}^L
\int_{I_\ell}
\|C_{a,n,L}^{(k),\tau}(t_{\ell-1})
-
C_{a,n}^{(k),\tau}(t)\|_n^2\,\dd t
\right)^{1/2}
\right\|_{L^p}
\lesssim L^{-1/2}.
\]

Finally, expand the difference of the gradient fields term by term:
\[
\begin{aligned}
G_{n,L}^{(k)}(t)-G_n^{(k)}(t)
={}&
\frac1n
\sum_{a=1}^{d_{\rm out}}
\bigl(
B_{a,n,L}^{(k),\tau}(t_\ell)
-
B_{a,n}^{(k),\tau}(t)
\bigr)
\bigl(C_{a,n,L}^{(k),\tau}(t_{\ell-1})\bigr)^\top
\\
&+
\frac1n
\sum_{a=1}^{d_{\rm out}}
B_{a,n}^{(k),\tau}(t)
\bigl(
C_{a,n,L}^{(k),\tau}(t_{\ell-1})
-
C_{a,n}^{(k),\tau}(t)
\bigr)^\top .
\end{aligned}
\]
For vectors \(u,v\), we have
\(n^{-1}\|uv^\top\|_{\mathrm F}=\|u\|_n\|v\|_n\).
Therefore, the preceding columnwise estimates,
\eqref{eq:depth-backward-regularity-clean}, and
\cref{prop:training-bounds} give
\[
\left\|
\|G_{n,L}^{(k)}-G_n^{(k)}\|_{L_t^2(\mathrm F)}
\right\|_{L^p}
\lesssim
L^{-1/2}.
\]
Since \(N_{\rm tr}\) and \(d_{\rm out}\) are fixed, the finite sums above
only change the constants.

The nonexpansiveness of \(\Pi_\tau\) closes the same training-step
induction as in \cref{prop:main-depth}. Hence,
\begin{equation}
\max_{\substack{0\le k\le T\\1\le i\le N_{\rm tr}}}
\left\|
f_{n,L}^{(k),\tau}(x_i)-F_{i,n}^{(k),\tau}
\right\|_{L^2(\Omega;\mathbb R^{d_{\rm out}})}
\le
C_{\rm D}L^{-1/2}.
\label{eq:finite-data-vector-depth}
\end{equation}

\paragraph{Width comparison.}
Define the input covariance matrix by
\(\Sigma_{ij}^{\rm in}:=d^{-1}x_i^\top x_j\), and choose
\(S^{\rm in}\) such that
\(\Sigma^{\rm in}=S^{\rm in}(S^{\rm in})^\top\).
There exist independent standard Gaussian vectors
\(\zeta_1,\ldots,\zeta_{N_{\rm tr}}\in\mathbb R^n\) such that
\[
W_0x_i
\overset d=
\sum_{r=1}^{N_{\rm tr}}
S_{ir}^{\rm in}\zeta_r,
\qquad
1\le i\le N_{\rm tr}.
\]
All initialization randomness is contained in the fixed family
\(\zeta_1,\ldots,\zeta_{N_{\rm tr}},
\xi_1,\ldots,\xi_{d_{\rm out}}\).

As in \eqref{eq:surrogate-linear-representation}, the surrogate forward
and backward states are finite linear combinations of these Gaussian
vectors with matrix coefficients. In the width limit, each matrix
coefficient is represented by an element of \(\mathcal H_0\). We therefore
set
\(\mathcal H:=\mathcal H_0^{\oplus(N_{\rm tr}+d_{\rm out})}\)
and write its elements as
\((h_1,\ldots,h_{N_{\rm tr}+d_{\rm out}})\).

At finite width, \(\Phi_{s,t}\) left-multiplies every matrix coefficient.
In the width limit, \(u_{s,t}\) left-multiplies each \(h_j\), so
\[
U_{s,t}(h_1,\ldots,h_{N_{\rm tr}+d_{\rm out}})
:=
\bigl(
u_{s,t}h_1,\ldots,
u_{s,t}h_{N_{\rm tr}+d_{\rm out}}
\bigr).
\]

Order the components of \(\mathcal H\) so that the first
\(N_{\rm tr}\) correspond to
\(\zeta_1,\ldots,\zeta_{N_{\rm tr}}\), and the last
\(d_{\rm out}\) correspond to
\(\xi_1,\ldots,\xi_{d_{\rm out}}\).
Starting from \(V^{(0)}=0\), suppose that \(V^{(k)}\) has been defined
and let \(P_{s,t}^{(k)}\) satisfy \eqref{eq:limit-volterra}. For each
training input \(i\), define
\[
y_i^{(k)}(t)
:=
P_{0,t}^{(k)}
\bigl(
S_{i1}^{\rm in}1,\ldots,
S_{iN_{\rm tr}}^{\rm in}1,
0,\ldots,0
\bigr).
\]
For each output coordinate \(a\), define
\[
b_a^{(k)}(t)
:=
\bigl(P_{t,1}^{(k)}\bigr)^*
\bigl(
0,\ldots,0,
\underbrace{1}_{N_{\rm tr}+a},
0,\ldots,0
\bigr).
\]
Let \(\mathbf Y_\infty^{(k)}(t)\) have columns
\(y_1^{(k)}(t),\ldots,y_{N_{\rm tr}}^{(k)}(t)\), and let
\(\mathbf B_\infty^{(k)}(t)\) have columns
\(b_1^{(k)}(t),\ldots,b_{d_{\rm out}}^{(k)}(t)\).

Let
\(\mathbf Y^\star:=[y_1^\star,\ldots,y_{N_{\rm tr}}^\star]\).
Define the limiting output and residual matrices by
\[
\mathbf F_\infty^{(k),\tau}
:=
\bigl(\mathbf B_\infty^{(k)}(1)\bigr)^*
\mathbf Y_\infty^{(k)}(1),
\qquad
\mathbf E_\infty^{(k),\tau}
:=
\mathbf F_\infty^{(k),\tau}-\mathbf Y^\star.
\]

For the finite-width continuous-depth model, let
\[
\mathbf Y_n^{(k),\tau}(t)
:=
\bigl[
Y_{1,n}^{(k),\tau}(t),\ldots,
Y_{N_{\rm tr},n}^{(k),\tau}(t)
\bigr],
\]
and define
\[
\mathbf F_n^{(k),\tau}
:=
\frac1n\Xi^\top\mathbf Y_n^{(k),\tau}(1),
\qquad
\mathbf E_n^{(k),\tau}
:=
\mathbf F_n^{(k),\tau}-\mathbf Y^\star.
\]
The finite-width and limiting gradient fields are
\[
\begin{aligned}
G_n^{(k)}(t)
&=
\frac{1}{nN_{\rm tr}}
\mathbf B_n^{(k),\tau}(t)
\mathbf E_n^{(k),\tau}
\bigl(\mathbf Y_n^{(k),\tau}(t)\bigr)^\top,
\\
\mathcal G^{(k)}(t)
&=
\frac1{N_{\rm tr}}
\mathbf B_\infty^{(k)}(t)
\mathbf E_\infty^{(k),\tau}
\bigl(\mathbf Y_\infty^{(k)}(t)\bigr)^* .
\end{aligned}
\]
Define
\(J_k^\tau
:=\int_0^1\|\mathcal G^{(k)}(t)\|_{\rm HS}^2\,\dd t\) and
\(\alpha_k^\tau
:=\eta\min\{1,\tau/\sqrt{J_k^\tau}\}\),
with multiplier one when \(J_k^\tau=0\). Set
\(V^{(k+1)}:=V^{(k)}+\alpha_k^\tau\mathcal G^{(k)}\).
This completes the limiting recursion at stage \(k\).

We next extend the surrogate construction of \cref{app:width-proof}.
For \(r,k\le T\), define
\[
\mathbf Q_{rk}(t)
:=
\bigl(\mathbf Y_\infty^{(r)}(t)\bigr)^*
\mathbf Y_\infty^{(k)}(t),
\qquad
\mathbf R_{rk}(t)
:=
\bigl(\mathbf B_\infty^{(r)}(t)\bigr)^*
\mathbf B_\infty^{(k)}(t).
\]
Let \(\mathbf Y_{0,n}\in\mathbb R^{n\times N_{\rm tr}}\) have
\(i\)th column \(W_0x_i\). Define
\[
\begin{aligned}
\widetilde{\mathbf Y}_n^{(k)}(t)
={}&
\Phi_{0,t}\mathbf Y_{0,n}
-
\frac1{N_{\rm tr}}
\sum_{r<k}\alpha_r^\tau
\int_0^t
\Phi_{s,t}
\widetilde{\mathbf B}_n^{(r)}(s)
\mathbf E_\infty^{(r),\tau}
\mathbf Q_{rk}(s)\,\dd s,
\\
\widetilde{\mathbf B}_n^{(k)}(t)
={}&
\Phi_{t,1}^{\top}\Xi
-
\frac1{N_{\rm tr}}
\sum_{r<k}\alpha_r^\tau
\int_t^1
\Phi_{t,s}^{\top}
\widetilde{\mathbf Y}_n^{(r)}(s)
\bigl(\mathbf E_\infty^{(r),\tau}\bigr)^\top
\mathbf R_{rk}(s)\,\dd s .
\end{aligned}
\]

Every column of
\(\widetilde{\mathbf Y}_n^{(k)}\) and
\(\widetilde{\mathbf B}_n^{(k)}\) is a finite linear combination of
\(\zeta_1,\ldots,\zeta_{N_{\rm tr}},
\xi_1,\ldots,\xi_{d_{\rm out}}\).
As in \eqref{eq:surrogate-linear-representation}, their matrix
coefficients are finite sums of multiple depth integrals of matrix-flow
words. Since \(N_{\rm tr}\) and \(d_{\rm out}\) are fixed,
\cref{prop:surrogate-approximation-clean,prop:surrogate-trace-clean}
apply entrywise. Hence, for every fixed \(p\ge2\),
\[
\begin{aligned}
&
\max_{0\le k\le T}
\left\|
\frac1n\Xi^\top\widetilde{\mathbf Y}_n^{(k)}(1)
-
\mathbf F_\infty^{(k),\tau}
\right\|_{L^p(\Omega;\mathrm F)}
\\
&\quad+
\max_{0\le r,k\le T}
\sup_{t\in[0,1]}
\left\|
\frac1n
\bigl(\widetilde{\mathbf Y}_n^{(r)}(t)\bigr)^\top
\widetilde{\mathbf Y}_n^{(k)}(t)
-
\mathbf Q_{rk}(t)
\right\|_{L^p(\Omega;\mathrm F)}
\\
&\quad+
\max_{0\le r,k\le T}
\sup_{t\in[0,1]}
\left\|
\frac1n
\bigl(\widetilde{\mathbf B}_n^{(r)}(t)\bigr)^\top
\widetilde{\mathbf B}_n^{(k)}(t)
-
\mathbf R_{rk}(t)
\right\|_{L^p(\Omega;\mathrm F)}
\lesssim
n^{-1/2}.
\end{aligned}
\]

We now compare the finite-width and surrogate states by induction over
\(k\). Assume that, for every \(r<k\), the forward and backward state
matrices are \(O(n^{-1/2})\) from their surrogate values and
\(\alpha_n^{(r),\tau}-\alpha_r^\tau=O(n^{-1/2})\) in every fixed
\(L^p\) norm.

For \(r<k\), define the mixed Gram matrices
\[
\widehat{\mathbf Q}_{rk,n}(t)
:=
\frac1n
\bigl(\mathbf Y_n^{(r),\tau}(t)\bigr)^\top
\widetilde{\mathbf Y}_n^{(k)}(t),
\qquad
\widehat{\mathbf R}_{rk,n}(t)
:=
\frac1n
\bigl(\mathbf B_n^{(r),\tau}(t)\bigr)^\top
\widetilde{\mathbf B}_n^{(k)}(t).
\]
The induction hypothesis and the surrogate estimates give
\[
\begin{aligned}
\sup_{t\in[0,1]}
\left\|
\widehat{\mathbf Q}_{rk,n}(t)-\mathbf Q_{rk}(t)
\right\|_{L^p(\Omega;\mathrm F)}
&\lesssim n^{-1/2},
\\
\sup_{t\in[0,1]}
\left\|
\widehat{\mathbf R}_{rk,n}(t)-\mathbf R_{rk}(t)
\right\|_{L^p(\Omega;\mathrm F)}
&\lesssim n^{-1/2}.
\end{aligned}
\]

Set
\(\mathbf D_{Y,k}(t)
:=\mathbf Y_n^{(k),\tau}(t)
-\widetilde{\mathbf Y}_n^{(k)}(t)\).
Subtracting the surrogate forward equation from the finite-width forward
equation gives
\[
\begin{aligned}
\mathbf D_{Y,k}(t)
={}&
-\int_0^t
\Phi_{s,t}V_n^{(k)}(s)\mathbf D_{Y,k}(s)\,\dd s
\\
&-
\frac1{N_{\rm tr}}
\sum_{r<k}
\int_0^t
\Phi_{s,t}
\Bigl[
\alpha_n^{(r),\tau}
\mathbf B_n^{(r),\tau}(s)
\mathbf E_n^{(r),\tau}
\widehat{\mathbf Q}_{rk,n}(s)
\\
&\hspace{40mm}
-
\alpha_r^\tau
\widetilde{\mathbf B}_n^{(r)}(s)
\mathbf E_\infty^{(r),\tau}
\mathbf Q_{rk}(s)
\Bigr]
\,\dd s .
\end{aligned}
\]
Each difference inside the brackets is \(O(n^{-1/2})\) by the induction
hypothesis, the output comparison at the previous stages, and the mixed
Gram estimate above. Applying \cref{prop:training-stability} to each
column gives
\[
\left\|
\sup_{t\in[0,1]}
n^{-1/2}
\|\mathbf Y_n^{(k),\tau}(t)
-\widetilde{\mathbf Y}_n^{(k)}(t)\|_{\rm F}
\right\|_{L^p}
\lesssim
n^{-1/2}.
\]
The same argument after depth reversal and transposition, using
\(\widehat{\mathbf R}_{rk,n}\), gives
\[
\left\|
\sup_{t\in[0,1]}
n^{-1/2}
\|\mathbf B_n^{(k),\tau}(t)
-\widetilde{\mathbf B}_n^{(k)}(t)\|_{\rm F}
\right\|_{L^p}
\lesssim
n^{-1/2}.
\]

The state estimates and the surrogate comparisons now give
\[
\left\|
\mathbf F_n^{(k),\tau}
-
\mathbf F_\infty^{(k),\tau}
\right\|_{L^p(\Omega;\mathrm F)}
\lesssim n^{-1/2}.
\]
Define the finite-width Gram matrices by
\[
\mathbf Q_{rk,n}(t)
:=
\frac1n
\bigl(\mathbf Y_n^{(r),\tau}(t)\bigr)^\top
\mathbf Y_n^{(k),\tau}(t),
\qquad
\mathbf R_{rk,n}(t)
:=
\frac1n
\bigl(\mathbf B_n^{(r),\tau}(t)\bigr)^\top
\mathbf B_n^{(k),\tau}(t).
\]
Then, for every \(r\le k\),
\[
\begin{aligned}
\sup_{t\in[0,1]}
\left\|
\mathbf Q_{rk,n}(t)-\mathbf Q_{rk}(t)
\right\|_{L^p(\Omega;\mathrm F)}
&\lesssim n^{-1/2},
\\
\sup_{t\in[0,1]}
\left\|
\mathbf R_{rk,n}(t)-\mathbf R_{rk}(t)
\right\|_{L^p(\Omega;\mathrm F)}
&\lesssim n^{-1/2}.
\end{aligned}
\]

The continuous-depth gradient norm can be written as
\[
\mathcal J_n^{(k)}
=
\frac1{N_{\rm tr}^2}
\int_0^1
\tr\!\left(
\mathbf E_n^{(k),\tau}
\mathbf Q_{kk,n}(t)
\bigl(\mathbf E_n^{(k),\tau}\bigr)^\top
\mathbf R_{kk,n}(t)
\right)\,\dd t,
\]
while
\[
J_k^\tau
=
\frac1{N_{\rm tr}^2}
\int_0^1
\tr\!\left(
\mathbf E_\infty^{(k),\tau}
\mathbf Q_{kk}(t)
\bigl(\mathbf E_\infty^{(k),\tau}\bigr)^\top
\mathbf R_{kk}(t)
\right)\,\dd t.
\]
Hence,
\[
\|\mathcal J_n^{(k)}-J_k^\tau\|_{L^p}
\lesssim n^{-1/2}.
\]

Define
\(\alpha_n^{(k),\tau}
:=\eta\min\{1,\tau/\sqrt{\mathcal J_n^{(k)}}\}\),
with multiplier one when \(\mathcal J_n^{(k)}=0\).
Since the map
\(x\mapsto\eta\min\{1,\tau x^{-1/2}\}\), with value \(\eta\) at \(x=0\),
is Lipschitz on \([0,\infty)\),
\[
\|\alpha_n^{(k),\tau}-\alpha_k^\tau\|_{L^p}
\lesssim n^{-1/2}.
\]
This closes the induction at stage \(k\).

Since \(T\) is fixed,
\begin{equation}
\max_{\substack{0\le k\le T\\1\le i\le N_{\rm tr}}}
\left\|
F_{i,n}^{(k),\tau}
-
f_{i,\infty}^{(k),\tau}
\right\|_{L^2(\Omega;\mathbb R^{d_{\rm out}})}
\le
C_{\rm W}n^{-1/2}.
\label{eq:finite-data-vector-width}
\end{equation}
\end{proof}

\section{Additional Experiments Details}
\label{app:exp}

\subsection{Experimental and Training Details}
\label{app:experimental-details}

All experiments use the linear residual network
\[
h_0(x)=W_0x,
\qquad
h_\ell(x)
=
h_{\ell-1}(x)
+
\frac{1}{\sqrt L}W_\ell h_{\ell-1}(x),
\qquad
f_{n,L}(x)=\frac{1}{n}\xi^\top h_L(x).
\]
At initialization, the entries of $W_0$, $W_\ell$, and $\xi$ are
independent centered Gaussians with variances $d^{-1}$, $n^{-1}$, and
$1$, respectively. Only the residual matrices $W_1,\ldots,W_L$ are trained. The
input projection and readout remain fixed at initialization. HIGGS uses two independent readout vectors $\xi^{(0)}$ and $\xi^{(1)}$
with i.i.d.\ $\mathcal N(0,1)$ entries, each scaled by $n^{-1}$ in the
output layer. Their output difference defines the logit margin. Unless stated otherwise,
training uses 32-bit floating-point arithmetic, with no momentum, weight
decay, learning-rate schedule, or early stopping.

\paragraph{Synthetic regression.}
We minimize
\[
\mathcal L(\theta)
=
\frac{1}{2N}
\sum_{i=1}^N
\bigl(f_\theta(x_i)-y_i\bigr)^2
\]
by full-batch gradient descent with learning rate $1$. We fix a single
dataset draw (dataset seed $0$) with input dimension $d=16$, $N=128$
training examples, and $256$ test examples. The input covariance has a
geometrically spaced spectrum with condition number $10$, and the noiseless
targets are normalized by their empirical RMS on the training set. The
joint-path experiment uses model seeds $0,\ldots,4$ and records
initialization together with $15$ gradient updates. The width and depth
sweeps use model seeds $0,\ldots,19$. Uncertainty intervals for these sweeps
are obtained by resampling model seeds while keeping the dataset fixed.

\paragraph{Empirical rate fits.}
For \cref{fig:width-depth-rate}, the empirical reference
$\widehat f_\infty^{(k)}$ is the coordinatewise average of the $25$ output
vectors at the largest common scale $m=3072$, corresponding to five
width--depth scaling paths and five model seeds. 
The width sweep fixes $L_{\rm ref}=12{,}288$ and uses
\[
n\in\{384,768,1536,3072,6144,12288\},
\]
while the depth sweep fixes $n_{\rm ref}=12{,}288$ and uses
\[
L\in\{5,10,15,20,30,40,60,80,100,120,160,200,240,320,
384,768,1536,3072,6144,12288\}.
\]
The bands in \cref{fig:width-depth-rate}(a)--(b) show one standard deviation
of the seedwise RMS errors, while the band in
\cref{fig:width-depth-rate}(c) is obtained from $500$ paired model-seed
bootstrap refits.

For the width exponent, we fit the slope of
$\log\operatorname{Err}_k(n,L_{\rm ref})$ against $\log n$ over
$n\geq1536$. For the depth exponent, we fit the decreasing regime of the RMS
error curve. The large-$L$ error level is estimated from $L\geq384$ and is
used only to determine this fitting range. The regression itself is performed
on the raw RMS error.

\begin{table}[t]
  \centering
  \caption{
  \textbf{Empirical convergence exponents across training.}
  Confidence intervals are $95\%$ model-seed bootstrap intervals.
  }
  \label{tab:empirical-rate-fits}
  \begin{tabular}{
    @{}c
    @{\hspace{1.5em}}c
    @{\hspace{0.8em}}l
    @{\hspace{2.5em}}c
    @{\hspace{0.8em}}l@{}
  }
    \toprule
    $k$ & $\alpha_n(k)$ & $95\%$ CI
        & $\alpha_L(k)$ & $95\%$ CI \\
    \midrule
    $1$  & $0.427$ & $[0.360, 0.492]$ & $0.738$ & $[0.671, 0.807]$ \\
    $2$  & $0.441$ & $[0.368, 0.508]$ & $0.789$ & $[0.749, 0.837]$ \\
    $3$  & $0.434$ & $[0.369, 0.493]$ & $0.898$ & $[0.870, 0.929]$ \\
    $4$  & $0.437$ & $[0.364, 0.497]$ & $0.979$ & $[0.943, 1.017]$ \\
    $5$  & $0.425$ & $[0.350, 0.498]$ & $0.899$ & $[0.866, 0.936]$ \\
    $6$  & $0.416$ & $[0.335, 0.497]$ & $0.932$ & $[0.898, 0.971]$ \\
    $7$  & $0.413$ & $[0.328, 0.499]$ & $0.966$ & $[0.932, 1.006]$ \\
    $8$  & $0.418$ & $[0.329, 0.506]$ & $0.983$ & $[0.956, 1.013]$ \\
    $9$  & $0.430$ & $[0.341, 0.524]$ & $1.003$ & $[0.975, 1.035]$ \\
    $10$ & $0.447$ & $[0.354, 0.541]$ & $1.019$ & $[0.992, 1.052]$ \\
    $11$ & $0.466$ & $[0.367, 0.561]$ & $1.029$ & $[1.000, 1.063]$ \\
    $12$ & $0.485$ & $[0.379, 0.580]$ & $0.970$ & $[0.933, 1.006]$ \\
    $13$ & $0.503$ & $[0.391, 0.604]$ & $0.968$ & $[0.931, 1.005]$ \\
    $14$ & $0.519$ & $[0.402, 0.622]$ & $0.964$ & $[0.927, 1.002]$ \\
    $15$ & $0.535$ & $[0.409, 0.645]$ & $0.960$ & $[0.922, 0.998]$ \\
    \bottomrule
  \end{tabular}
\end{table}
\cref{tab:empirical-rate-fits} reports the
full trajectories in \cref{fig:width-depth-rate}(c). The width rate remains
close to $n^{-1/2}$ throughout training, whereas the depth-dependent error
decays more rapidly and approaches an empirical $L^{-1}$ rate after the
initial iterations.

\paragraph{HIGGS.}
We use a fixed split of $10{,}500{,}000$ training examples and
$500{,}000$ test examples. Each of the $28$ features is standardized using
the mean and population standard deviation computed on the training split.
Writing $g(x)$ for the class-$1$ versus class-$0$ logit margin, we minimize
the binary logistic loss
\[
\mathcal L_{\rm HIGGS}
=
\frac{1}{N}
\sum_{i=1}^{N}
\left[
\log\bigl(1+e^{g(x_i)}\bigr)-y_i g(x_i)
\right].
\]
We optimize this loss by minibatch SGD with learning rate $1$ and batch size
$4096$. Training runs for $5126$ updates, corresponding to two epochs after
dropping the final incomplete batch in each epoch. The same two training-set
permutations are used for every width, depth, and model seed. We use five
model seeds and evaluate the logit margin on a fixed $4096$-example test
probe. As above, only the residual matrices are updated.

\subsection{Additional Results on Width--Depth Commutativity}
\label{app:joint-additional}

\begin{figure}[H]
  \centering
  \includegraphics[
    width=\textwidth,
    height=0.60\textheight,
    keepaspectratio
  ]{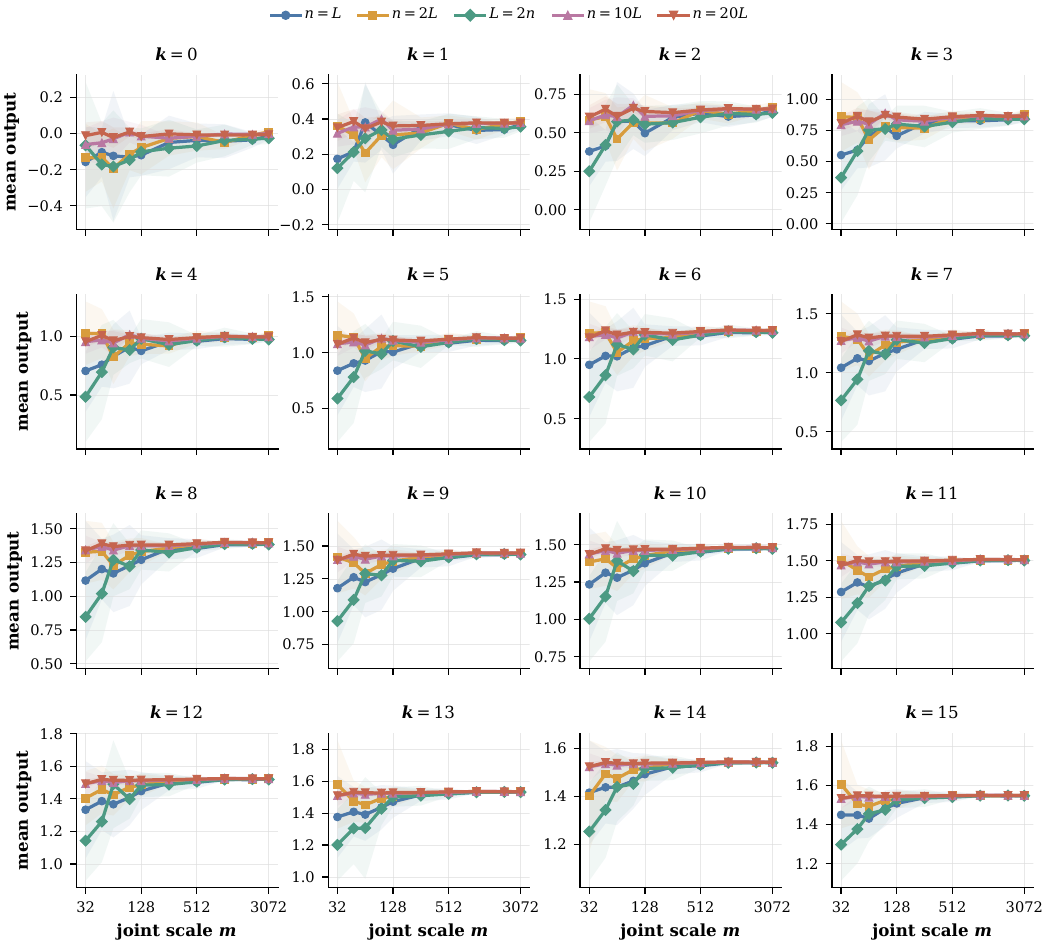}
  \caption{
  \textbf{Synthetic width--depth commutativity across all saved iterations.}
  Each panel shows the output for training input $86$ at one iteration
  $k=0,\ldots,15$. Curves correspond to the five width--depth scaling paths
  in \cref{sec:exp-joint} and show means over five model seeds, with shaded
  regions denoting $\pm1$ standard deviation.
  }
  \label{fig:joint-synthetic-all}
\end{figure}

\begin{figure}[!htbp]
  \centering
  \includegraphics[width=\textwidth]
  {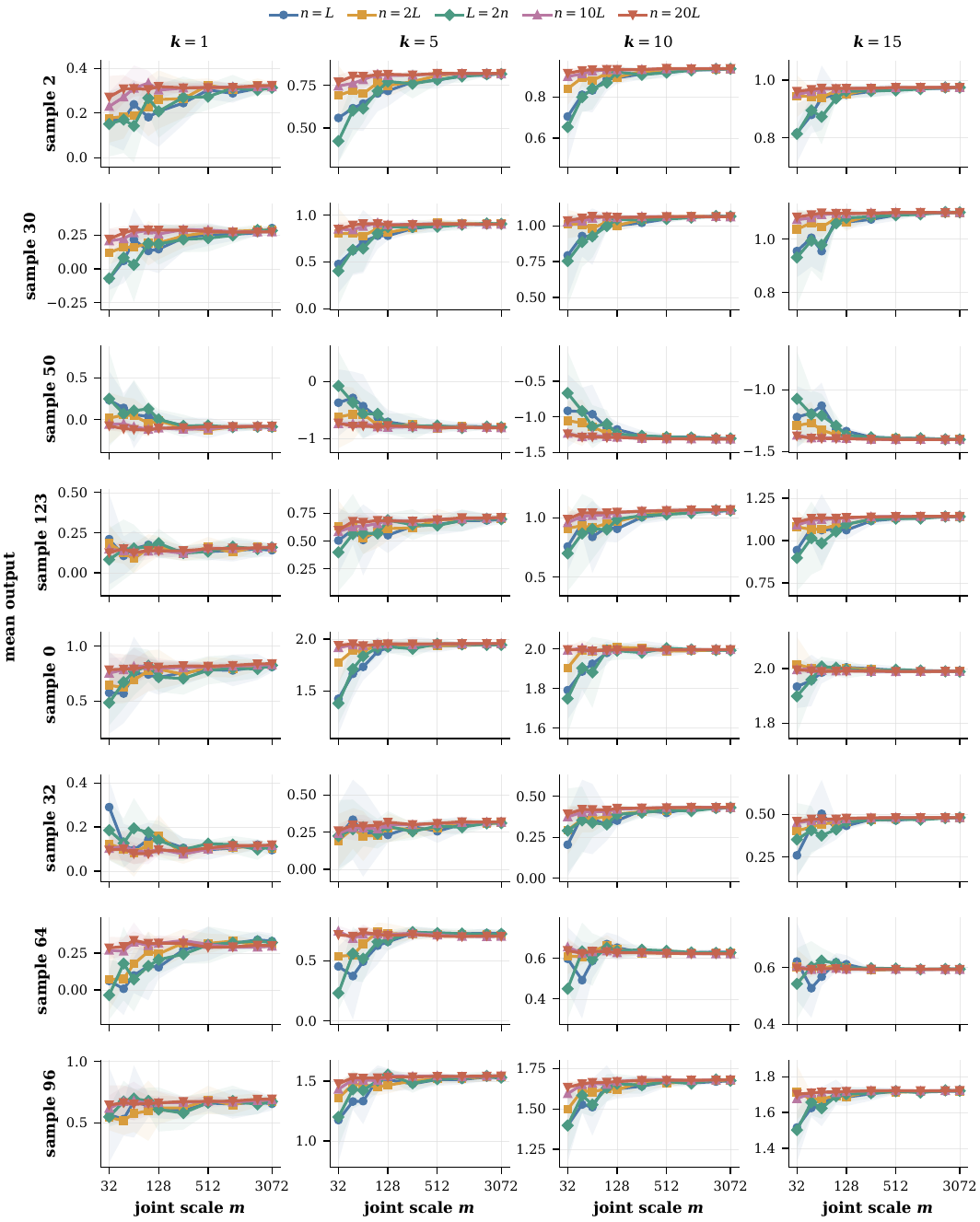}
  \caption{
  \textbf{Synthetic width--depth commutativity at additional training inputs.}
  Rows correspond to four fixed training inputs, and columns show the
  representative iterations in \cref{sec:exp-joint}. Curves correspond to
  the same five width--depth scaling paths and show means over five model
  seeds, with shaded regions denoting $\pm1$ standard deviation.
  }
  \label{fig:joint-multiple-points}
\end{figure}

\paragraph{Synthetic regression.}
\cref{sec:exp-joint} reports four representative training iterations for one
fixed training input. We additionally report the complete trajectory over
$k=0,\ldots,15$ in \cref{fig:joint-synthetic-all}, using the common scales
\[
m\in\{32,48,64,96,128,256,512,1024,2048,3072\},
\]
with five model seeds at each scale. We also report four additional fixed
training inputs in \cref{fig:joint-multiple-points}.

\begin{figure}[!htbp]
  \centering
  \includegraphics[width=\textwidth]
  {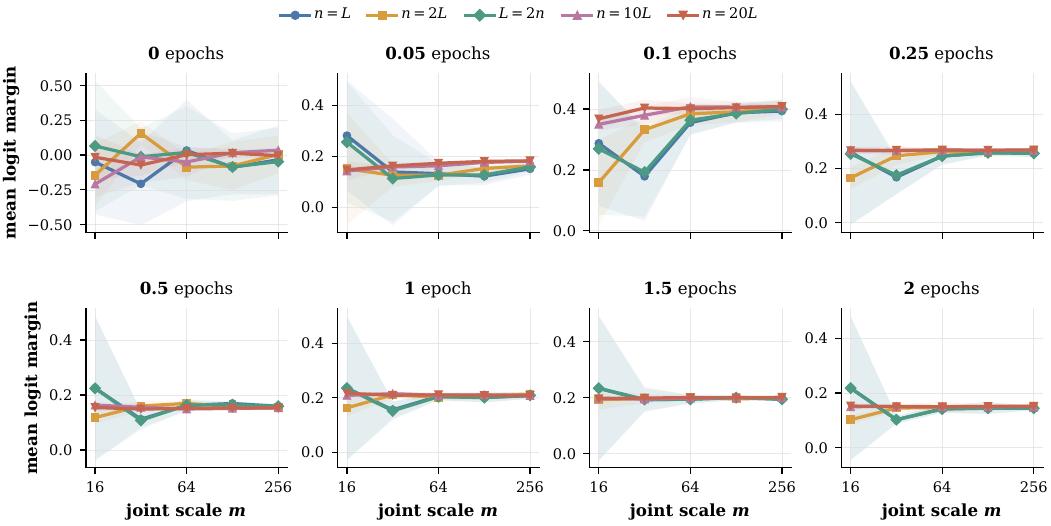}
  \caption{
  \textbf{HIGGS width--depth commutativity across all saved checkpoints.}
  Each panel shows the class-$1$ versus class-$0$ logit margin for the fixed
  test example used in \cref{sec:exp-joint}, from initialization through two
  training epochs. Curves show means over five model seeds, with shaded
  regions denoting $\pm1$ standard deviation, over the common scales
  $m\in\{16,32,64,128,256\}$.
  }
  \label{fig:joint-higgs-all}
\end{figure}

\begin{figure}[!htbp]
  \centering
  \includegraphics[width=\textwidth]
  {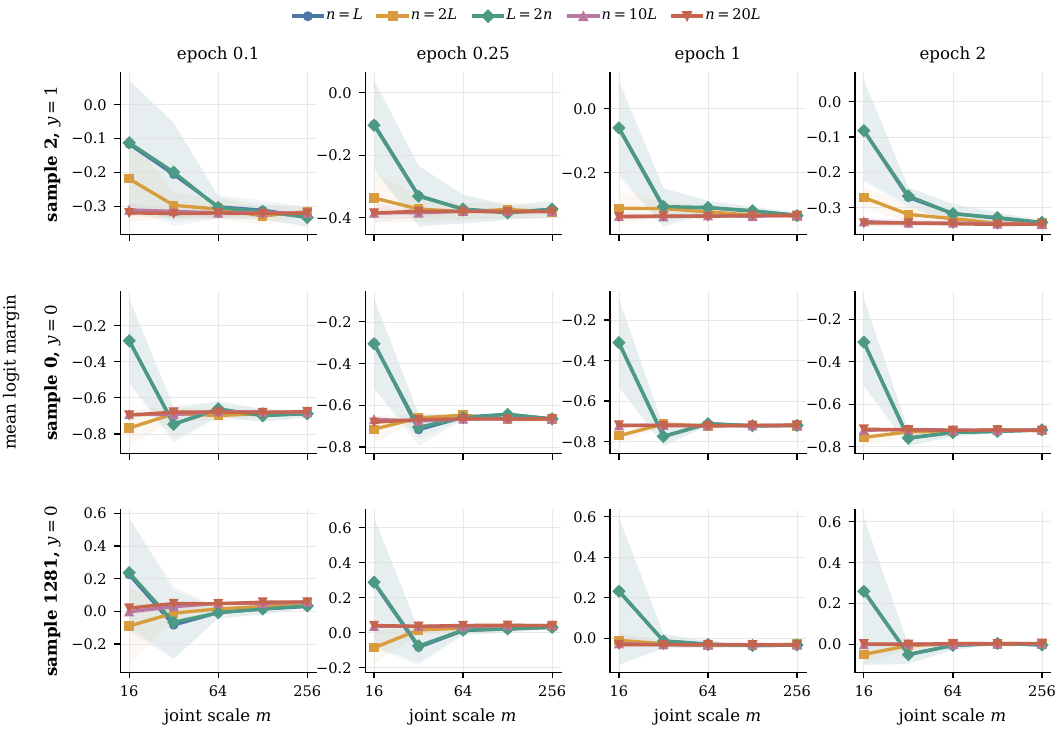}
  \caption{
  \textbf{HIGGS width--depth commutativity at additional test examples.}
  Rows correspond to three fixed test examples, and columns show the
  representative checkpoints in \cref{sec:exp-joint}. Curves correspond to
  the same five width--depth scaling paths and show means over five model
  seeds, with shaded regions denoting $\pm1$ standard deviation.
  }
  \label{fig:joint-higgs-multiple-points}
\end{figure}

\paragraph{HIGGS.}
\cref{sec:exp-joint} reports four representative training checkpoints for one
fixed HIGGS test example. We additionally report all eight saved checkpoints,
including initialization, in \cref{fig:joint-higgs-all}, together with three
additional fixed test examples in
\cref{fig:joint-higgs-multiple-points}.

\subsection{Single-Datum Experiment}
\label{app:single-datum}

The main experiments use a multi-sample regression task. To more closely
match the single-datum setting of our theory, we additionally train on a
single input--target pair with $q_0=1$, $y_\star=1$, learning rate $1$, and
five independent model seeds. We evaluate the same five width--depth scaling paths at $m=\min(n,L)\in\{128,256,512,1024,2048,3072\},$ for training iterations $k=0,\ldots,10$.

\begin{figure}[!htbp]
  \centering
  \includegraphics[
    width=\textwidth,
    height=0.68\textheight,
    keepaspectratio
  ]{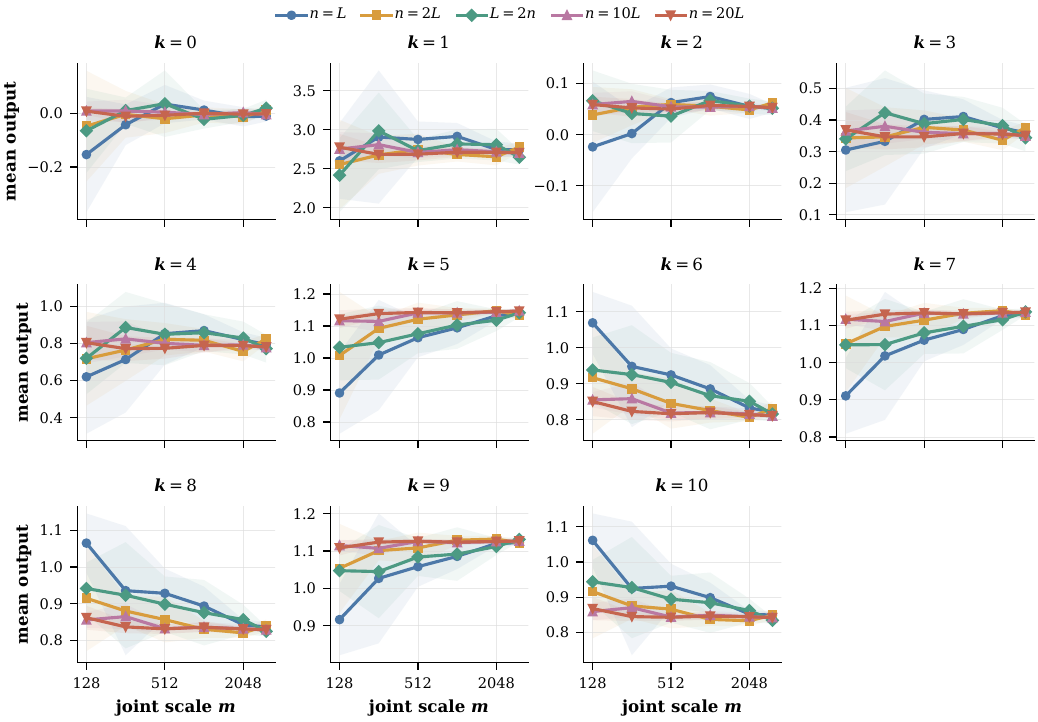}
  \caption{
  \textbf{Width--depth commutativity in the single-datum setting.}
  Each panel shows the trained scalar output at one training iteration along
  the five width--depth scaling paths. Curves show the mean over five model
  seeds, with shaded regions denoting $\pm1$ standard deviation. The paths
  exhibit substantial finite-scale differences at small $m$, but these
  differences diminish as both width and depth increase.
  }
  \label{fig:single-datum-joint}
\end{figure}

Figure~\ref{fig:single-datum-joint} shows the same transition from
finite-scale path dependence to large-scale path independence as in the
multi-sample experiment.

\subsection{Experimental Details}
\label{app:computational-details}

\paragraph{Lazy Gaussian implementation.}
The largest synthetic model has $(n,L)=(61{,}440,3{,}072)$, for which
explicitly storing the residual matrices in float32 would require
approximately $46.4$ TB of memory. We instead use an exact-in-law lazy
Gaussian implementation based on Gaussian conditioning and deferred
decisions \citep{Bayati_2011}, which generates the required matrix--vector
products without materializing the residual matrices. This reduces the peak
GPU memory usage to $46.9$ GB and makes the largest experiments computationally
feasible.

\begin{table}[H]
  \centering
  \caption{
  \textbf{Memory usage for the reported experiments.}
  Dense memory is the float32 storage required to materialize all residual
  matrices; peak memory is the maximum recorded CUDA allocation.
  }
  \label{tab:computational-resources}
  \small
  \setlength{\tabcolsep}{6pt}
  \begin{tabular}{lcccc}
    \toprule
    Experiment
    & Maximum $(n,L)$
    & Backend
    & Dense memory
    & Peak memory \\
    \midrule
    Single datum
    & $(61{,}440,3{,}072)$
    & lazy
    & $46.4$ TB
    & $31.7$ GB \\
    Synthetic commutativity
    & $(61{,}440,3{,}072)$
    & lazy
    & $46.4$ TB
    & $46.9$ GB \\
    Linear rate sweeps
    & $(12{,}288,12{,}288)$
    & lazy
    & $7.42$ TB
    & $25.4$ GB \\
    HIGGS commutativity
    & $(5{,}120,256)$
    & dense
    & $26.8$ GB
    & $26.9$ GB \\
    Nonlinear activations
    & $(5{,}120,256)$
    & dense
    & $26.8$ GB
    & $27.6$ GB \\
    \bottomrule
  \end{tabular}
\end{table}

\end{document}